\documentclass[10pt, letter, onecolumn]{arxiv}

\usepackage{kantlipsum, lipsum}
\usepackage{dm-colors}
\usepackage{amsmath}
\usepackage{pstricks, pst-node}
\usepackage{verbatim}
\usepackage{multirow}
\usepackage{scalerel}
\usepackage{booktabs}
\usepackage{enumitem}
\usepackage{xspace}
\usepackage{bm}
\usepackage{bbm}
\usepackage{mathtools}
\usepackage{soul}
\usepackage{epsfig}
\usepackage{graphicx}
\usepackage{tcolorbox}
\usepackage{subcaption}
\usepackage{amssymb}
\usepackage{colortbl}
\usepackage{csquotes}
\usepackage{etex}
\usepackage{setspace}
\usepackage{svg}
\usepackage{colortbl}
\usepackage{tabularx,ragged2e}
\usepackage{placeins}
\usepackage{makecell}
\usepackage[symbol]{footmisc}
\usepackage[bibstyle=nature,citestyle=numeric,%
            natbib=true,backend=biber,maxbibnames=99,%
            giveninits=false,sorting=none]{biblatex}
\usepackage{nameref}
\usepackage{varioref}
\usepackage[pagebackref=false,breaklinks=false,%
            colorlinks=true,bookmarks=true,citecolor=ourdarkblue,%
            urlcolor=ourdarkblue,linkcolor=ourdarkblue]{hyperref}
\usepackage{appendix}
\usepackage[noabbrev,capitalize]{cleveref}
\usepackage{etoc}

\graphicspath{{figures/}}

\def\etc{{\em etc}\xspace}

\def\ourmodel{{TxGemma}\xspace}
\def\ouragent{{Agentic-Tx}\xspace}
\def\basemodel{{Gemma-2}\xspace}
\def\oldbasemodel{{PaLM-2}\xspace}
\def\ouroldmodel{{Tx-LLM}\xspace}

\def\ourmodelpredict{{\ourmodel-Predict}\xspace}
\def\ourmodelpredictsmall{{\ourmodel-2B-Predict}\xspace}
\def\ourmodelpredictnine{{\ourmodel-9B-Predict}\xspace}
\def\ourmodelpredictlargest{{\ourmodel-27B-Predict}\xspace}
\def\ourmodelconverse{{\ourmodel-Chat}\xspace}
\def\ourmodelconversenine{{\ourmodel-9B-Chat}\xspace}
\def\ourmodelconverselargest{{\ourmodel-27B-Chat}\xspace}

\def\basemodelsmall{{\basemodel-2B}\xspace}
\def\basemodelnine{{\basemodel-9B}\xspace}
\def\basemodellargest{{\basemodel-27B}\xspace}

\def\ouroldmodels{{\ouroldmodel~S}\xspace}
\def\ouroldmodelm{{\ouroldmodel~M}\xspace}

\def\ntasks{{66}\xspace}
\def\nbetterthansota{{26}\xspace}
\def\natleastnearsota{{50}\xspace}
\def\ntools{{18}\xspace}

\title{TxGemma: \\Efficient and Agentic LLMs for Therapeutics}

\author[$\ast,\dagger$,1]{Eric Wang}
\author[$\ast$,1]{Samuel Schmidgall}
\author[1]{Paul F. Jaeger}
\author[2]{Fan Zhang}
\author[2]{Rory Pilgrim}
\author[2]{\\Yossi Matias}
\author[1]{Joelle Barral}
\author[1]{David Fleet}
\author[$\dagger$,1]{Shekoofeh Azizi}

\affil[1]{Google DeepMind, }
\affil[2]{Google Research }

\renewcommand{\correspondingauthor}[1]{
    $\ast$~Equal contributions.\\%
    $\dagger$~Corresponding authors: \{shekazizi,ericzwang\}@google.com
}

\begin{document}
\begin{refsection}

\begin{abstract}
Therapeutic development is a costly and high-risk endeavor that is often plagued by high failure rates. To address this, we introduce \ourmodel, a suite of efficient, generalist large language models (LLMs) capable of therapeutic property prediction as well as interactive reasoning and explainability. Unlike task-specific models, \ourmodel synthesizes information from diverse sources, enabling broad application across the therapeutic development pipeline. The suite includes 2B, 9B, and 27B parameter models, fine-tuned from \basemodel on a comprehensive dataset of small molecules, proteins, nucleic acids, diseases, and cell lines. Across 66 therapeutic development tasks, \ourmodel achieved superior or comparable performance to the state-of-the-art generalist model on 64 (superior on 45), and against state-of-the-art specialist models on 50 (superior on 26). Fine-tuning \ourmodel models on therapeutic downstream tasks, such as clinical trial adverse event prediction, requires less training data than fine-tuning base LLMs, making \ourmodel suitable for data-limited applications. Beyond these predictive capabilities, \ourmodel features conversational models that bridge the gap between general LLMs and specialized property predictors. These allow scientists to interact in natural language, provide mechanistic reasoning for predictions based on molecular structure, and engage in scientific discussions. Building on this, we further introduce \ouragent, a generalist therapeutic agentic system powered by Gemini 2.5 that reasons, acts, manages diverse workflows, and acquires external domain knowledge. \ouragent surpasses prior leading models on the Humanity's Last Exam benchmark (Chemistry \& Biology) with 52.3\% relative improvement over o3-mini (high) and 26.7\% over o3-mini (high) on GPQA (Chemistry).  On ChemBench, \ourmodel excels with improvements of 6.3\% (ChemBench-Preference) and 2.4\% (ChemBench-Mini) over o3-mini (high), as well as 17.7\% and 5.6\% over o1, respectively.  \ourmodel's collection is released as \href{https://github.com/google-gemini/gemma-cookbook/tree/main/TxGemma}{open models}, enabling researchers to adapt and validate it on their own diverse datasets, thus facilitating more challenging real-world applications.

\end{abstract}

\maketitle


\begin{figure}[]
    \centering
    \includegraphics[width=1.0\textwidth]{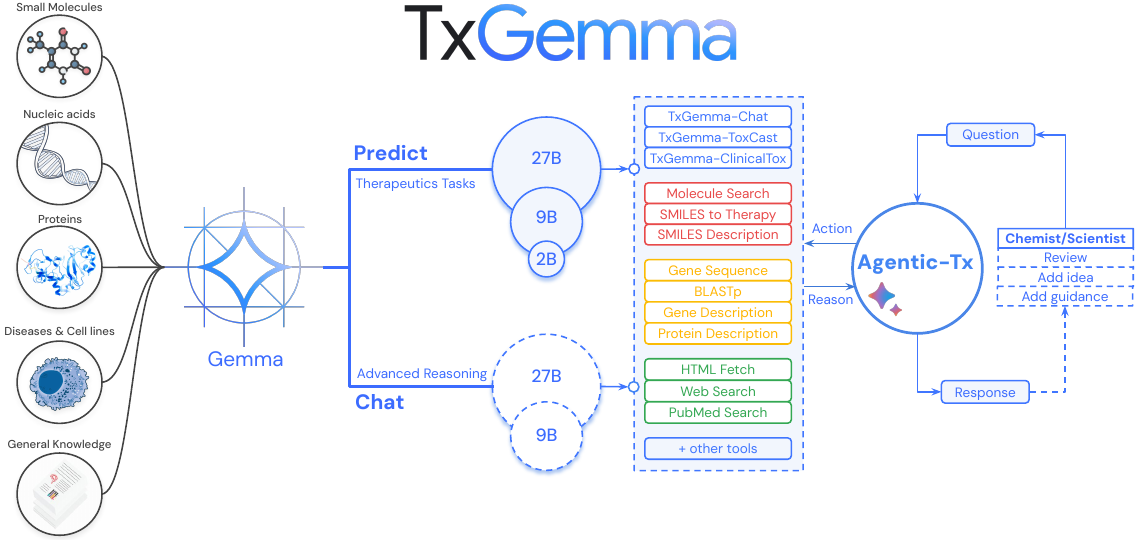} \\
    \vspace{10pt}
    \includegraphics[width=0.48\textwidth]{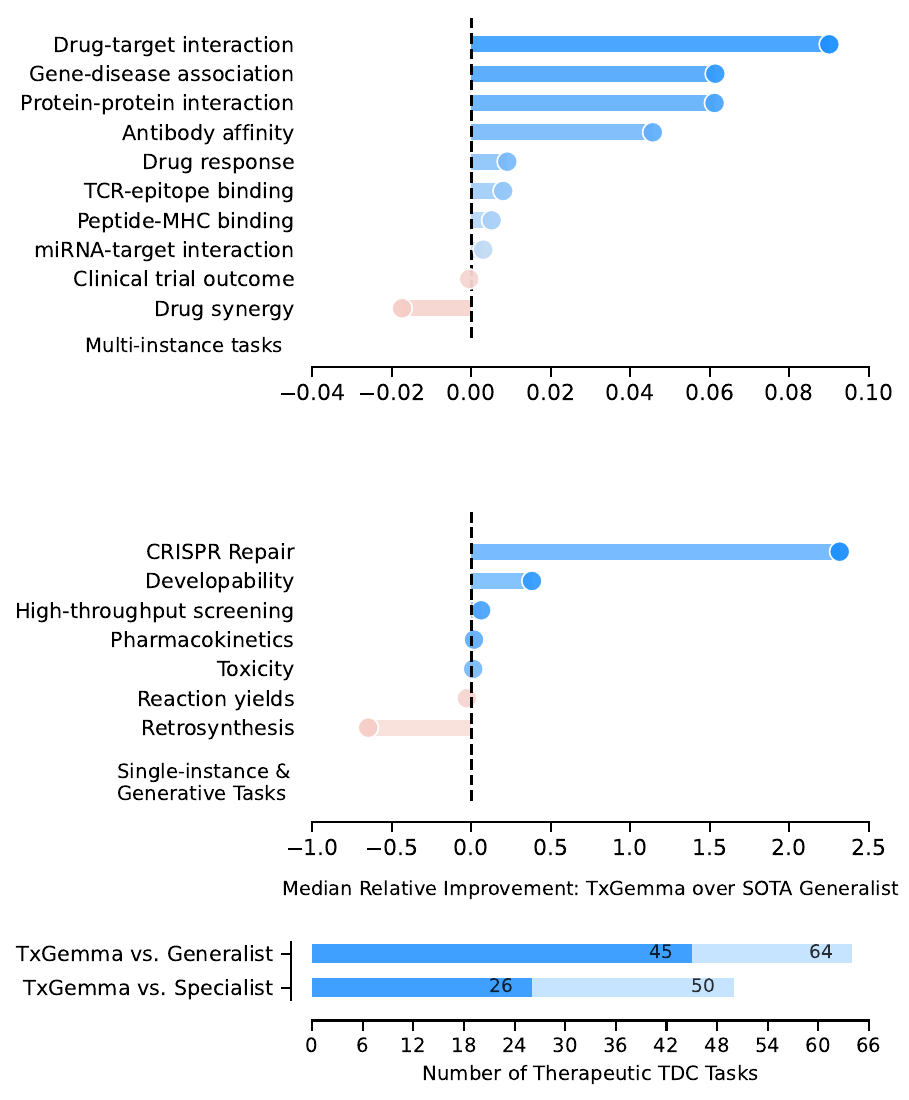} \hfill
    \includegraphics[width=0.48\textwidth]{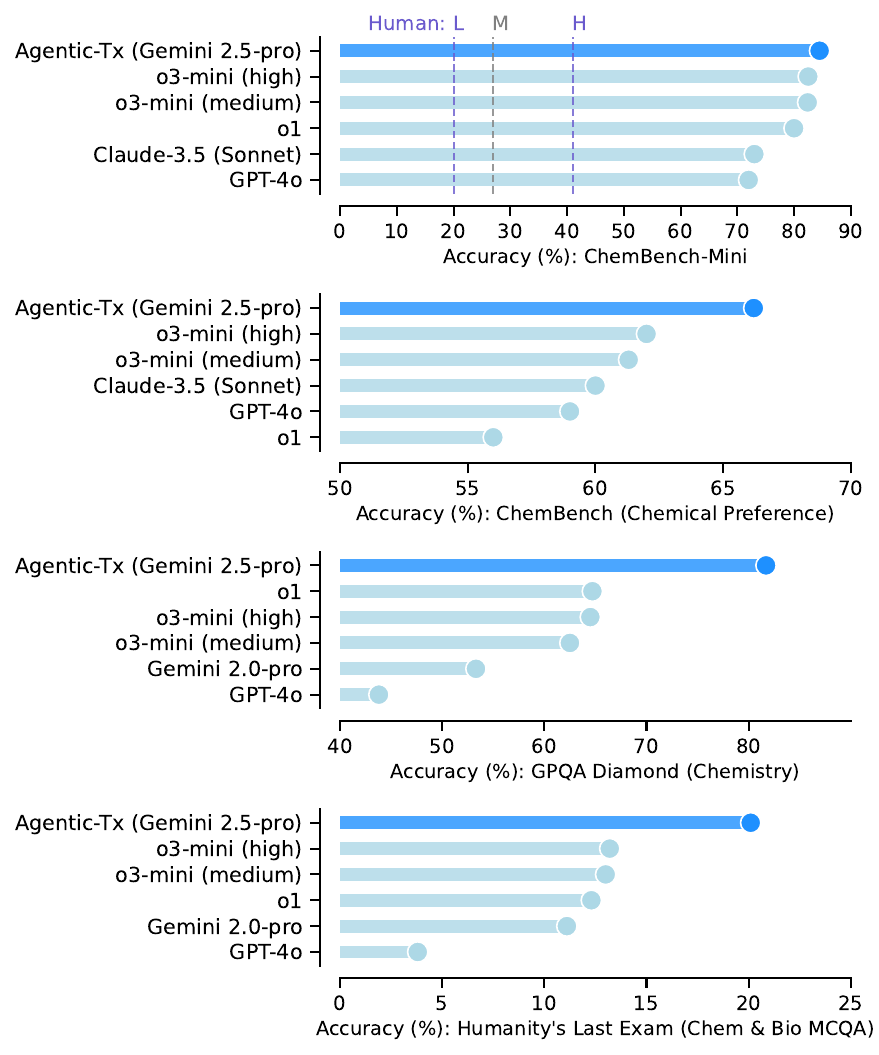}
    \caption{\small{\textbf{Overview of \ourmodel.} (\textbf{top}) All \ourmodel variants are trained on diverse data sources of the Therapeutic Data Commons (TDC). \ourmodelpredict comes in three size variants (2B, 9B, and 27B) and is trained for high-performance predictions on a broad set of therapeutic development tasks. \ourmodelconverse features two variants (9B and 27B) and is trained on a combination of TDC data with general \basemodel instruction tuning data to retain conversational and reasoning capabilities. \ouragent, a therapeutics-focused agentic system powered by Gemini 2.5, has access to \ntools tools including \ourmodelpredict and \ourmodelconverse to collect external knowledge and manages complex tasks in either autonomous or interactive settings. (\textbf{bottom-right}) Absolute performance of \ouragent compared to best-in-class models on three complex therapeutic-related reasoning benchmarks. The state-of-the-art (SOTA) values are obtained from~\cite{mirza2024large,OpenAIReasoningLLMs} and details are listed in~\cref{tab:tx-agent-perf}. Dashed lines: L=lowest, M=mean, H=highest human scores. (\textbf{bottom-left}) Relative performance changes of \ourmodelpredict compared to the SOTA generalist model for each task type. The assignment of the 66 evaluated TDC tasks to task types is shown in Tables~\ref{tab-sup:binary_dataset_sizes} and~\ref{tab-sup:regression_generation_dataset_sizes}.} The bottom bar chart shows a summary of results where \ourmodelpredict outperforms or nearly matches SOTA (light blue), and outperforms SOTA (darker blue).}
    \label{fig:overview} 
\end{figure}

\section{Introduction}
\label{sec:introduction}

The pharmaceutical industry faces significant challenges in bringing new therapeutics to market. High attrition rates and lengthy, costly development timelines~\cite{sun2022why,hinkson2020accelerating} necessitate innovative approaches to therapeutic development. Success requires a drug candidate to not only demonstrate efficacy but also possess favorable safety, metabolic stability, pharmacokinetic/pharmacodynamic properties and developability, among other characteristics. Determining these diverse characteristics often relies on a large array of complex and expensive experimental procedures, highlighting the need for more efficient methods.

Computational approaches, such as machine learning, are emerging as powerful tools to address these challenges. Leveraging predictive models trained on curated datasets allows researchers to prioritize promising candidates early in the development process, reducing reliance on costly experimental assays \cite{kumar2012fragment}. Publicly available databases of molecular properties and biological activity are crucial for training and validating these models. In this area, a major development was the curation of the Therapeutics Data Commons (TDC)~\cite{velez2024tdc,huang2021therapeutics,huang2022artificial}, which contains datasets and benchmarks for many different tasks throughout the therapeutic development pipeline, ranging from early-stage target identification to late-stage clinical trial approval.

Recent advancements in large language models (LLMs) offer a compelling opportunity to leverage available datasets and address limitations in the therapeutic development process. LLMs have demonstrated the capacity to integrate and learn from diverse data sources across various domains, including scientific applications~\cite{bubeck2023sparks,taylor2022galactica,telenti2024large}. Their potential to connect disparate aspects of drug development, such as chemical structure, biological activity, and clinical trial outcomes, is particularly exciting. In this context, we have previously introduced \ouroldmodel, a LLM fine-tuned from a collection of question-answer instruction-tuning datasets based on TDC~\cite{chaves2024txllm}. While promising, the model's lack of conversational capabilities prevented reasoning or user interaction, limiting its value for scientists who require a model that can understand complex queries and engage in nuanced discussions.

In this work, we introduce \ourmodel, a suite of \emph{efficient, generalist} LLMs trained for therapeutics. Building on, but significantly extending, our previous work~\cite{chaves2024txllm}, \ourmodel leverages LLMs to synthesize information from diverse sources. The suite includes 2B, 9B, and 27B parameter models, fine-tuned from Gemma-2~\cite{team2024gemma,team2024gemma2} using a collection of therapeutic instruction-tuning datasets encompassing small molecules, proteins, nucleic acids, diseases, and cell lines. For the first time in therapeutic AI, \ourmodel features conversational counterparts capable of reasoning and explanation, moving beyond black-box predictions to facilitate mechanistic understanding and scientific discussions. Our key contributions are as follows:

\begin{itemize}[leftmargin=2.0em,rightmargin=0em]
    \item \textbf{Efficient Generalist Therapeutic LLMs:} \ourmodel represents a potential shift from task-specific AI to \emph{efficient} \emph{generalist} models in therapeutic development. These efficient LLMs (2B-27B parameters) offer a competitive alternative to specialized models, achieving strong performance across a broad range of predictive and generative tasks. Out of 66 therapeutic development tasks curated by TDC, \ourmodelpredict outperforms or nearly matches the state-of-the-art generalist model on 64 (outperforms on 45) and state-of-the-art specialist models on 50 (outperforms on 26). Additionally, fine-tuning \ourmodel models on clinical trial adverse event prediction requires less data to achieve strong performance compared to base \basemodel models, an important advantage for data-limited fields.
    \vspace{+6pt}
    \item \textbf{Explainable and Interactive Therapeutic Models:} \ourmodelconverse introduces reasoning and explanation capabilities, bridging the gap between general LLMs and specialized property predictors. Scientists can interact with \ourmodelconverse using natural language, exploring complex questions, receive explanations for predictions (e.g., based on molecular structure), and engage in scientific discussions.
    \vspace{+6pt}
    \item \textbf{Agentic Orchestration of Therapeutic Development Workflows:} We further introduce \ouragent, a therapeutics-focused agentic system powered by Gemini 2.5, demonstrating how \ourmodel models can be integrated as tools. Equipped with \ntools tools, \ouragent solves complex, multi-step problems, achieving state-of-the-art results on reasoning-intensive chemistry and biology benchmarks, including Humanity's Last Exam~\cite{phan2025humanity} and ChemBench~\cite{mirza2024large}.
    \vspace{+6pt}
    \item \textbf{Enabling Innovative Research with Open Models:} Understanding the prevalence of proprietary data in therapeutic research, we release \ourmodel models trained only on datasets with commercial licenses as open models to empower researchers to adapt and refine them on their own data. This facilitates validation and potential performance improvements tailored to their specific research needs, paving the way for therapy safety and efficacy in more challenging real-world therapeutic applications.

\end{itemize}

\section{Methods}
\label{sec:methods}

\subsection{Data}
\label{subsec:datasets}
\textbf{Therapeutic Data Commons (TDC)}
We leverage the Therapeutic Data Commons (TDC)~\cite{huang2021therapeutics,velez2024tdc}, a comprehensive collection of 66 AI-ready datasets spanning the drug discovery and development pipeline. TDC includes over 15 million datapoints across various biomedical entities and encompasses single-instance prediction, multi-instance prediction, and generation tasks~\cite{huang2021therapeutics}. We focus on TDC tasks relevant to drug discovery, incorporating diverse therapeutic representations: SMILES strings (small molecules), amino acid sequences (proteins and peptides, including specialized representations for MHC molecules and T-cell receptors), nucleotide sequences (nucleic acids), and natural language text (disease/cell line names) (see \cref{tab-sup:data-representation} for examples). Many tasks combine multiple representations. (See \cref{tab-sup:excluded_results} for task inclusion criteria and \cref{tab:example_prompts_binary,tab:example_prompts_regression_generation} for biological contexts of certain tasks.)

\textbf{Therapeutic Instruction-Tuning}
Following~\citet{chaves2024txllm}, we transform the raw TDC data into an instruction-tuning format suitable for LLMs.  Each data point is formatted as a prompt:
\begin{itemize}[leftmargin=2.5em,rightmargin=0em]
    \item \textbf{Instruction:} Briefly describes the task.
    \item \textbf{Context:} Provides 2-3 sentences of relevant biochemical background, derived from TDC descriptions and literature.
    \item \textbf{Question:} Queries a specific therapeutic property, incorporating textual representations of therapeutics and/or targets (e.g., \textit{``Does the following molecule cross the blood-brain barrier? <molecule>''}).
    \item \textbf{Answer:} Formatted as (A)/(B) for binary classification, a binned continuous value for regression, or a SMILES string for generation.
\end{itemize}

This process yields 7,080,338 training, 956,575 validation, and 1,917,297 test data points (\cref{fig-sup:dataset_size}, \cref{tab-sup:binary_dataset_sizes,tab-sup:regression_generation_dataset_sizes}). Data splits closely follow TDC's recommended methodologies (random, scaffold, cold-start, combination, temporal) (\cref{tab-sup:binary_dataset_sizes}, \cref{tab-sup:regression_generation_dataset_sizes}).  Detailed task descriptions are in \cref{tab-sup:binary_descriptions,tab-sup:regression_generation_descriptions}.

We employ a few-shot prompting strategy to promote in-context learning~\cite{brown2020language}, using a blend of 70\% zero-shot and 30\% few-shot prompts~\cite{longpre2023flan,chaves2024txllm}. For few-shot prompts, we randomly sample examples from the training set (\cref{tab:example_prompts_fewshot}), as intra-training set similarity is higher than training-test set similarity (\cref{fig:nn_distribution}). The number of examples is uniformly selected between 1 and 10 so that few-shot prompting is robust to the number of examples during evaluation.

\subsection{Modeling}\label{subsec:modeling}

\textbf{Base LLM. } \ourmodel is built upon the \basemodel~\cite{team2024gemma2} family of lightweight, state-of-the-art open LLMs. \basemodel models utilize a decoder-only transformer architecture, incorporating architectural modifications such as interleaved local-global attention and group-query attention, and are trained using Gemini technology~\cite{team2023gemini}. We utilize \basemodel base models at 2B, 9B, and 27B parameters. 2B and 9B \basemodel models were initially trained via knowledge distillation~\cite{team2024gemma2}.

\textbf{Predictive Model Fine-Tuning. } We fine-tune the 2B, 9B, and 27B \basemodel base models on the therapeutic instruction-tuning data derived from TDC, creating \ourmodelpredictsmall, \ourmodelpredictnine, and \ourmodelpredictlargest, respectively. Training was performed across all TDC tasks, with mixture ratios proportional to the number of training data points (see \cref{tab-sup:binary_dataset_sizes,tab-sup:regression_generation_dataset_sizes} for data distribution). This encompassed all approximately 7 million training examples, comprising 3.3 million from regression/generation and 3.7 million from binary classification tasks. Fine-tuning proceeded for 67B tokens (12 epochs) using 256 TPUv4 chips with 8-way data replication, 4-way sequence sharding, and 4-way model sharding. In this work, ``\ourmodel'' generally refers to the generalist, predictive \ourmodelpredictlargest.

\textbf{Conversational Model Fine-Tuning. } We also trained conversational counterparts, \ourmodelconversenine and \ourmodelconverselargest, by supplementing the therapeutic instruction-tuning data with general instruction-tuning data, as detailed in the \basemodel report~\cite{team2024gemma2}. The training data mixture comprised 30\% therapeutic data and 70\% general instruction-tuning data. Conversational models were trained using the same number of tokens and TPU configuration as the predictive models.

\subsection{Evaluating Predictive Performance}
\textbf{Prompting strategy}
For test set evaluations, we use 10-shot prompting, selecting exemplars from the nearest neighbors within the combined training and validation set (not the test set), as detailed in~\cref{tab:example_prompts_fewshot}. Nearest neighbors were determined using different methods based on molecule type. For small molecules, we used RDKit \cite{landrum2016rdkit} to generate Morgan fingerprints (radius 2 and size 2048), representing molecular substructures as binary vectors. Subsequently, we used Chemfp \cite{dalke2019chemfp} to compute Tanimoto similarities, which quantify fingerprint overlap. For amino acid and nucleotide sequences, nearest neighbors were defined by percent sequence identity, determined through multiple sequence alignments performed with Clustal Omega~\cite{sievers2011fast}.

\textbf{Performance Metrics and Statistical Tests}
We assess performance using the preferred metrics for each task, as defined by TDC~\cite{huang2021therapeutics} and used by previous models. Binary classification tasks are assessed with area under the receiver operating characteristic curve (AUROC), area under the precision-recall curve (AUPRC), and accuracy. Regression tasks use Spearman's and Pearson correlation coefficients, mean absolute error (MAE), and mean squared error (MSE). The USPTO generation task uses "set accuracy," scoring 1 for perfect overlap between predicted and true reactant sets, and 0 otherwise. Bootstrapped metrics are calculated using 1000 samples. To compare overall performance between two models across all TDC tasks, we use the non-parametric Wilcoxon signed-rank test and report the corresponding p-value (details in Appendix~\ref{subsec:agg_comparison}).

\subsection{Agentic System} \label{sec:agentic_methods}
One limitation of LLMs for discovery is that, while their prediction capabilities are powerful, they do not have access to up-to-date external knowledge, such as research articles or domain-specific prediction models. These knowledge cut-offs prevent the model from answering questions outside of its training scope. Additionally, some questions involve multiple reasoning steps to solve, for example, the question ``What structural modifications could improve the potency of the given drug?'' requires iteratively searching the drug's structural space and then prompting \ourmodel to predict potency.

\ouragent, our therapeutics-focused agentic system powered by Gemini 2.5~\cite{team2023gemini}, extends \ourmodel's capabilities by orchestrating such complex workflows. \ouragent employs a modular, tool-usage paradigm, in contrast to \ourmodel's direct generation of solutions.

\begin{figure}[!t]
    \centering
    \includegraphics[width=1.0\textwidth]{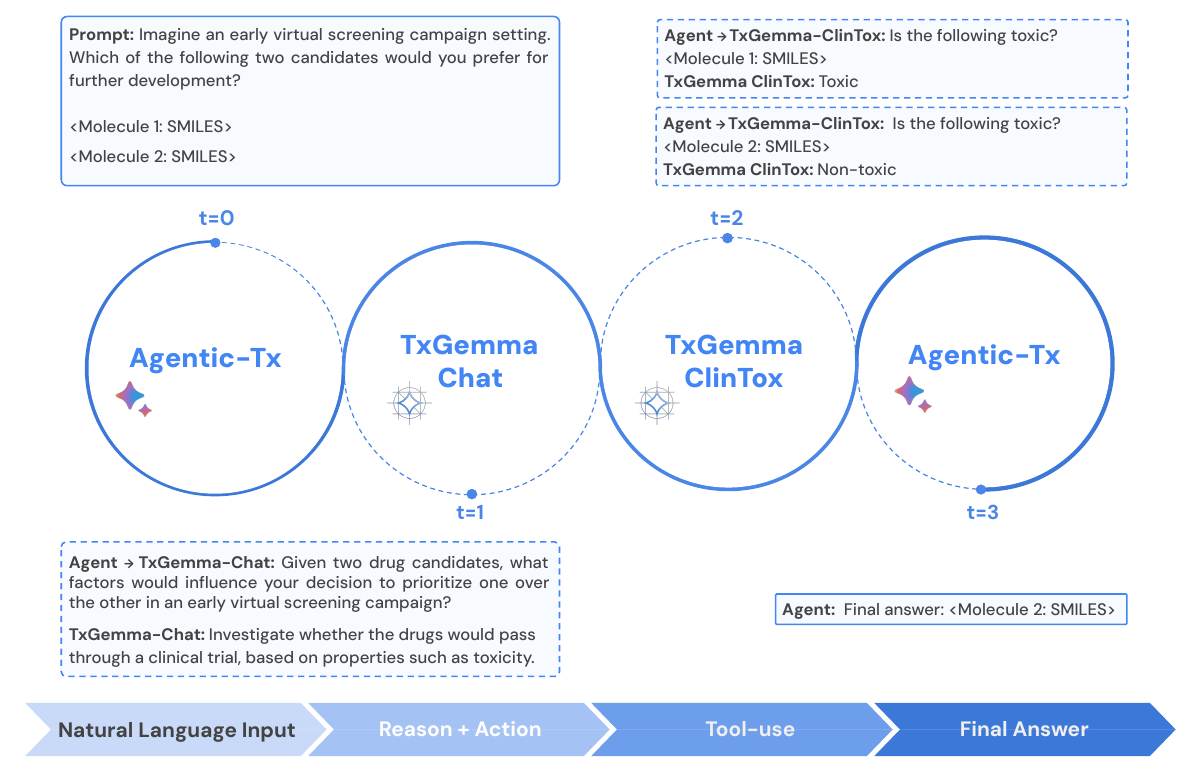}
    \caption{\textbf{Example workflow of agentic planning and execution with \ouragent.} \ouragent uses the ReAct framework~\cite{yao2022react} to interleave thought with tool-usage. When a user poses a query, \ouragent checks whether the query structure matches any defined tool trigger. If so, the query is routed to the corresponding tool, which (i) parses the request, (ii) invokes specialized logic, and (iii) returns a structured answer to the agent. The agent then composes a user-facing response. This adaptive tool-use mechanism is especially helpful for tasks that require external references, chemical data transformations, or precise chemical information, areas where self-contained LLMs often hallucinate. In the displayed example, \ouragent uses two tools to solve a complex therapeutic task: \ourmodelconverse and the clinical toxicity prediction tool based on \ourmodelpredict.}
    \label{fig:txagent-overview}
\end{figure}

\textbf{Reasoning and Action Framework}
\ouragent utilizes the ReAct framework~\cite{yao2022react}, allowing it to interleave reasoning steps (``thoughts'') with actions (tool use). The agentic system receives a task or question and iteratively takes actions based on its current context. Each action typically involves using a tool, which returns an observation. Key to ReAct is this iterative process of observing, reasoning, and acting, allowing \ouragent to dynamically adjust its approach based on the information it gathers. Because tools may return large outputs, we summarize these observations in order to maintain a concise and relevant context. This iterative process of observing, reasoning, acting, and updating its context allows \ouragent to dynamically adjust its approach and gather the necessary information to answer the initial query. Finally, \ouragent integrates the gathered information and formulates a user-friendly response.

\textbf{Agentic Tools}
\ouragent is equipped with \ntools tools across four categories (detailed tool descriptions are in \cref{tab-sup:txagent-tools}). They can be broadly categorized as:
\begin{enumerate}
    \item \textbf{\ourmodel-based Tools:} These provide access to \ourmodel's capabilities. The \textit{Chat} tool enables interaction with \ourmodelconverselargest. The \textit{ClinicalTox} and \textit{ToxCast} tools utilize \ourmodelpredictlargest for toxicity predictions. \textit{IC$_{50}$} returns the predicted normalized IC$_{50}$ between a drug and protein, the \textit{Mutagenicity} tool predicts drug mutagenicity, and the \textit{Phase1 Trial} tool predicts whether a drug would pass a Phase 1 clinical trial.
    \item \textbf{General Tools:} These query external knowledge resources, including PubMed, Wikipedia, and the web.
    \item \textbf{Molecule Tools:} These leverage domain-specific libraries for tasks such as retrieving molecular descriptors (e.g., from PubChem) and performing chemical structure conversions.
    \item \textbf{Gene \& Protein Tools:} These leverage domain-specific libraries for tasks involving genes or proteins, such as retrieving gene descriptions and protein descriptions (e.g., from the NCBI Gene database).
\end{enumerate}

\section{Results}
\label{sec:results}

\subsection{\ourmodel Predictive Performance}

\begin{figure}[!t]
    \centering
    \includegraphics[width=1.0\textwidth]{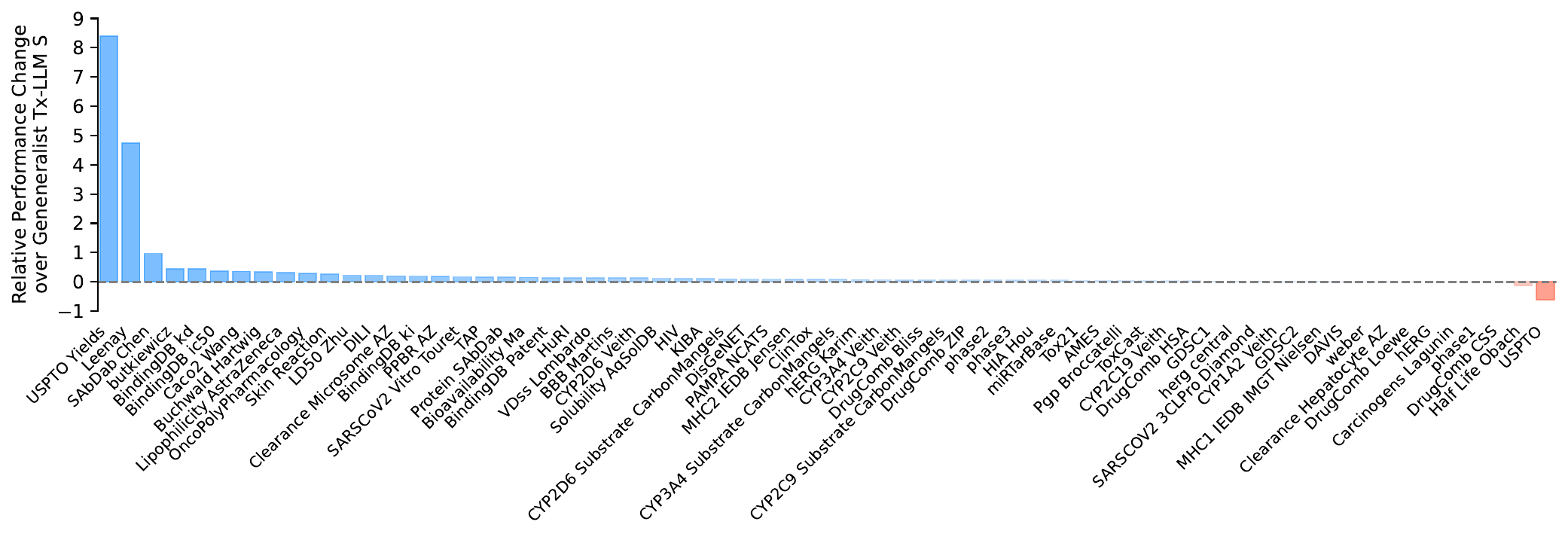}
    \includegraphics[width=1.0\textwidth]{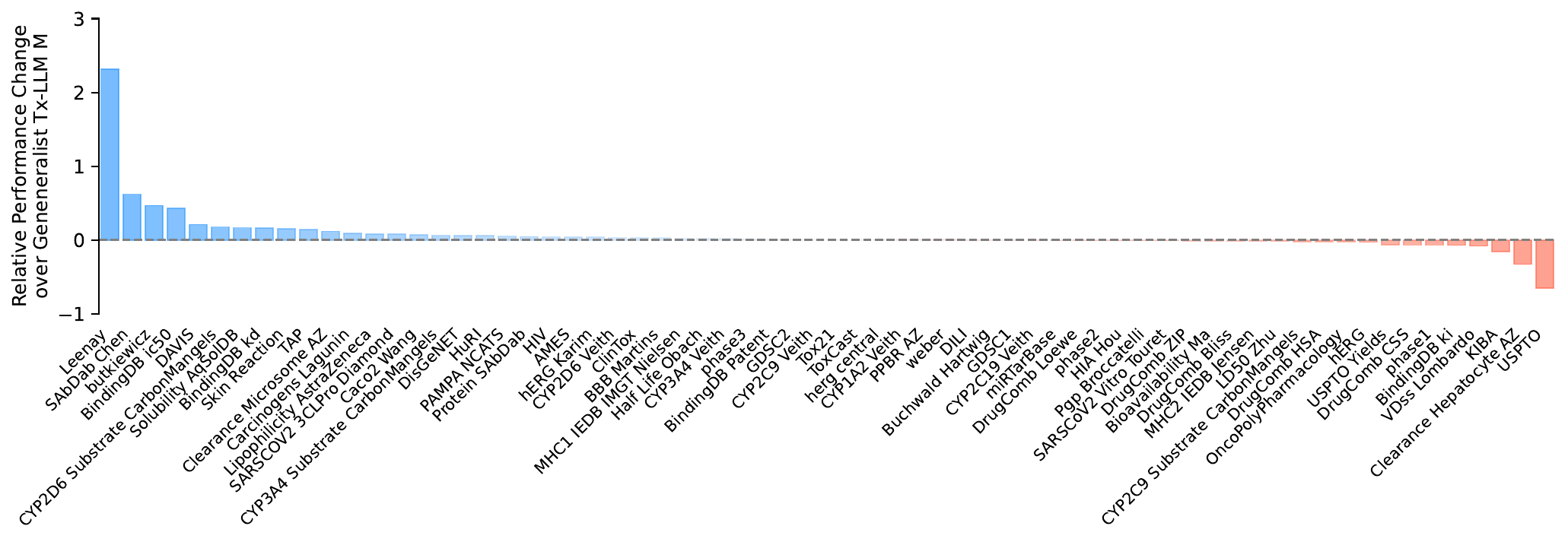}
    \vspace{0pt}
    \caption{\textbf{Comparison of \ourmodelpredict's performance with therapeutic generalist models.} \textbf{(top)} relative performance improvement of \ourmodelpredictlargest in comparison to \ouroldmodels. \ourmodelpredictlargest outperforms \ouroldmodels on 62 and underperforms on only 4. \textbf{(bottom)} relative performance improvement of \ourmodelpredictlargest in comparison to \ouroldmodelm. \ourmodelpredictlargest outperforms \ouroldmodelm on 45 out of 66 tasks, while underperforming on 21. When aggregating performance over task, we observe a net improvement of \ourmodelpredictlargest over \ouroldmodel models, with a statistically significant difference (p=0.003, Wilcoxon signed-rank test). These results establish \ourmodelpredictlargest as a competitive and functionally enhanced alternative at practical model sizes. Values for each task can be found in \cref{tab:binary_results_chat_and_txllm,tab:regression_generation_chat_and_txllm}.}
    \label{fig:generalist-specialist-relative-all}
\end{figure}

\subsubsection{Comparison with best-in-class therapeutic models}\label{sec:eval_protocol}

To provide a comprehensive evaluation of our models' predictive capabilities, we benchmark against both specialist and generalist baselines. For specialist comparisons, we define best-in-class performance metrics for each task using previous models. Specifically, we utilize TDC leaderboard scores for tasks where available (ADMET, DrugCombo, DTI DG). For remaining tasks, values are reported from a literature review and are detailed in \cref{tab:binary_results,tab:regression_generation_results}. These specialist performance values align with those reported in~\citet{chaves2024txllm}.  Additionally, we compare our models against three prominent therapeutic generalist and multi-task models: \ouroldmodel~\cite{chaves2024txllm}, LlaSMol~\cite{yu2024llasmol}, and MolE~\cite{mendez2024mole}. \ouroldmodel, with its two size-variants S and M, shares similar training data to our approach enabling a direct comparison across all tasks. LlaSMol a suite of generalist models built upon fine-tuned open-source LLMs trained for small-molecule applications~\cite{yu2024llasmol}. Similarly, MolE was developed as a graph-based multi-task foundation model for small molecules. LlaSMol and MolE, specialized for small molecules, offer strong baselines for small molecule tasks.

\textbf{\ourmodel shows improved performance compared to therapeutic generalist models} 
In \cref{fig:generalist-specialist-relative-all}, we compare the performance of \ourmodelpredictlargest to the two existing models in the \ouroldmodel~\cite{chaves2024txllm} family, \ouroldmodelm and \ouroldmodels, built over \oldbasemodel on TDC tasks. \ourmodelpredictlargest surpasses \ouroldmodelm on 45 out of 66 tasks, while underperforming on 21. In addition, it outperforms \ouroldmodels on 62 and underperforms \ouroldmodels on only 4. Aggregating performance over task, we observe a statistically significant improvement of \ourmodelpredictlargest over \ouroldmodel models (p=0.003, Wilcoxon signed-rank test). These results demonstrate that \ourmodel provides a highly competitive alternative to its predecessor with improved functionality at a substantially reduced model size.

\begin{figure}[!t]
    \centering
    \includegraphics[width=0.9\textwidth]{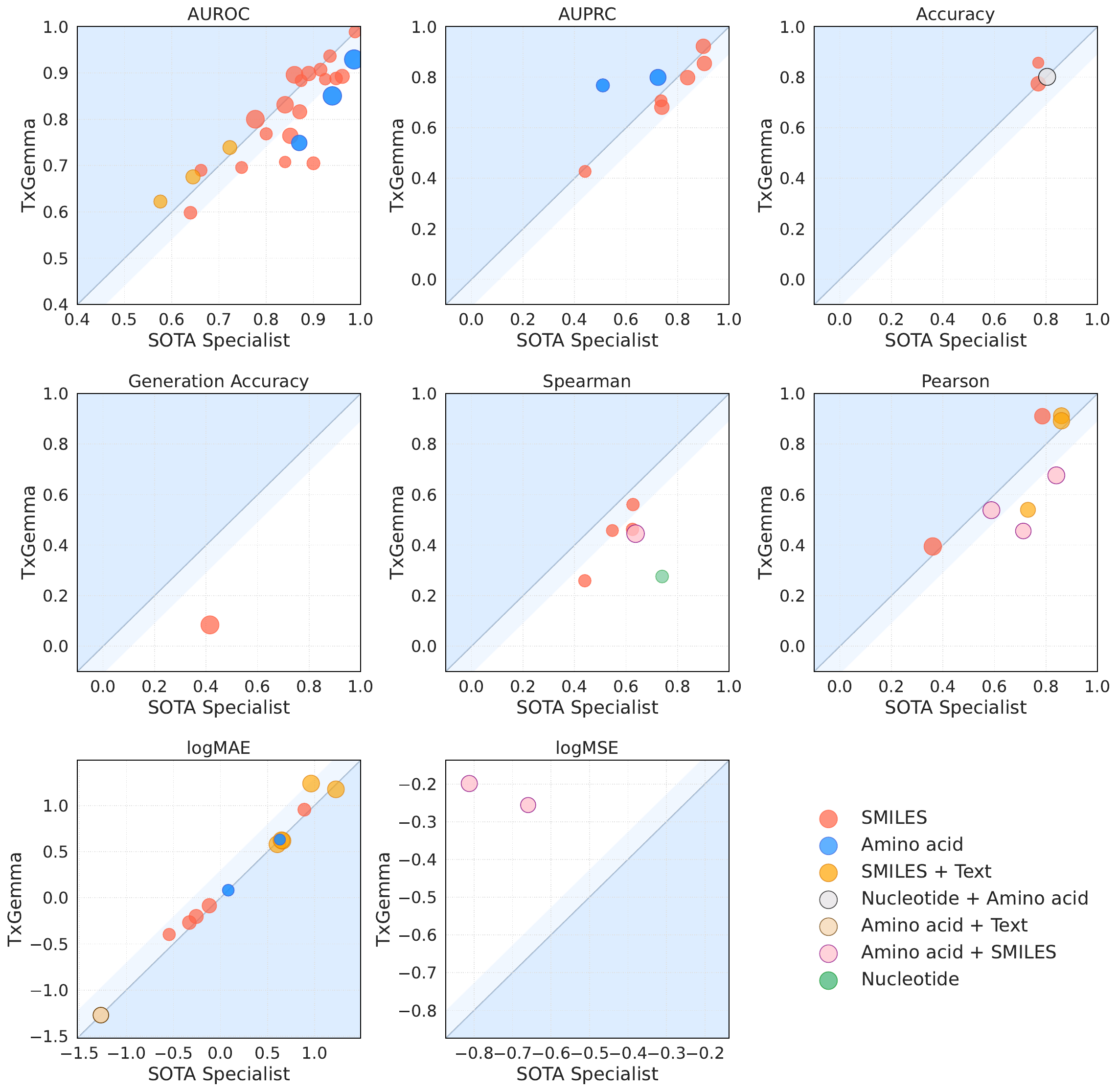}
    \vspace{6pt}
    \caption{\textbf{Comparison of \ourmodel's performance with best-in-class specialist models.} \ourmodelpredictlargest is evaluated on each task in TDC and compared to the corresponding best-in-class competitor. The panels depict different metrics used to evaluate the tasks. Tasks are colored by their feature types including one or a combination of SMILE, Amino acid, Nucleotide and text as indicated in the legend. Marker sizes illustrate the number of data points in the task on a log scale. The larger shaded area in blue indicates where \ourmodel outperforms best-in-class models, while the narrower light blue shaded area indicates where \ourmodel is performing near best-in-class model (defined as within 10\%). MAE and MSE values are log-transformed since the magnitudes of these values depend on the units of outputs. Generation accuracy is the fraction of correct SMILES strings in the USPTO generation task. Values for each task can also be found in \cref{tab:binary_results,tab:regression_generation_results}.} 
    \label{fig:comparison-specialist-scatter}
\end{figure}

\begin{table}[t]
\centering
\small
\caption{\small{\textbf{Comparative performance of \ourmodel and MolE on small molecule tasks.} Details of the predictive performance of \ourmodelpredictlargest and MolE, a graph-based molecular multi-task foundation model, across various pharmacokinetics and toxicity tasks. Bold values indicate the best performance for each task. Metrics for MolE are reported from \citet{mendez2024mole}. \ourmodelpredictlargest values are bootstrapped averages and 95\% CIs. These pharmacokinetics and toxicity tasks are publicly available in TDC~\cite{huang2021therapeutics}.}}
\label{tab:mole-comparison}
\renewcommand{\arraystretch}{1.05}
\centerline{
\begin{tabular}{l|l|c|cc}
\toprule
Task Type & Task & Metric & MolE~\cite{mendez2024mole} & \ourmodelpredictlargest \\ \midrule
\multirow{17}{*}{Pharmacokinetics} & Caco2 Wang                & MAE ($\downarrow$)   & \textbf{0.329} & 0.401 (0.358-0.449) \\
                             & Lipophilicity AstraZeneca & MAE ($\downarrow$)    & \textbf{0.406} & 0.538 (0.507-0.570)  \\
                             & Solubility AqSolDB        & MAE ($\downarrow$)   & \textbf{0.776} & 0.907 (0.870-0.948) \\
                             & PPBR AZ                   & MAE ($\downarrow$)   & \textbf{7.229} & 9.048 (8.141-10.111) \\
                             & HIA Hou                   & AUROC ($\uparrow$) & 0.984          & \textbf{0.988} (0.972-0.999) \\
                             & Pgp Broccatelli           & AUROC ($\uparrow$) & 0.930          & \textbf{0.937} (0.904-0.964) \\
                             & Bioavailability Ma        & AUROC ($\uparrow$) & 0.640          & \textbf{0.694} (0.575-0.801) \\
                             & BBB Martins                & AUROC ($\uparrow$) & 0.903          & \textbf{0.908} (0.872-0.938) \\
                             & CYP3A4 Substrate CarbonMangels & AUROC ($\uparrow$) & \textbf{0.692} & 0.691 (0.601-0.784) \\
                             & CYP2D6 Veith              & AUPRC ($\uparrow$) & 0.679          & \textbf{0.683} (0.639-0.726) \\
                             & CYP3A4 Veith              & AUPRC ($\uparrow$) & \textbf{0.876} & 0.854 (0.836-0.872) \\
                             & CYP2C9 Veith              & AUPRC ($\uparrow$) & 0.782          & \textbf{0.798} (0.767-0.826) \\
                             & CYP2D6 Substrate CarbonMangels & AUPRC ($\uparrow$) & 0.692          & \textbf{0.711} (0.570-0.830) \\
                             & CYP2C9 Substrate CarbonMangels & AUPRC ($\uparrow$) & 0.409          & \textbf{0.438} (0.302-0.576) \\
                             & VDss Lombardo             & Spearman ($\uparrow$) & \textbf{0.644} & 0.559 (0.457-0.655) \\
                             & Half Life Obach          & Spearman ($\uparrow$) & \textbf{0.578} & 0.458 (0.306-0.594) \\
                             & Clearance Microsome AZ   & Spearman ($\uparrow$) & \textbf{0.632} & 0.462 (0.353-0.565) \\
                             & Clearance Hepatocyte AZ  & Spearman ($\uparrow$) & \textbf{0.456} & 0.260 (0.129-0.384) \\ \midrule
\multirow{3}{*}{Toxicity}    & LD50 Zhu                  & MAE ($\downarrow$)   & \textbf{0.602} & 0.627 (0.597-0.660) \\
                             & hERG                      & AUROC ($\uparrow$) & 0.835          & \textbf{0.885} (0.813-0.946) \\
                             & AMES                      & AUROC ($\uparrow$) & \textbf{0.834} & 0.816 (0.795-0.838) \\
                             & DILI                      & AUROC ($\uparrow$) & 0.852          & \textbf{0.886} (0.810-0.947) \\

\bottomrule
\end{tabular}}
\end{table}

\begin{table}[!ht]
\centering
\small
\caption{\small{\textbf{Comparative performance of \ourmodel and LlaSMol on small molecule tasks.} Comparison of \ourmodelpredictlargest with LlaSMol$_{\text{Mistral}}$ (best LlaSMol model at 7B) across shared small-molecule tasks. Bold values indicate the best performance for each task. Metrics for LlaSMol$_{\text{Mistral}}$ are reported from \citet{yu2024llasmol}. \ourmodelpredict values are bootstrapped averages and 95\% CIs. These pharmacokinetics, toxicity, and high-throughput screening data and tasks are publicly available in TDC~\cite{huang2021therapeutics}}}
\label{tab:llasmole-comparison}
\renewcommand{\arraystretch}{1.05}
\centerline{
\begin{tabular}{@{\hspace{0em}}l@{\hspace{0.2em}}|@{\hspace{0.2em}}c@{\hspace{0.2em}}|@{\hspace{0.2em}}c@{\hspace{0.2em}}|c@{\hspace{0.5em}}c@{\hspace{0.5em}}c@{\hspace{0em}}}
\toprule
Task Type & Task & Metric & LlaSMol$_{\text{Mistral}}$~\cite{yu2024llasmol} &  \ourmodelpredictlargest & \ourmodelpredictnine \\ \midrule
\multirow{3}{*}{Pharmacokinetics} & BBBP$^\dagger$ & Accuracy ($\uparrow$) & 0.746 & \textbf{0.869} (0.835-0.901) & 0.847 (0.813-0.881) \\
                          & ESOL$^\dagger$ & RMSE ($\downarrow$) & \textbf{1.150} & 1.250 (1.185-1.321) & 1.360 (1.246-1.480) \\
                          & Lipo$^\dagger$ & RMSE ($\downarrow$) & 1.010 & \textbf{0.710} (0.668-0.752) & 0.742 (0.700-0.787) \\ \midrule
Toxicity & Clintox & Accuracy ($\uparrow$) & \textbf{0.931} & 0.926 (0.896-0.956) & 0.925 (0.892-0.953) \\ \midrule
High-throughput screening & HIV$^*$ & Accuracy ($\uparrow$) & 0.967 & \textbf{0.968} (0.964-0.972) & 0.965 (0.961-0.969) \\
\bottomrule
\end{tabular}}
{\raggedright
\vspace{0.1in}
\scriptsize{
$*$ To predict whether compounds have anti-HIV properties. \\ 
$\dagger$ Task name is modified to match the nomenclature from \citet{yu2024llasmol}. \\
}}
\end{table}

\textbf{\ourmodel is competitive with specialist therapeutic models }
\cref{fig:comparison-specialist-scatter} and \cref{fig-sup:comparison-reletive-mix} compare \ourmodel's performance with best-in-class specialist model across tasks containing various combinations of SMILES, amino acid, nucleotide, and text inputs. In a comparison with specialist best-in-class models, \ourmodelpredictlargest outperforms the state-of-the-art (SOTA) on \nbetterthansota and performs near SOTA on \natleastnearsota. This is a substantial improvement over its predecessor \ouroldmodelm, which outperformed SOTA on 22 tasks and near SOTA on 43. These results demonstrate the improved capabilities of \ourmodelpredictlargest and its competitiveness with current specialist models designed for specific tasks and therapeutic feature types. 

\textbf{\ourmodel performs similarly to multi-task models specialized for small molecules} 
\cref{tab:mole-comparison,fig-sup:mole} compare the predictive performance of \ourmodelpredictlargest with MolE, a graph-based multi-task foundation model for small molecules. MolE performs within the 95\% CIs of \ourmodelpredictlargest for 15 out of 22 tasks. Furthermore, both \ourmodelpredictlargest and \ourmodelpredictnine outperform LlaSMol$_{\text{Mistral}}$ (7B), the top performing model from the LlaSMol suite, on 2 of 5 shared tasks and within 95\% CIs on 2 additional tasks (\cref{tab:llasmole-comparison,fig-sup:llasmol}). All metrics for MolE and LlaSMol are reported from \citet{mendez2024mole,yu2024llasmol}. Given their specialization in small-molecule tasks, LlaSMol and MolE provide strong baselines for evaluating generalist models. Notably, \ourmodel, a generalist model encompassing diverse drug types and many different tasks, achieves competitive performance with these dedicated models designed for a narrower range of small-molecule tasks.


\subsection{\ourmodel Conversational Capabilities}
\label{conversational}
While \ourmodelpredictlargest performs well on prediction tasks, training solely on instruction tuning data for therapeutic properties limits its conversational capacity. \ourmodelpredictlargest can engage in general conversation, but its performance deteriorates when prompts deviate from the expected format. \cref{fig-sup:27b-predict-conv-failure} shows an example of such decline in \ourmodelpredictlargest's conversational capabilities. To expand the \ourmodel family's capabilities and provide a more versatile tool with the ability to explain its reasoning, we trained \ourmodelconverse with a mix of therapeutic and general instruction-tuning data as detailed in \cref{subsec:modeling}. We evaluate these new conversational capabilities through a combination of standard LLM benchmarks and qualitative examples. We also run our models through assurance evaluations, as done for Gemma-3 \cite{team2025gemma3}, to verify that \ourmodel models adhere to safety policies.

\textbf{\ourmodelconverse bridges the gap between property predictors and general language models} To assess the performance of \ourmodelconverse as a general conversational LLM, we evaluated it on the Massive Multitask Language Understanding (MMLU)~\cite{hendrycks2020measuring} benchmark, a comprehensive suite of 57 diverse tasks spanning mathematics, history, computer science, law, \etc. This benchmark evaluates knowledge, reasoning, and problem-solving abilities across a wide range of academic subjects, providing a measure of overall language understanding. It comprises 14,079 multiple-choice questions, each with four possible answers.  For this multiple-choice format, we took the model's prediction as the option with the highest log-likelihood in a zero-shot setting and report overall accuracy as well as per-subject accuracy.

\cref{fig-sup:comparison-mmlu} compares the performance of \ourmodelconverselargest, \ourmodelpredictlargest, and \basemodellargest on MMLU, a standard benchmark for evaluating general LLMs. \ourmodelconverselargest achieves an accuracy of 73.87\%, slightly lower than \basemodellargest's 75.38\%, but \ourmodelconverselargest shows slight improvements in areas such as medical genetics, high school statistics, and college chemistry. Furthermore, \ourmodelconverselargest significantly outperforms \ourmodelpredictlargest, which has an accuracy of 53.60\%. This suggests that while fine-tuning solely on therapeutic data can diminish general knowledge acquired during pre-training, incorporating general instruction-tuning data can mitigate this effect.

\begin{figure}[!t]
    \centering
    \includegraphics[width=1.0\textwidth]{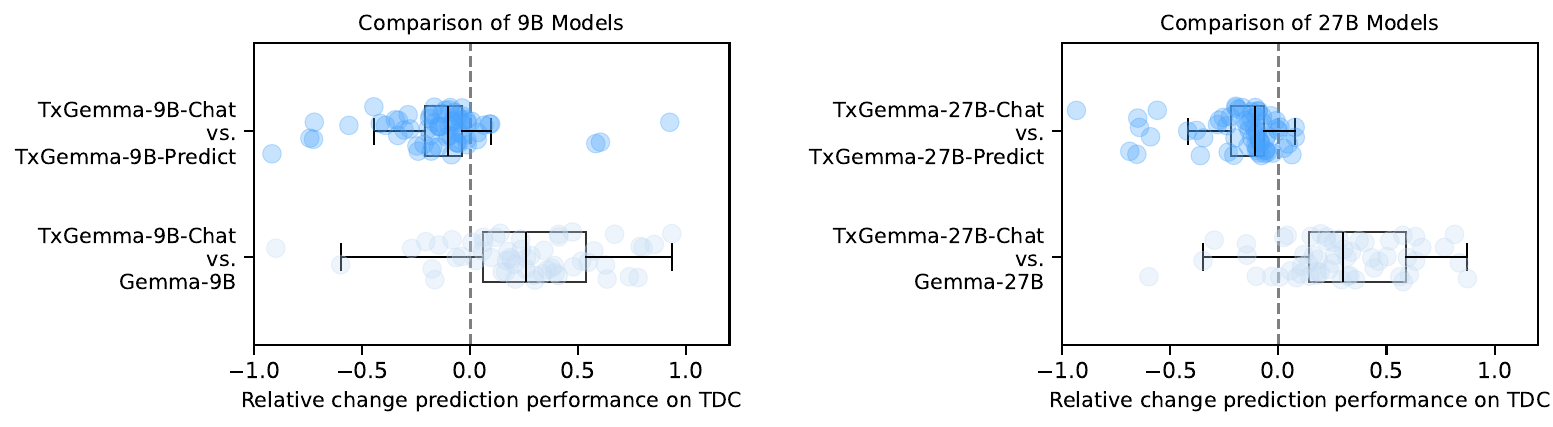}
    \vspace{-6pt}
    \caption{\small{\textbf{\ourmodelconverse bridges the gap between property predictors and general LLMs.}  Each point represents a therapeutic task in the TDC. The figure depicts relative predictive performance changes of \ourmodelconverse in comparison to \ourmodelpredict (top) and \basemodel (bottom) for 9B variants \textbf{left} and 27B variants in \textbf{right}. As expected, \ourmodelpredictlargest outperforms \ourmodelconverselargest on therapeutic tasks, with \ourmodelconverselargest showing a 10.69\% median relative performance reduction. However, \ourmodelconverselargest exceeds the \basemodellargest baseline by 29.67\% on TDC therapeutic tasks. Similarly, \ourmodelconversenine's performance is 10.32\% lower than \ourmodelpredictnine's.  Values for each task can be found in \cref{tab:binary_results_chat_and_txllm,tab:regression_generation_chat_and_txllm}.}}
    \label{fig:chat-base-relative}
\end{figure}

Furthermore, we assess \ourmodelconverselargest on all therapeutic tasks within TDC. \cref{fig:chat-base-relative} compares the relative performance changes of \ourmodelconverselargest to \ourmodelpredictlargest and \basemodellargest for both 9B and 27B variants across these tasks. As anticipated, \ourmodelpredictlargest outperforms \ourmodelconverselargest on these predictive tasks, with a median relative performance reduction of 11\% observed for \ourmodelconverselargest. Nevertheless, \ourmodelconverselargest surpasses the baseline \basemodellargest, demonstrating a median relative improvement of 30\%. Similarly, \ourmodelconversenine shows a 10\% median relative performance reduction compared to \ourmodelpredictnine. Regression tasks experience the greatest performance decline from the general-purpose training. These results demonstrate how \ourmodelconverse bridges the gap between therapeutic property predictors and general LLMs, functioning as a unified model for both capabilities.

\begin{figure}[!htp]
    \centering
    \includegraphics[width=0.98\textwidth]{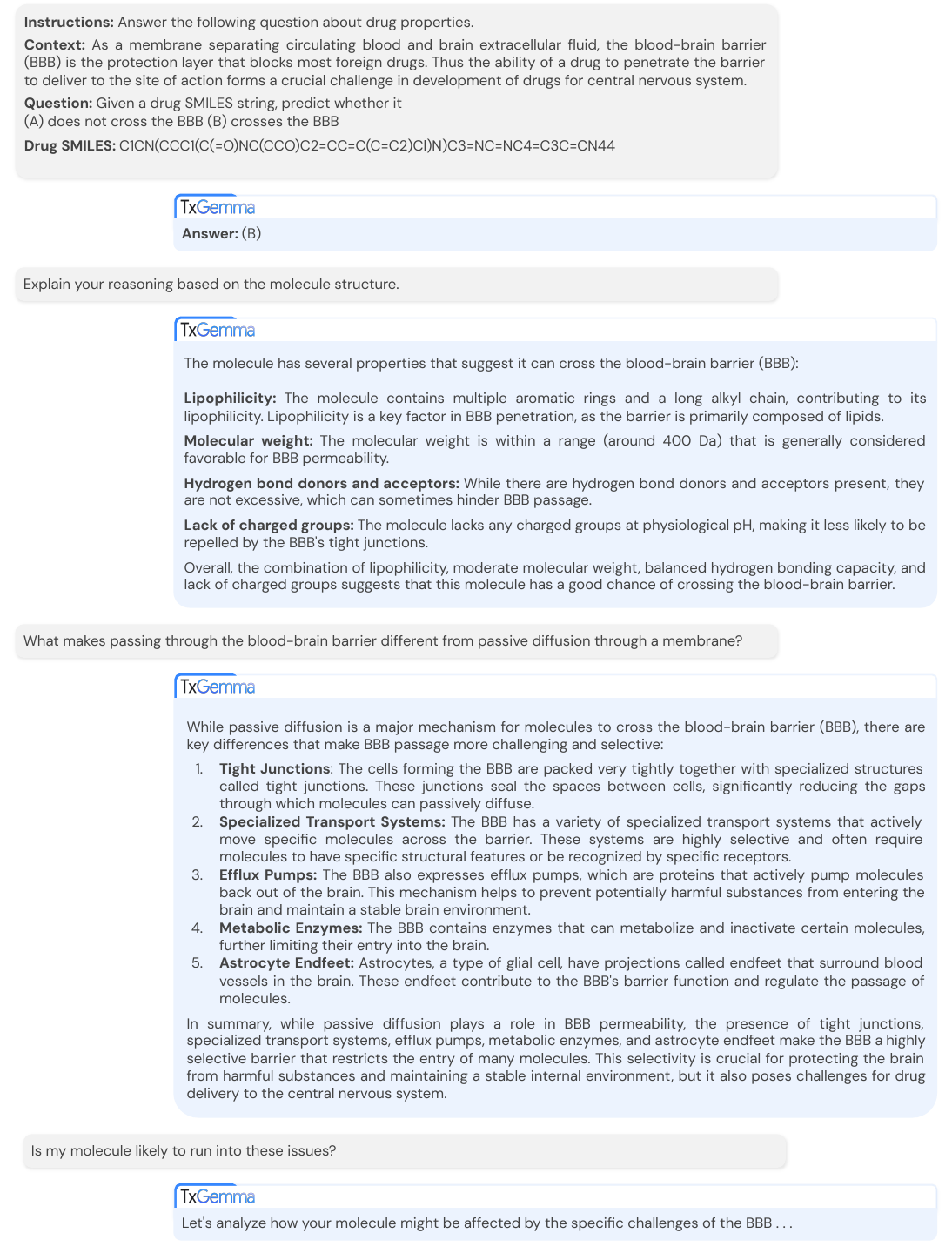}
    \caption{\textbf{\ourmodelconverse can be prompted to explain its reasoning or provide additional context.} Following an initial question, \ourmodelconverselargest is asked to explain its reasoning based on molecule structure in the second turn. The model uses its understanding of chemistry and biology to justify its answer and can continually engage with the user on follow-up questions.}
    \label{fig:txgemma-long-convo}
\end{figure}

\textbf{\ourmodelconverse can provide reasoning for complex tasks.} A particularly compelling application of conversational models lies in prompting them to explain their predictions to users. While general LLMs may possess some foundational knowledge concerning therapeutic challenges, they are not accurate for property prediction (\cref{fig:chat-base-relative}). In \cref{fig:txgemma-long-convo}, we prompt \ourmodelconverselargest to answer a question regarding blood-brain barrier permeability using the BBB Martins prompt format. \ourmodelconverselargest provides only the answer in the initial turn, but when given a subsequent prompt to articulate its rationale, the model provides mechanistic reasoning for its answer based on molecular solubility and the structure of the input molecule derived from the SMILES string. All of this reasoning occurred directly within the model weights, without requiring any preprocessing of the SMILES string.

Interestingly, prompting structures enable \ourmodelconverse to provide additional reasoning on complex tasks. For instance, while the relationship between blood-brain barrier permeability and lipophilicity is intuitive, some tasks such as predicting clinical trial approval are more challenging to reason over. If \ourmodelconverselargest is prompted to provide reasoning in the same manner as in \cref{fig:txgemma-long-convo} for predicting clinical trial approval, \ourmodelconverselargest refuses and directs the user to alternative sources. However, when modifying the original prompt, instructing the model to output reasoning steps before the final answer, it bypasses the refusal and restores reasoning capabilities (\cref{fig-sup:txgemma-convo-trial}).

\subsection{Agentic Planning and Execution based on \ourmodel}
\textbf{\ouragent demonstrates competitive performance on therapeutic benchmarks}. We evaluate the capability of \ouragent to assist with therapeutics tasks by means of questions from three benchmarks: GPQA (Diamond)~\cite{rein2024gpqa}, ChemBench~\cite{mirza2024large}, and Humanity's Last Exam (HLE)~\cite{phan2025humanity}. Within each benchmark, we use existing selections of therapeutic-relevant questions; for GPQA we evaluate GPQA-Chemistry (47 questions), for ChemBench we evaluate ChemBench-Chemical Preference which aims to select an ideal candidate molecule for therapeutic development (1,001 question) and ChemBench-mini, which evaluates across 8 categories of chemistry from toxicity/safety to organic chemistry (236 questions).
Finally, for HLE, we evaluate HLE-Chemistry and HLE-Biology (235 questions). For open-ended questions in HLE, we observed a high variation of metric scores depending on the selection of the LLM-rater model~\cite{phan2025humanity}. To ensure an objective accuracy measure, we restrict the evaluation to  multiple choice questions (MCQs).

\begin{table}[t]
\centering
\small
\caption{\small{\textbf{Performance of \ouragent.} Accuracy of \ouragent compared with SOTA models on ChemBench, GPQA, and HLE benchmarks.}}
\label{tab:tx-agent-perf}
\renewcommand{\arraystretch}{1.}
\centerline{
\begin{tabular}{@{\hspace{0em}}l@{\hspace{0.2em}}|@{\hspace{0.2em}}c@{\hspace{0.2em}}|@{\hspace{0.2em}}c@{\hspace{0.2em}}|c@{\hspace{0.2em}}|c@{\hspace{0em}}}
\toprule
\multirow{2}{*}{\textbf{Model}} & \multicolumn{2}{c|}{ChemBench}                   & GPQA (Diamond)  & Humanity's Last Exam \\\cmidrule{2-5}
                                & \hspace{0.3cm} Mini \hspace{0.3cm} & Preference  & \hspace{0.3cm} Chemistry  \hspace{0.3cm}  & Chemistry \& Biology  \\
\midrule
\ouragent (Gemini 2.5-Pro) \hspace{0.2cm} & \textbf{84.5} & \textbf{66.2} & \textbf{81.7} & \textbf{20.1} \\
\ouragent (Gemini 2.0-Pro) \hspace{0.2cm} & 83.4          & 65.5          & 62.4          & 14.5 \\
\ouragent (Gemini 1.5-Pro) \hspace{0.2cm} & 80.6         & 65.0           & 51.8          & 11.9 \\
\midrule
Claude-3.5 (Sonnet) & ~73.0$^{*}$  & ~~60.0$^{*\dagger}$ &~~40.4~~                & -     \\
GPT-4o              & ~72.0$^{*}$  & ~~59.0$^{*}$~       & ~~43.8$^{**}$          & 3.8   \\
Gemini 2.5-pro      & ~82.8~       & ~~65.5~~            & ~~79.5~~               & 17.9  \\
Gemini 2.0-pro      & ~79.6~       & ~~58.4~~            & ~~53.3~~               & 11.1  \\
Gemini 1.5-pro      & ~74.9~       & ~~55.6~~            & ~~48.2~~               & 10.6  \\
PaperQA2~\cite{skarlinski2024language} 
                    & ~67.0$^{*}$  & ~~56.0$^{*}$~       & -                      & -     \\ 
o1                  & ~80.0$^{*}$  & ~~56.0$^{*}$~       & ~~64.7$^{**}$          & 12.3  \\
o3-mini (medium)    & ~82.4~       & ~~61.3~~            & ~~62.5~~               & 13.0  \\ 
o3-mini (high)      & ~82.5~       & ~~62.0~~            & ~~64.5~~               & 13.2  \\ \midrule
Human Expert (Average Performance) & ~27.0~              & -                      & - & - \\ 
\bottomrule
\end{tabular}}
{\raggedright
\vspace{0.1in}
\footnotesize{
\hspace{0.3in} ($\dagger$) Using ReAct framework, ($^*$) Extracted from ~\cite{mirza2024large}, ($^{**}$) Extracted from ~\cite{OpenAIReasoningLLMs}\\
}}
\end{table}

As shown in Table \ref{tab:tx-agent-perf}, \ouragent (Gemini 2.5-Pro), \ouragent (Gemini 2.0-Pro), and \ouragent (Gemini 1.5-Pro) achieve competitive or greater accuracy compared to existing SOTA models across several benchmarks. Specifically, \ouragent (Gemini 2.5-Pro) and \ouragent (Gemini 2.0-Pro) surpasses prior SOTA models on the exceptionally difficult Humanity's Last Exam benchmark (Chemistry \& Biology tasks), with \ouragent (Gemini 2.5-Pro) achieving 52.3\% relative improvement over o3-mini (high) and 13.4\% over Gemini 2.5-pro, as well as on ChemBench, with relative improvements of 6.3\% (ChemBench-Preference) and 2.4\% (ChemBench-Mini) over o3-mini (high) and 1.1\% (ChemBench-Preference) and 2.0\% (ChemBench-Mini) over Gemini 2.5-pro. On GPQA (Diamond), \ouragent also achieves SOTA accuracy with 26.7\% relative improvements over o3-mini and 2.7\% over Gemini 2.5-pro. All variants of \ouragent outperform their corresponding base Gemini models across all benchmarks, indicating the effectiveness of the \ouragent framework in enhancing LLMs efficacy for advanced reasoning within this domain. This suggest that agentic workflows such as ours represent useful tools for therapeutic development, particularly in areas requiring domain knowledge and the selection of candidate molecules. The agent's ability to leverage external tools and perform multi-step reasoning enables it to address more complex queries beyond the scope of traditional LLMs. 

\textbf{\ouragent effectively leverages various tools based on the therapeutic task requirement.} In \cref{fig-sup:tool-use-frequency}, we investigate tool usage frequency within the \ouragent system across the ChemBench-Preference and Biology and Chemistry (B\&C) HLE datasets. Our analysis reveals that \ouragent tool usage distribution varies significantly depending on the task and dataset. For the ChemBench-Preference task, which focuses on selecting ideal candidate molecules for therapeutic development, the \ouragent system exhibits a high frequency of usage for tools such as SMILES description and toxicity prediction. This suggests a strong emphasis on molecular characterization and safety assessment in this task correctly invoked by \ouragent. In contrast, on the B\&C HLE dataset, tool usage is predominantly concentrated on general knowledge retrieval tools like PubMed or Wikipedia search. This indicates that the \ouragent system relies heavily on accessing and synthesizing broad biological or chemical knowledge to address questions in these domains. In \cref{fig-sup:tool-use-per-question}, we investigate the breakdown of tool interactions per question and explore how these interactions contribute to performance variations. Our analysis shows that each question can involve up to 8 tool calls, and the high usage of tools such as SMILES description and toxicity prediction tools correlates with overall performance improvement. These results highlight the \ouragent system's adaptive nature, demonstrating its ability to leverage different tools based on the specific requirements of the task.

\paragraph{\ouragent inference time is suitable for real time human interaction} Analysis of \ouragent's inference time indicates efficient performance characteristics. The median time observed for tool execution is 0.55 seconds. The fastest tool (Gene Sequence) completes execution in 0.15 seconds, while the slowest (ToxCast) requires 28.2 seconds. This suggests that \ouragent operates within a timeframe conducive to real-time user interaction. The observed latencies demonstrate suitability for integration into workflows where immediate feedback and responsiveness are desired. The system's ability to maintain a median inference time below one second contributes to an efficient user experience.

\subsection{Additional Analysis and Ablations}

\textbf{Data contamination analysis and data leakage considerations} To assess potential data contamination from the \basemodel pretraining data, we calculated the overlap between features in the therapeutic instruction-tuning data and the pretraining corpus. For multi-instance tasks, contamination was defined as the presence of \textit{any} constituent feature (e.g., drug SMILES or target protein sequence in drug-target binding) in the pretraining data. The majority of tasks showed no direct contamination (\cref{fig:contamination-percent-decoy}). For tasks with some contamination, filtering contaminated datapoints and recalculating \ourmodelpredictlargest performance revealed no significant changes (\cref{fig:contamination-performance}).

While direct contamination was minimal, we further investigated potential \textit{indirect} contamination. Although SMILES strings are less common in general web text, pretraining on molecular \textit{names} could have created learned associations between names and SMILES, potentially influencing test set performance. To test this, we compared the similarity of \ourmodelpredictlargest embeddings for PubChem molecules represented as SMILES strings and their corresponding IUPAC names, against the similarity of embeddings for SMILES strings paired with decoy (randomly selected, incorrect) names. The similarities were statistically equivalent (\cref{fig:contamination-percent-decoy}), confirmed by a two one-sided t-test ($p=3 \times 10^{-12}$, $\delta = 0.02$). This suggests that \ourmodelpredictlargest did not learn spurious name-SMILES associations during pretraining, likely because names and SMILES were encountered in separate training phases and for different molecules. Therefore, both direct and indirect contamination from pretraining are unlikely to significantly affect our results.

\textbf{Fine-tuning \ourmodel models improves data efficiency} Given the scarcity of therapeutic data and the potential of \ourmodel to serve as a pretrained model for further adaptation, we investigated \ourmodel's data efficiency and generalization to new tasks in out-of-distribution settings. Specifically, we fine-tuned the baseline model \basemodellargest as well as our \ourmodelpredictlargest on adverse event prediction data from TrialBench~\cite{chen2024trialbench}. Serious adverse events are critical in assessing the safety profile of a new treatment and accurate prediction of these events allows for better risk management and resource allocation~\cite{chen2024trialbench}. To ensure a fair evaluation of generalization, we filtered the TrialBench test set to exclude samples overlapping with phase 1, 2, or 3 of clinical trial outcome prediction data in TDC. In addition, datapoints without available SMILES strings are excluded. This lead to 14,368 train and 3,184 test samples.

\begin{figure}
    \centering
    \includegraphics[width=\textwidth]{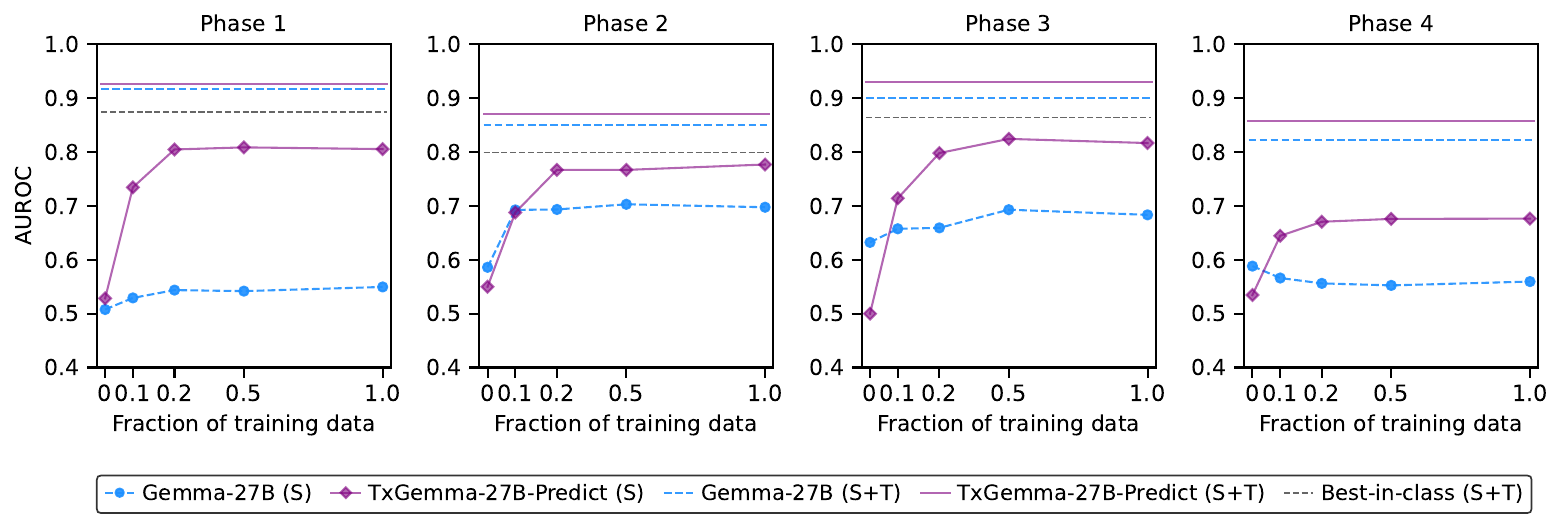}
    \vspace{-6pt}
    \caption{\textbf{\ourmodel improves efficiency at adverse event prediction from SMILES strings.} The figure shows the AUROC of predicting adverse events in a clinical trial from the drug SMILES strings as a function of the training data fraction for \basemodellargest and \ourmodelpredictlargest. Clinical trials are separated based on trial phase, and datapoints without available SMILES strings are excluded. To assess model performance with additional textual information, separate models trained on both SMILES strings and additional textual information are indicated by colored dashed lines, and SOTA models are indicated by gray dashed lines. (S) denotes models trained with SMILES strings only, and (S+T) those trained with SMILES and textual information (\cref{tab:example_prompts_adverse}).}
    \label{fig:adverse_finetuning}
\end{figure}

We consider two settings. Initially, we focus exclusively on drug SMILES strings as the only feature contributing to clinical trial outcome, thereby isolating the influence of therapeutic information by excluding this additional context. To simulate data limitations, we fine-tuned \ourmodelpredictlargest and the baseline \basemodellargest on varying fractions of the training data, and then evaluated the newly fine-tuned models performance on the test set after 30 epochs of training (\cref{fig:adverse_finetuning}). Overall, \ourmodelpredictlargest achieved higher AUROCs with lower amounts of training data, matching the performance of \basemodellargest with less than 10\% of retraining data. In the second setting, we explored the performance ceiling by incorporating textual information about the clinical trials, increasing the number of tokens provided to the models by a factor of 4 (\cref{tab:example_prompts_adverse}). This is the setting used by the best-in-class model for adverse event prediction \cite{chen2024trialbench}. The addition of textual information allowed our models to consistently outperform existing SOTA methods \cite{chen2024trialbench}. However, the performance difference between \ourmodelpredictlargest and \basemodellargest decreased in this scenario because the additional textual information diluted the relative importance of the drug SMILES strings.

\textbf{\ourmodel inference time is suitable for virtual screening} In \cref{fig:speedtest}, we plot the inference speed of \ourmodel models of all sizes normalized by the number of TPUv5e chips used for serving. All model sizes are suitably fast for virtual screening, as even the largest 27B model is able to inference around 9,000 samples per day per TPU chip. Using 64 chips for serving, this would yield around 600,000 samples per day for the 27B model, and the smallest 2B model would reach 3,000,000 samples per day.

\textbf{Correlation between clinical trial approval and toxicity predictions} We investigated the correlation between \ourmodel's clinical trial approval predictions (based on SMILES and target disease) and its toxicity predictions (using TDC's AMES, DILI, and hERG tasks). \cref{fig:toxtop-correlation} shows a consistent, but weak (0.15-0.35), positive Spearman correlation across all phases. This suggests \ourmodel associates lower predicted toxicity with approval, but may also consider other factors such as efficacy or drug-likeness.

\textbf{Impact of feature types} \cref{fig:txgeamma-featuretype} presents a performance breakdown of \ourmodelpredictlargest by feature type, compared to \ouroldmodelm. In both models, tasks incorporating both SMILES strings and textual features (e.g., disease names, cell line names/descriptions) show the most significant improvement over SOTA. This suggests that the contextual knowledge acquired during LLM pretraining could aid in synthesizing textual information with molecular representations.

\textbf{Model size and domain fine-tuning ablations} Figure~\ref{fig:size_ablation} compares the performance of \ourmodelpredict models across different sizes (2B, 9B, and 27B) on TDC tasks. Pairwise comparisons using the Wilcoxon signed-rank test indicate that model size is a significant factor: \ourmodelpredictlargest outperforms \ourmodelpredictnine ($p=0.013$) and \ourmodelpredictsmall ($p=6.2 \times 10^{-6}$), and \ourmodelpredictnine outperforms \ourmodelpredictsmall ($p=0.048$). Furthermore, comparing \ourmodel models to their corresponding base \basemodel models reveals the significant impact of domain fine-tuning. All \ourmodel models significantly outperform their \basemodel counterparts ($p < 10^{-10}$, Wilcoxon signed-rank test), underscoring the importance of specialized training for therapeutic tasks.

\section{Related work}

\textbf{Task-specific models for chemistry and therapeutics.} In recent years, there has been a surge in the development of deep learning models designed for various chemistry applications. Amongst those, graph neural networks (GNNs) have been applied for a wide variety of molecular prediction or generation tasks because small molecules are naturally represented as graphs~\cite{torng2019graph,stark2022equibind,xiong2020pushing,heid2022machine,yang2019analyzing,morrone2020docking,mohr2022data,stokes2020deep,mendez2024mole}. Another common representation for small molecules is molecular fingerprints~\cite{rogers2010extended}, which are binary vectors that capture the local environment of each atom~\cite{torng2019graph,tayyebi2023prediction,belenahalli2023development}.

TxGNN trained a GNN on medical knowledge graphs in order to perform zero-shot drug repurposing for diseases with limited treatment options~\cite{huang2024foundation}. AlphaFold and its successors have also significantly advanced the field of protein structure prediction and protein design~\cite{jumper2021highly,tunyasuvunakool2021highly,senior2020improved,abramson2024accurate,zambaldi2024novo}. These models have been influential for both mechanistic research and the development of structure-based drugs~\cite{ren2023alphafold}.

\textbf{Large language models for biology and chemistry.} Transformer-based models \cite{vaswani2023attention} have fueled the development of LLMs, which are trained on massive textual datasets with subsequent instruction-tuning~\cite{zhang2024instruction} or alignment~\cite{kaufmann2023survey}. LLMs have demonstrated exceptional proficiency in various tasks, including text summarization, translation, and question answering~\cite{brown2020language,liu2019text,kenton2019bert}. Their ability to encode vast amounts of information and generalize to new tasks has sparked considerable interest in their potential applications across diverse domains.

There has been increasing interest in applying the development for LLMs to scientific research. BrainGPT fine-tuned LLMs on neuroscience literature and found greater performance than domain experts~\cite{luo2024large}. LlaSMol fine-tuned LLMs on small molecule datasets and achieved near-SOTA performance on multiple tasks~\cite{yu2024llasmol}. CLAMP used separate modules for natural language and molecular inputs, combining them together in a contrastive pre-training objective~\cite{seidl2023enhancing}. Protein language models~\cite{rives2019biological,lin2023evolutionary,alley2019unified,ferruz2022protgpt2} and genomic language models~\cite{nguyen2024sequence,dalla2024nucleotide,cornman2024omg} have used self-supervised pretraining to generate embeddings useful for downstream tasks. ProtLLM~\cite{zhuo2024protllm}, BioT5~\cite{pei2024biot5}, and GraphToken~\cite{anonymous2024parameter} combine molecule or proteins with LLMs using textual or multi-modal strategies. Cellular foundation models such as scGPT~\cite{cui2024scgpt}, GenePT~\cite{chen2024genept}, Geneformer~\cite{theodoris2023transfer}, Nicheformer~\cite{schaar2024nicheformer}, and Cell2Sentence~\cite{cell2sentence} represent cells based on their gene expression to differentiate cell types and understand gene perturbations. NatureLM~\cite{xia2025naturelm} trained a foundation model that represents small molecules, proteins, RNA, and materials as sequences over a wide variety of scientific tasks.

\textbf{Agentic Systems.} Unlike traditional passive models, agentic systems proactively choose actions to achieve goals~\cite{wang2024survey, shanahan2023role, qian2023communicative, hong2023metagpt, talebirad2023multi}, involving planning~\cite{hao2023reasoning, huang2022language, song2023llm, wang2023describe, yao2024tree} and interaction with external tools~\cite{parisi2022talm, schick2023toolformer, qin2024tool, cai2023large}. LLMs have enabled such systems by processing complex information and generating action-driving responses. The ReAct framework~\cite{yao2022react} combines reasoning, action, and observation, with variations incorporating self-reflection~\cite{shinn2024reflexion} or model architectures for internal tool usage~\cite{schick2023toolformer}. Agentic frameworks enable automating tasks like software development~\cite{qian2023communicative, yang2024swe, qian2023experiential, gottweis2025towards} and scientific research~\cite{schmidgall2025agent, swanson2024virtual, lu2024ai} including biomedical applications such as nanobody design~\cite{swanson2024virtual}, drug discovery~\cite{m2024augmenting}, or reaction optimization~\cite{boiko2023autonomous}. ChemCrow~\cite{m2024augmenting} is an agent designed to perform chemistry experiments in drug discovery and materials design. The coscientist by \citet{boiko2023autonomous} designs and performs chemical experiments by integrating web knowledge, code execution, and experiment automation, demonstrating successful reaction optimization of palladium-catalysed cross-couplings. The multi-agent system AI co-scientist~\cite{gottweis2025towards} is designed for hypothesis generation over a variety of scientific fields. TxAgent was developed as an agentic framework that provides multi-step reasoning and tool use aimed towards therapeutic applications, processing clinical information to support tasks like treatment recommendation \cite{gao2025txagent}. In contrast to recommending existing therapeutics, \ouragent generally focuses on developing new therapeutics.

\section{Discussion}
\label{sec:discussion}
\ourmodel's performance suggests a paradigm shift in therapeutic AI development, demonstrating the viability of generalist LLMs. Despite the established dominance of specialist models in niche areas, \ourmodel, a relatively lightweight and efficient generalist, achieves competitive results across a wide array of therapeutic tasks. This highlights the potential for broadly trained LLMs, such as those leveraging the comprehensive dataset Therapeutics Data Commons (TDC), to serve as powerful preliminary tools for hypothesis generation, information synthesis, and candidate prioritization. While specialist models would likely retain their value for complex, domain-specific challenges, future research should explore synergistic approaches that combine the strengths of both generalist and specialist therapeutic AI.

A significant advancement with \ourmodelconverse is its ability to provide reasoning for its predictions, a first in therapeutic AI and a feature lost in \ourmodelpredict, likely due to ``catastrophic forgetting''~\cite{aleixo2023catastrophic}. While explainability may introduce a small trade-off in raw predictive power, it provides a crucial window into the model's decision-making, a factor of paramount importance in therapeutic development. For instance, explaining blood-brain barrier permeability based on molecular structure provides valuable insights for medicinal chemists. Beyond its research applications, \ourmodelconverse holds a significant educational potential, enabling students and researchers to explore complex therapeutic concepts. At the same time, it is important to acknowledge that provided explanations are correlations, not necessarily causal, and must be interpreted with caution. The model's occasional inability to explain certain predictions reveals its knowledge boundaries. Future research should prioritize improving reliability and comprehensive explanations. Even with current limitations, \ourmodelconverse represents an important improvement over the ``black-box'' paradigm.

Expanding beyond single-step predictions, \ouragent demonstrates the potential for LLMs to orchestrate complex workflows. By integrating \ourmodel with a suite of external tools (PubMed, Wikipedia, chemical databases, \etc), \ouragent can tackle multi-step reasoning tasks that would be difficult for a standalone LLM.  Its strong performance on benchmarks like ChemBench Chemical Preference and Humanity's Last Exam (HLE) highlights the synergistic value of integrating domain-specific knowledge from \ourmodel with general reasoning and information retrieval. This modular, tool-based design further ensures flexibility and extensibility, allowing for future integration of new tools and data. Importantly, it solves the issue of knowledge cut-off in LLMs by providing access to up-to-date information. \ouragent with its autonomous and collaborative operation is a powerful asset for augmenting researchers and advancing therapeutic development.

The data efficiency of \ourmodel is clearly demonstrated in fine-tuning experiments on TrialBench. It achieves robust performance on novel tasks with substantially less training data compared to baseline models, showcasing the benefits of pre-training on a broad and diverse dataset like TDC. This efficiency is particularly critical in therapeutic domains, where data is often proprietary and limited. Moreover, our finding that adding textual context, while improving overall results, can dilute the influence of molecular representations emphasizes the importance of balancing the benefits of additional information with strategic feature selection.  

Although our \textit{in-silico} results across a diverse range of therapeutic tasks are highly encouraging, we acknowledge that \ourmodel's performance has not yet been validated in real-world, wet-lab experiments. Prospective validation in these settings represents a crucial next step. However, a cornerstone of this work is our commitment to open model release. By making \ourmodel readily accessible to the research community, we aim to facilitate its rigorous validation and adaptation. Researchers can tailor \ourmodel to their specific datasets, encompassing tasks and distribution shifts beyond the scope of TDC. Given the predominantly proprietary nature of therapeutic data, we believe this collaborative, community-driven approach is essential for translating \ourmodel into tangible therapeutic applications

\section{Conclusion}
\label{sec:conclusion}
In conclusion, this work introduced \ourmodel, a suite of efficient, generalist LLMs designed to improve therapeutic development. By leveraging extensive therapeutic instruction-tuning datasets and building upon the foundation of \basemodel, \ourmodel achieves exceptional performance across a wide range of predictive and generative therapeutic tasks, surpassing or matching both generalist and specialist state-of-the-art models. Notably, \ourmodel's conversational counterparts, a first in therapeutic AI, provide reasoning and explanations, moving beyond traditional black-box predictions to facilitate mechanistic understanding and scientific discourse. Furthermore, the integration of \ourmodel into an agentic system, \ouragent, demonstrates its capacity to solve complex, multi-step problems, achieving state-of-the-art results on challenging reasoning-intensive tasks. Finally, and critically, the open release of \ourmodel empowers the research community and scientist to adapt and refine the models on their own private data, potentially leading to significant advancements in drug discovery and development. Through these contributions, \ourmodel represents a meaningful step towards more efficient, transparent, and collaborative AI-driven therapeutic research.

\vspace{15pt}
\subsubsection*{Acknowledgments}
This project was a collaboration between teams at Google DeepMind and Google Research. We thank Marcus~Brubaker, David Belanger, Justin Chen, and David Steiner  for the feedback and insight which significantly contributed to the enhancement of this report. We thank Tris Warkentin, Glenn Cameron, Victor~Cotruta, Fereshteh Mahvar, Tiffany Chen, Omar Sansevier, Kathleen Kenealy, Joe Fernandez, Gus~Martins, Nabila~Babar, Sara Smoot, Antonia Paterson, Pankil Botadra, Metin Toksoz-Exley, Tim Thelin, Can~``John''~Kirmizi, and Fayaz Jamil for their collaborative efforts in enabling the open model launch of \ourmodel. We also thank Phoebe Kirk, Rachelle~Sico, Yun Liu, Anand Rao, Jon Small, Juanita Bawagan, Jane Park, Jenn~Sturgeon, Fred Alcober, Samantha~Heyman, Abhinav Das for their valuable insights and technical support. We are also grateful to Zoubin Ghahramani, Raia Hadsell, Avinatan Hassidim, Katherine~Chou, Dale Webster, Jon~Shlens, and Pushmeet Kohli for their support during the course of this project.

\subsubsection*{Inclusion and ethics} 
While AI offers transformative potential in drug discovery, ethical considerations and transparency remain crucial. Biases in training data can lead to inequities, highlighting the need for diverse datasets and explainable AI systems. Our model, while still in the research stage, highlights the continuous need for development and refinement in this field. We acknowledge the difficulty in explaining the inner workings of complex models, but remain dedicated to advancing research in this area.

\subsubsection*{Data availability}
The Therapeutics Data Commons (TDC) datasets used for developing, benchmarking, and evaluating \ourmodel are publicly available on their \href{https://tdcommons.ai/}{website}. The benchmarking datasets used in this study—\href{https://huggingface.co/datasets/cais/hle}{Humanity's Last Exam (HLE)}, \href{https://huggingface.co/datasets/Idavidrein/gpqa}{GPQA (Diamond)}, \href{https://github.com/lamalab-org/chembench/tree/main}{ChemBench}, and \href{https://huggingface.co/datasets/ML2Healthcare/ClinicalTrialDataset}{TrialBench (Serious Adverse Event Prediction)}—are all publicly available via their respective websites. 

\subsubsection*{Code availability}
All of the components used in this work are available publicly. For reproducibility, we have documented technical methods and data curation detail in depth, while keeping the paper accessible to clinical and general scientific audiences. Specifically, all the data needs to reproduce this work is publicly accessible to the community. \ourmodel, a collection of lightweight state-of-the-art, open language models, are provided for researchers in three model size of 2B, 9B, and 27B and is accessible through \href{https://console.cloud.google.com/vertex-ai/publishers/google/model-garden/txgemma}{Vertex AI Model Garden} and \href{https://huggingface.co/collections/google/txgemma-release-67dd92e931c857d15e4d1e87}{Hugging Face}. \ourmodel's Github repository including supporting code and colab notebooks for quick start are also available at: \href{https://github.com/google-gemini/gemma-cookbook/tree/main/TxGemma}{https://github.com/google-gemini/gemma-cookbook/tree/main/TxGemma}. We have specifically provided starter colabs for \href{https://github.com/google-gemini/gemma-cookbook/blob/main/TxGemma/\%5BTxGemma\%5DQuickstart_with_Hugging_Face.ipynb}{inference}, \href{https://github.com/google-gemini/gemma-cookbook/blob/main/TxGemma/\%5BTxGemma\%5DFinetune_with_Hugging_Face.ipynb}{fine-tuning}, and exploring \href{https://github.com/google-gemini/gemma-cookbook/blob/main/TxGemma/\%5BTxGemma\%5DAgentic_Demo_with_Hugging_Face.ipynb}{agentic capabilities}. \ourmodel remains a research model and requires refinement. We look forward to working with research partners, regulators, and providers to validate and explore safe onward uses of \ourmodel.

\subsubsection*{Author Contributions} E.W., S.S., and S.A. made substantial contributions to the conception, design, and evaluation of this work. They played a key role in data analysis, interpretation of results, and the drafting and revision of the manuscript. P.F.J. contributed to drafting and revision of the manuscript. F.Z. contributed to the data processing and model training in the manuscript. R.P. contributed to obtaining necessary legal approvals, and organizational support. All authors participated in critically reviewing and revising the manuscript and interpreting the data and findings.

\subsubsection*{Competing interests}
This study was funded by Alphabet Inc and/or a subsidiary thereof (‘Alphabet’). E.W., S.S., P.F.J., F.Z., R.P., Y.M., J.B., D.F., and S.A. are employees of Alphabet and may own stock as part of the standard compensation package.

\newpage
\setlength\bibitemsep{3pt}
\balance
\clearpage

\printbibliography
\end{refsection}

\newpage
\begin{refsection}
\begin{appendices}
\clearpage

\renewcommand{\thefigure}{S.\arabic{figure}}
\renewcommand{\thetable}{S.\arabic{table}}

\setcounter{figure}{0}
\setcounter{table}{0}

\noindent \textbf{\LARGE{Supplementary Material}}\\
\normalfont

\noindent\textbf{\fontsize{13.}{15}\selectfont Version control} \\\\
\normalfont
\noindent\textbf{V0 (25 March 2025) $\rightarrow$ V1} 
\begin{itemize}[leftmargin=2.5em,rightmargin=0em]
\item Upgraded the \ouragent system's orchestrator from Gemini 2.0 to Gemini 2.5. This enhancement results in significant performance improvements in complex workflow orchestration, as detailed in \cref{tab:tx-agent-perf}.
\item Added performance results of \ourmodelpredict and \ourmodelconverse (trained only on commercially licensed datasets) for binary classification (\cref{tab:binary_results_commercial}), regression, and generation tasks (\cref{tab:regression_generation_commercial}).
\end{itemize}

\section{Summary}
\begin{itemize}[leftmargin=2.5em,rightmargin=0em]
\item Data details as listed in \cref{subsec-sup:additional_data_details}:
\begin{itemize}
    \item \cref{tab-sup:excluded_results}: Excluded TDC tasks and reasons for exclusion.
    \item \cref{tab-sup:binary_dataset_sizes}: Number of samples in training, validation, and test sets for all binary classification tasks.
    \item \cref{tab-sup:regression_generation_dataset_sizes}: Number of samples in training, validation, and test sets for all regression and generation tasks.
    \item \cref{tab-sup:binary_descriptions}: Descriptions of the binary classification tasks.
    \item \cref{tab-sup:regression_generation_descriptions}: Descriptions of the regression and generation tasks.
    \item \cref{tab-sup:data-representation} Types of features in the processed TDC data along with illustrative examples.
    \item \cref{fig-sup:dataset_size}: Distribution of TDC task sizes, aggregated over train, validation, and test sets.
\end{itemize}

\item Method and modeling details as listed in \cref{subsec:additional_method_details}:
\begin{itemize}
    \item \cref{tab:example_prompts_binary} Examples of prompts for binary classification tasks.
    \item \cref{tab:example_prompts_regression_generation} Examples of prompts for regression and generation tasks.
    \item \cref{tab:example_prompts_fewshot} Example of a 10-shot prompt for a binary classification task.
    \item \cref{tab:example_prompts_adverse} Example of prompts for predicting adverse events in clinical trials.
    \item \cref{tab-sup:example-react} Example of \ouragent response to a chemical preference question.
    \item \cref{tab-sup:txagent-tools} List of tools available to \ouragent.
    \item \cref{fig:nn_distribution} Distribution of Tanimoto similarities for 10 nearest neighbors by dataset splits in the AMES task.
    \item \cref{subsec:agg_comparison} Details about Wilcoxon signed-rank test used to assess model performance.
\end{itemize}

\item Additional results as listed in \cref{subsec:additional_results}:
\begin{itemize}
    \item Additional prediction results for \ourmodel (\cref{sec:appendix_predict_results})
    \begin{itemize}
    \item \cref{tab:binary_results} Performance on binary classification tasks for specialist SOTA, base \basemodel, and \ourmodelpredict models.
    \item \cref{tab:regression_generation_results} Performance on regression and generation tasks for specialist SOTA, base \basemodel, and \ourmodelpredict models.
    \item \cref{tab:binary_results_chat_and_txllm} Performance on binary classification tasks for \ourmodelpredict, \ourmodelconverse, and \ouroldmodel models.
    \item \cref{tab:regression_generation_chat_and_txllm} Performance on regression and generation tasks for \ourmodelpredict, \ourmodelconverse, and \ouroldmodel models.
    \item \cref{tab:binary_results_commercial} Performance on binary classification tasks for \ourmodelpredict and \ourmodelconverse models trained only on datasets with commercial licenses.
    \item \cref{tab:regression_generation_commercial} Performance on regression and generation tasks for \ourmodelpredict and \ourmodelconverse models trained only on datasets with commercial licenses.
    \item \cref{fig-sup:comparison-reletive-mix} Performance of \ourmodelpredictlargest compared to generalist and specialist SOTA models.
    \item \cref{fig-sup:llasmol} Comparison of \ourmodelpredictlargest with LlaSMol on select small molecule tasks.
    \item \cref{fig-sup:mole} Comparison of \ourmodelpredictlargest with MolE on select small molecule tasks.
    \item \cref{fig:speedtest} Inference speed of \ourmodel models at various sizes.
    \item \cref{fig:contamination-percent-decoy} Percent contamination for datasets and cosine similarity analysis.
    \item \cref{fig:contamination-performance} Performance on contaminated datasets before and after filtering out contaminated datapoints.
    \item \cref{fig:txgeamma-featuretype} Performance by feature type of all \ourmodelpredict sizes.
    \item \cref{fig:size_ablation} Comparison of \ourmodelpredict performances over different sizes and with \basemodel models.
    \item \cref{fig:toxtop-correlation} Correlations of \ourmodelpredictlargest predictions for toxicity and clinical trial approval tasks.
    \end{itemize}
    \item Conversing with \ourmodelpredictlargest and \ourmodelconverselargest (\cref{sec:appendix_converse_results})
    \begin{itemize}
       \item \cref{fig-sup:comparison-mmlu} Comparison of \ourmodelpredictlargest, \ourmodelconverselargest, and \basemodellargest on MMLU.
        \item \cref{fig-sup:27b_nontdc_prompts} Example of a dialogue with \ourmodelpredictlargest about general topics.
        \item \cref{fig-sup:27b-predict-conv-failure} Example of a multi-turn dialogue with \ourmodelpredictlargest about its predictions.
        \item \cref{fig-sup:txgemma-convo-trial} Example of a prompt format the enables \ourmodelconverse to provide reasoning for challenging tasks.
        \end{itemize}
   
    \item Additional \ouragent Results (\cref{sec:appendix_agent_tooluse})
    \begin{itemize}
        \item \cref{fig-sup:tool-use-frequency} \ouragent tool use frequencies for chemical preference and HLE benchmarks.
        \item \cref{fig-sup:tool-use-per-question} \ouragent tool use frequency per question for chemical preference questions.
    \end{itemize}
    \item Proof-of-concept example using \ourmodel (\cref{sec:ovarian_cancer})
    \begin{itemize}
       \item \cref{fig:end_to_end} Illustration of a possible application of \ourmodel to end-to-end therapeutic development.
    \end{itemize}
 \end{itemize}
 
\end{itemize}

\clearpage
\section{Data details}
\label{subsec-sup:additional_data_details}
This section provides a breakdown of the tasks used in our study, including information on excluded tasks and the size of training, validation, and test sets for binary classification, regression, and generation tasks.

As previously mentioned, we excluded a small number of tasks from TDC for various reasons. \cref{tab-sup:excluded_results} provides an overview of the excluded tasks and the rationale behind their exclusion. The primary reasons for exclusion were the tasks' relevance to the study, limitations of LLMs, and specific data characteristics, such as the absence of clear metrics or redundancy. For instance, tasks like QM7b, QM8, and QM9, which focus on predicting quantum properties, were not directly relevant to the study's focus on therapeutic development. Similarly, IEDB Jespersen and PDB Jespersen were excluded due to their small size and the complexity of implementing token prediction, as opposed to binary classification, within an LLM framework. Tasks such as DrugBank DDI, TWOSIDES, and USPTO Catalyst posed challenges due to the large number of potential labels, making them difficult for LLMs to process effectively. MOSES, ZINC, and ChEMBL were excluded because they lacked well-defined evaluation metrics. Finally, USPTO 50K and USPTO Reaction were excluded as they either overlapped with or were subsets of the USPTO task.

\begin{figure}[h]
    \centering
    \includegraphics[width=0.45\textwidth]{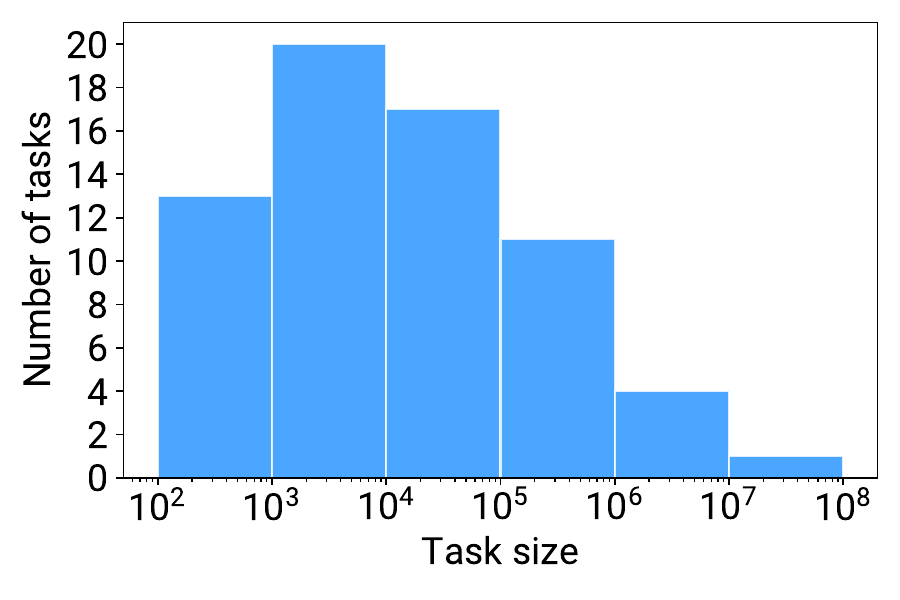}
    \caption{\textbf{Distribution of TDC task sizes, aggregated over train, validation, and test sets.} For tasks containing multiple datasets, such as ToxCast which contains data for more than 600 different assays, the task size is calculated by summing over the sizes for each dataset.}
    \label{fig-sup:dataset_size}
\end{figure}

\cref{tab-sup:binary_dataset_sizes,tab-sup:regression_generation_dataset_sizes} specify the number of samples in the training, validation, and test sets for the included binary classification, regression, and generation tasks, respectively. Substantial variability in task sizes across different tasks is shown in these tables. The binary classification tasks range from 196 to 1,406,988 samples, while the regression and generation tasks range from 345 to 775,767 samples. This variability highlights the diverse data availability landscape across various tasks. \cref{fig-sup:dataset_size} provides a visual representation of the distribution of TDC task sizes, aggregated across train, validation, and test sets. For tasks encompassing multiple subtasks, like ToxCast, the task size is computed by summing the sizes of each individual dataset.

\cref{tab-sup:binary_descriptions,tab-sup:regression_generation_descriptions} provide a brief description of the tasks, as well as the types of inputs (e.g. protein, small molecules, etc.). These tasks are diverse and encompass many different aspects of development. Some tasks corresponding to gene-disease association or protein-protein interaction prediction are useful for early-stage development, in order to identify mechanisms of disease and relevant targets. Predictions of antibody affinity, drug-target interaction, high-throughput screening, drug synergy are useful for intermediate development steps that involve proposing candidate therapeutics based on their interaction with a target. Predictions of toxicity, pharmacokinetics, and developability are useful for filtering candidates down based on favorable druglike properties. Predictions of clinical trial outcome, reaction yields, retrosynthesis are useful for late-stage development where understanding the likelihood of clinical trial approval and manufacturing potential are critical. There are also tasks that are highly specific for particular therapeutics types, which include predictions of CRISPR repair, peptide-MHC binding, miRNA-Target interaction, and TCR-epitope binding.

Binary classification tasks always output ``(A)'' or ``(B)'', where ``(A)'' is a negative answer to the question which is specified in the prompt and ``(B)'' is a positive answer. Regression tasks output an integer between 0 and 1000, which can be transformed back into the original task-specific label space. The output of the USPTO generation task is the SMILES string of the predicted molecules. \cref{tab-sup:data-representation} lists the different types of inputs in the processed TDC data along with illustrative examples.

\begin{table}[htbp]
\centering
\small
\caption{\textbf{Excluded TDC tasks and reasons for exclusion.} The tasks were excluded primarily due to their relevance to the study, limitations inherent to large language models (LLMs), and specific data characteristics, such as a lack of clear evaluation metrics or redundancy. }
\label{tab-sup:excluded_results}
\renewcommand{\arraystretch}{1.1}
\begin{tabular}{l|p{13cm}}
\toprule
Task Name & Reason for Exclusion \\
\midrule
QM7b & Prediction of quantum properties is not closely related to therapeutic development. \\ 
QM8 & Prediction of quantum properties is not closely related to therapeutic development. \\ 
QM9 & Prediction of quantum properties is not closely related to therapeutic development. \\ 
IEDB Jespersen & Amount of data is small, and token prediction is more difficult to implement in a LLM than binary classification. \\ 
PDB Jespersen & Amount of data is small, and token prediction is more difficult to implement in a LLM than binary classification. \\ 
DrugBank DDI & Large number of possible labels is difficult to implement in a LLM. \\ 
TWOSIDES & Large number of possible labels is difficult to implement in a LLM. \\ 
USPTO Catalyst & Large number of possible labels is difficult to implement in a LLM. \\ 
MOSES & No clear metric. \\ 
ZINC & No clear metric. \\ 
ChEMBL & No clear metric. \\ 
USPTO 50K & Subset of USPTO. \\ 
USPTO Reaction & Same data as USPTO. \\ 
\bottomrule
\end{tabular}
\end{table}

\begin{table}[htbp]
\centering
\small
\caption{\textbf{Number of samples in training, validation, and test sets for all binary classification tasks.} The binary classification tasks range in size from a minimum of 196 samples (Carcinogens Lagunin) to a maximum of 1,406,988 samples (butkiewicz), highlighting the considerable variability in data availability across different tasks. The task type and split type are also indicated following the TDC classification and recommendation.}
\label{tab-sup:binary_dataset_sizes}
\centerline{
\begin{tabular}{l|ccccc}
\toprule
Task Name & Task Type & Split Type & Training Size & Validation Size & Test Size \\
\midrule
AMES & Toxicity & Scaffold & 5,093 & 728 & 1,457  \\ 
BBB Martins & Pharmacokinetics & Scaffold & 1,421 & 203 & 406  \\ 
Bioavailability Ma & Pharmacokinetics & Scaffold & 1,344 & 192 & 384  \\ 
CYP1A2 Veith & Pharmacokinetics & Scaffold & 8,805 & 1,257 & 2,517  \\
CYP2C19 Veith & Pharmacokinetics & Scaffold & 8,865 & 1,266 & 2,534  \\ 
CYP2C9 Substrate CarbonMangels & Pharmacokinetics & Scaffold & 467 & 67 & 135  \\ 
CYP2C9 Veith & Pharmacokinetics & Scaffold & 8,463 & 1,210 & 2,419  \\
CYP2D6 Substrate CarbonMangels & Pharmacokinetics & Scaffold & 465 & 67 & 135  \\ 
CYP2D6 Veith & Pharmacokinetics & Scaffold & 9,191 & 1,313 & 2,626  \\
CYP3A4 Substrate CarbonMangels & Pharmacokinetics & Scaffold & 468 & 67 & 135  \\ 
CYP3A4 Veith & Pharmacokinetics & Scaffold & 8,628 & 1,233 & 2,467  \\
Carcinogens Lagunin & Toxicity & Scaffold & 196 & 28 & 56  \\ 
ClinTox & Toxicity & Scaffold & 1,034 & 147 & 297  \\ 
DILI & Toxicity & Scaffold & 325 & 54 & 96  \\ 
HIA Hou & Pharmacokinetics & Scaffold & 403 & 58 & 117  \\ 
HIV$^*$ & High-throughput screening & Scaffold & 28,788 & 4,112 & 8,227  \\ 
HuRI & Protein-protein interaction & Cold-start & 45,855 & 987 & 3,694  \\ 
MHC1 IEDB IMGT Nielsen & Peptide-MHC binding & Random & 130,190 & 18,598 & 37,197  \\ 
MHC2 IEDB Jensen & Peptide-MHC binding & Random & 93,997 & 13,428 & 26,856  \\ 
PAMPA NCATS & Pharmacokinetics & Scaffold & 1,423 & 203 & 408  \\ 
Pgp Broccatelli & Pharmacokinetics & Scaffold & 851 & 122 & 245  \\ 
SARSCOV2 3CLPro Diamond & High-throughput screening & Scaffold & 616 & 88 & 176  \\ 
SARSCoV2 Vitro Touret & High-throughput screening & Scaffold & 1,038 & 148 & 298  \\ 
SAbDab Chen & Developability & Random & 1,686 & 241 & 482  \\ 
Skin Reaction & Toxicity & Scaffold & 282 & 40 & 82  \\ 
Tox21 & Toxicity & Scaffold & 54,556 & 7,790 & 15,600  \\ 
ToxCast & Toxicity & Scaffold & 1,073,279 & 153,099 & 307,282  \\ 
butkiewicz & High-throughput screening & Random & 1,406,988 & 200,998 & 40,1997  \\ 
hERG & Toxicity & Scaffold & 457 & 66 & 132  \\ 
hERG Karim & Toxicity & Scaffold & 9,411 & 1,344 & 2,690  \\ 
herg central & Toxicity & Scaffold & 214,825 & 30,689 & 61,379  \\ 
miRTarBase & miRNA-target interaction & Random & 559,591 & 79,948 & 159,889  \\ 
phase1 & Clinical trial outcome & Cold-start & 1,546 & 258 & 598  \\ 
phase2 & Clinical trial outcome & Cold-start & 5,792 & 716 & 1,282  \\
phase3 & Clinical trial outcome & Cold-start & 41,25 & 532 & 1,084  \\
weber & TCR-epitope binding & Cold-start & 33,013 & 4,748 & 9,421  \\ 
\bottomrule
\end{tabular}
}
{\raggedright
\vspace{0.05in}
\scriptsize{
$*$ To predict whether compounds have Anti-HIV properties. \\ 
}}
\end{table}
\begin{table}[htbp]
\centering
\small
\caption{\textbf{Number of samples in training, validation, and test sets for all regression and generation tasks.} The regression and generation tasks vary significantly in size, ranging from a minimum of 345 samples (Protein SAbDab) to a maximum of 775,767 samples (USPTO). The task type and split type are also indicated following the TDC classification and recommendation.}
\label{tab-sup:regression_generation_dataset_sizes}
\renewcommand{\arraystretch}{1.1}
\centerline{
\begin{tabular}{l|ccccc}
\toprule
Task Name & Task Type & Split Type & Training Size & Validation Size & Test Size \\
\midrule
BindingDB Patent & Drug-target interaction & Temporal & 146,800 & 36,630 & 49,028  \\ 
BindingDB ic50 & Drug-target interaction & Cold-start & 375,127 & 7,531 & 31,495  \\ 
BindingDB kd & Drug-target interaction & Cold-start & 19,034 & 376 & 2,321  \\ 
BindingDB ki & Drug-target interaction & Cold-start & 57,656 & 1,189 & 4,709  \\ 
Buchwald Hartwig & Reaction yields & Random & 2,768 & 396 & 791  \\ 
Caco2 Wang & Pharmacokinetics & Scaffold & 637 & 91 & 182  \\ 
Clearance Hepatocyte AZ & Pharmacokinetics & Scaffold & 848 & 122 & 243  \\ 
Clearance Microsome AZ & Pharmacokinetics & Scaffold & 770 & 111 & 221  \\ 
DAVIS & Drug-target interaction & Cold-start & 12,455 & 266 & 1,064  \\ 
DisGeNET & Gene-disease association & Random & 39,425 & 5,621 & 11,200  \\ 
DrugComb Bliss & Drug synergy & Combination & 207,772 & 29,618 & 59,708  \\ 
DrugComb CSS & Drug synergy & Combination & 207,772 & 29,618 & 59,708  \\ 
DrugComb HSA & Drug synergy & Combination & 207,772 & 29,618 & 59,708  \\ 
DrugComb Loewe & Drug synergy & Combination & 207,772 & 29,618 & 59,708  \\ 
DrugComb ZIP & Drug synergy & Combination & 207,772 & 29,618 & 59,708  \\ 
GDSC1 & Drug response & Random & 124,117 & 17,731 & 35,462  \\ 
GDSC2 & Drug response & Random & 64,892 & 9,270 & 18,541  \\ 
Half Life Obach & Pharmacokinetics & Scaffold & 465 & 67 & 135  \\ 
KIBA & Drug-target interaction & Cold-start & 59,326 & 1,042 & 4,524  \\ 
LD50 Zhu & Toxicity & Scaffold & 5,168 & 739 & 1,478  \\ 
Leenay & CRISPR repair & Random & 5,325 & 760 & 1,520  \\ 
Lipophilicity AstraZeneca & Pharmacokinetics & Scaffold & 2,940 & 420 & 840  \\ 
OncoPolyPharmacology & Drug synergy & Combination & 16,014 & 2,331 & 4,707  \\ 
PPBR AZ & Pharmacokinetics & Scaffold & 1,952 & 279 & 559  \\ 
Protein SAbDab & Antibody affinity & Random & 345 & 49 & 99  \\ 
Solubility AqSolDB & Pharmacokinetics & Scaffold & 6,988 & 998 & 1,996  \\ 
TAP & Developability & Random & 845 & 120 & 240  \\ 
USPTO & Retrosynthesis & Random & 775,767 & 110,824 & 221,648  \\ 
USPTO Yields & Reaction yields & Random & 597,546 & 85,364 & 170,728  \\ 
VDss Lombardo & Pharmacokinetics & Scaffold & 791 & 113 & 226  \\ 
\bottomrule
\end{tabular}
}
\end{table}
\newcolumntype{P}[1]{>{\centering\arraybackslash}p{#1}}
\begin{table}[htbp]
\centering
\small
\caption{\textbf{Inputs and task descriptions for binary classification tasks.} All output responses are either (A) for negative or (B) for positive.}
\label{tab-sup:binary_descriptions}
\renewcommand{\arraystretch}{1.4}
\centerline{
\scriptsize
\begin{tabular}{@{}P{3.2cm}@{\hspace{0.2em}}|@{\hspace{0.18em}}P{3.2cm}@{\hspace{0.18em}}|@{\hspace{0.15em}}P{11cm}@{}}
\toprule
Task Name & Input & Description \\
\midrule
AMES & Small molecule & Given a drug SMILES, predict whether it is mutagenic.  \\
BBB Martins & Small molecule & Given a drug SMILES, predict whether it can cross the blood-brain barrier.  \\
Bioavailability Ma & Small molecule & Given a drug SMILES, predict whether it is orally available.  \\
CYP1A2 Veith & Small molecule & Given a drug SMILES, predict whether it inhibits CYP1A2.  \\
CYP2C19 Veith & Small molecule & Given a drug SMILES, predict whether it inhibits CYP2C19.  \\
CYP2C9 Substrate CarbonMangels & Small molecule & Given a drug SMILES, predict whether it is a substrate to CYP2C9.  \\
CYP2C9 Veith & Small molecule & Given a drug SMILES, predict whether it inhibits CYP2C9.  \\
CYP2D6 Substrate CarbonMangels & Small molecule & Given a drug SMILES, predict whether it is a substrate to CYP2D6.  \\
CYP2D6 Veith & Small molecule & Given a drug SMILES, predict whether it inhibits CYP2D6.  \\
CYP3A4 Substrate CarbonMangels & Small molecule & Given a drug SMILES, predict whether it is a substrate to CYP3A4.  \\
CYP3A4 Veith & Small molecule & Given a drug SMILES, predict whether it inhibits CYP3A4.  \\
Carcinogens Lagunin & Small molecule & Given a drug SMILES, predict whether it is a carcinogen.  \\
ClinTox & Small molecule & Given a drug SMILES, predict whether it is toxic.  \\
DILI & Small molecule & Given a drug SMILES, predict whether it can cause liver injury.  \\
HIA Hou & Small molecule & Given a drug SMILES, predict whether it is absorbed in the human intestine.  \\
HIV$^*$ & Small molecule & Given a drug SMILES, predict whether it has anti-HIV activity.  \\
HuRI & Protein & Given the amino acid sequences of two proteins, predict whether the proteins interact.  \\
MHC1 IEDB IMGT Nielsen & Protein & Given the amino acid of the peptide and pseudo amino acid of MHC 1, predict whether the peptide binds to the MHC.  \\
MHC2 IEDB Jensen & Protein & Given the amino acid of the peptide and pseudo amino acid of MHC 2, predict whether the peptide binds to the MHC.  \\
PAMPA NCATS & Small molecule & Given a drug SMILES, predict whether it is permeable in a PAMPA assay.  \\
Pgp Broccatelli & Small molecule & Given a drug SMILES, predict whether it inhibits Pgp.  \\
SARSCOV2 3CLPro Diamond & Small molecule & Given a drug SMILES, predict whether it binds SARS-CoV-2 3CL protease.  \\
SARSCoV2 Vitro Touret & Small molecule & Given a drug SMILES, predict whether it inhibits SARS-CoV-2 replication.  \\
SAbDab Chen & Protein & Given an antibody heavy chain and light chain sequence, whether it is developable.  \\
Skin Reaction & Small molecule & Given a drug SMILES, predict whether it can cause skin reaction.  \\
Tox21 & Small molecule & Given a drug SMILES, predict whether it is toxic in various assays.  \\
ToxCast & Small molecule & Given a drug SMILES, predict whether it is toxic in various assays.  \\
butkiewicz & Small molecule & Given a drug SMILES, predict whether it is active against various proteins.  \\
hERG & Small molecule & Given a drug SMILES, predict whether it blocks hERG.  \\
hERG Karim & Small molecule & Given a drug SMILES, predict whether it inhibits hERG.  \\
herg central & Small molecule & Given a drug SMILES, predict whether it inhibits hERG.  \\
miRTarBase & Nucleic acid \& protein & Given the miRNA mature and target amino acid, predict whether they interact.  \\
phase1 & Small molecule \& disease & Given a drug SMILES and disease, predict whether the phase 1 trial will be approved.  \\
phase2 & Small molecule \& disease & Given a drug SMILES and disease, predict whether the phase 2 trial will be approved.  \\
phase3 & Small molecule \& disease & Given a drug SMILES and disease, predict whether the phase 3 trial will be approved.  \\
weber & Protein & Given the amino acid  of the epitope and a T-cell receptor (amino acid of the hypervariable CDR3 loop), predict whether the epitope binds to the TCR.  \\
\bottomrule
\end{tabular}
}
{\raggedright
\vspace{0.05in}
\scriptsize{
$*$ To predict whether compounds have Anti-HIV properties. \\ 
}}
\end{table}
\newcolumntype{P}[1]{>{\centering\arraybackslash}p{#1}}
\begin{table}[htbp]
\centering
\footnotesize
\caption{\textbf{Inputs and task descriptions for regression and generation tasks.} Regression task outputs are integers between 0 and 1000, which represents a binned transformation of the original numeric label. On evaluation, the integer output is transformed back into the original numeric label space. For the USPTO generation task, the output is the SMILES string of the predicted set of small molecules.}
\label{tab-sup:regression_generation_descriptions}
\renewcommand{\arraystretch}{1.4}
\centerline{
\begin{tabular}{@{}P{3.3cm}@{\hspace{0.2em}}|@{\hspace{0.18em}}P{3.5cm}@{\hspace{0.18em}}|@{\hspace{0.15em}}P{11cm}@{}}
\toprule
Task Name & Input & Description \\
\midrule
BindingDB Patent & Protein \&  small molecule & Given the target amino acid and drug SMILES, predict their binding affinity.  \\
BindingDB ic50 & Protein & Given the target amino acid and drug SMILES, predict their IC50.  \\
BindingDB kd & Protein & Given the target amino acid and drug SMILES, predict their Kd.  \\
BindingDB ki & Protein & Given the target amino acid and drug SMILES, predict their Ki.  \\
Buchwald Hartwig & Small molecule & Given a product, a catalyst, and a reactant SMILES, predict the reaction yield.  \\
Caco2 Wang & Small molecule & Given a drug SMILES, predict the cell effective permeability.  \\
Clearance Hepatocyte AZ & Small molecule & Given a drug SMILES, predict the activity of hepatocyte clearance.  \\
Clearance Microsome AZ & Small molecule & Given a drug SMILES, predict the activity of microsome clearance.  \\
DAVIS & Protein \&  small molecule & Given the target amino acid and drug SMILES, predict their binding affinity.  \\
DisGeNET & Protein \&  disease & Given the disease description and the amino acid of the gene, predict their association.  \\
DrugComb Bliss & Small molecule \& cell line & Given two drug SMILESs and a cell line description, predict the drug synergy level.  \\
DrugComb CSS & Small molecule \&  cell line & Given two drug SMILESs and a cell line description, predict the drug synergy level.  \\
DrugComb HSA & Small molecule \&  cell line & Given two drug SMILESs and a cell line description, predict the drug synergy level.  \\
DrugComb Loewe & Small molecule \&  cell line & Given two drug SMILESs and a cell line description, predict the drug synergy level.  \\
DrugComb ZIP & Small molecule \&  cell line & Given two drug SMILESs and a cell line description, predict the drug synergy level.  \\
GDSC1 & Small molecule \&  cell line & Given a drug SMILES and a cell line description, predict the drug sensitivity level.  \\
GDSC2 & Small molecule \&  cell line & Given a drug SMILES and a cell line description, predict the drug sensitivity level.  \\
Half Life Obach & Small molecule & Given a drug SMILES, predict the half life duration.  \\
KIBA & Protein \&  small molecule & Given the target amino acid and drug SMILES, predict their binding affinity.  \\
LD50 Zhu & Small molecule & Given a drug SMILES, predict its LD50 toxicity.  \\
Leenay & Nucleic acid & Given a GuideSeq sequence, predict various properties.  \\
Lipophilicity AstraZeneca & Small molecule & Given a drug SMILES, predict the lipohilicity.  \\
OncoPolyPharmacology & Cell line \&  small molecule & Given two drug SMILESs and a cell line description, predict the drug synergy level.  \\
PPBR AZ & Small molecule & Given a drug SMILES, predict the plasma protein binding rate.  \\
Protein SAbDab & Protein & Given the amino acid of the antibody and  antigen, predict the binding affinity.  \\
Solubility AqSolDB & Small molecule & Given a drug SMILES, predict the activity of solubility.  \\
TAP & Protein & Given an antibody heavy chain and light chain sequence, predict its CDR length.  \\
USPTO & Small molecule & Given the product SMILES, generate the reactant SMILESs.  \\
USPTO Yields & Small molecule & Given a catalyst SMILES, reactant SMILES, and product SMILES, predict the yield.  \\
VDss Lombardo & Small molecule & Given a drug SMILES, predict the volume of distributon.  \\
\bottomrule
\end{tabular}
}
\end{table}
\begin{table}[htbp]
\centering
\footnotesize
\caption{\textbf{Types of drugs and targets found in our data.} Features found in our data as well as their textual representation and an illustrative example. Protein sequences are divided into several subtypes: some proteins and peptides are represented using their full amino acid sequence whereas MHC molecules are represented using the amino acid pseudo-sequences that only use residues in contact with a peptide, and TCRs only use CDR3 hypervariable loops.}
\label{tab-sup:data-representation}
\begin{tabular}{@{}l|@{\hspace{.2em}}l|@{\hspace{.4em}}l@{}}
\toprule
Representation Type               & Representation                                          & Example                                \\ \midrule
Small Molecules                   & SMILES string                                           & CN1C(=O)CN=C(C2=CCCCC2)c2cc(Cl)ccc21   \\
Amino Acid: Proteins and peptides & Amino acid sequences                                    & QLADETLLKV                             \\
Amino Acid: MHC molecules         & Pseudo-sequences~$\dagger$                              & YFAMYGEKVAHTHVDTLYVRYHYYTWAEWAYTWY     \\
Amino Acid: T cell receptors      & CDR3 hypervariable loops                                & CSASEGTSSYEQYF                         \\
Nucleic acid                      & Nucleotide sequence                                     & ACAGCCCAGCAGUUAUCACGGG                 \\
Disease                           & English text                                            & Chronic myeloproliferative disease     \\
Cell Line                         & English text                                            & NU-1, stomach cell sourced from cancer \\ \bottomrule 
\end{tabular}
\\
{\raggedright
\vspace{0.1in}
\scriptsize{
$\dagger$ Only for residues in contact with a peptide. \\
}}
\end{table}

\clearpage
\section{Method details}
\label{subsec:additional_method_details}

This section elaborates on the modeling choices employed in the development of \ourmodel. \cref{tab:example_prompts_binary,tab:example_prompts_regression_generation} illustrate prompts used for binary classification, regression, and generation tasks, showcasing the input structure for the model including the instructions and context provided to the model.  \cref{tab:example_prompts_fewshot} provide a concrete example of few-shot prompting applied to a binary classification task using 10 examples with nearest-neighbor shots. Each dataset in our data is structured as a text prompt, consisting of instructions, context, a question, and the corresponding answer. To provide relevant background, we created 2-3 sentence contexts based on TDC dataset descriptions and literature searches.  Prompts used for predicting adverse events in clinical trials based on the TrialBench dataset \cite{chen2024trialbench} are shown in \cref{tab:example_prompts_adverse}. To illustrate the reasoning process of \ouragent, \cref{tab-sup:example-react} provides an example of the steps taken to answer a chemical preference question from ChemBench. \cref{tab-sup:txagent-tools} also provides a comprehensive list of the tools available of \ouragent. \cref{subsec:agg_comparison} provides details of the Wilcoxon signed-rank test used to assess the performance of our models across all tasks.

\begin{figure}[h]
    \centering
    \includegraphics[width=0.5\textwidth]{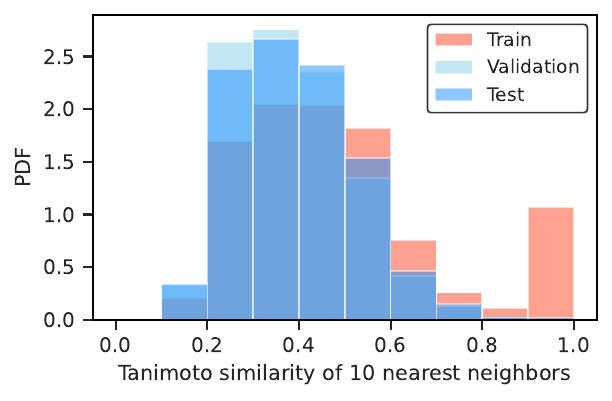}
    \caption{\small{\textbf{Distribution of the Tanimoto similarities for the 10 nearest neighbors in the AMES task.} Nearest neighbors are calculated from the training set for training and validation sets, and from both the training and validation sets for the test set.}}
    \label{fig:nn_distribution}
\end{figure}

We utilize random data points from the training set for few-shot learning during training. Although we use nearest neighbor shots for evaluation, we opt for random shots during training due to the higher intra-set similarity observed within the training data compared to between training and test sets, as illustrated in \cref{fig:nn_distribution}.

\subsection{Aggregated method comparison}
\label{subsec:agg_comparison}
For a pair of performances ($x_{i}$, $y_{i}$) of a task $i$, the test statistic of the Wilcoxon signed-rank test is calculated as the minimum of the positive-rank sum ($W^{+}$) and the negative-rank sum ($W^{-}$),

\begin{equation}\label{wilcoxon_pos}
W^{+} = \sum_{X_{i}>0}{R_{i}}
\end{equation}

\begin{equation}\label{wilcoxon_neg}
W^{-} = \sum_{X_{i}<0}{R_{i}}
\end{equation}

where $X_{i}=x_{i}-y_{i}$ and $R_{i}$ is the rank of $|x_{i}-y_{i}|$. In order to account for the differences in magnitudes for MAE and MSE metrics, we normalized all performances by the mean of the performances from both models. We also reversed the sign of MAEs and MSEs because lower MAEs and MSEs correspond to better performances.

\begin{table}[htbp]
\centering
\small
\caption{Example of prompts for binary classification tasks.}
\label{tab:example_prompts_binary}
\footnotesize
\begin{tabular}{p{16cm}}
\toprule

\color{black}
\textbf{Instructions}: Answer the following question about drug properties.

\textbf{Context}: As a membrane separating circulating blood and brain extracellular fluid, the blood-brain barrier (BBB) is the protection layer that blocks most foreign drugs. Thus the ability of a drug to penetrate the barrier to deliver to the site of action forms a crucial challenge in development of drugs for central nervous system.

\textbf{Question}: Given a drug SMILES string, predict whether it

(A) does not cross the BBB (B) crosses the BBB

Drug SMILES: CN1C(=O)CN=C(C2=CCCCC2)c2cc(Cl)ccc21

\textbf{Answer: (B)}\\\\

\color{black}
\textbf{Instructions}: Answer the following question about peptide-MHC binding.

\textbf{Context}: In the human body, T cells monitor the existing peptides and trigger an immune response if the peptide is foreign. To decide whether or not if the peptide is not foreign, the peptide must bind to a major histocompatibility complex (MHC) molecule. Therefore, predicting peptide-MHC binding affinity is pivotal for determining immunogenicity. In some experiments, the peptide binding is measured against cells that express multiple MHCs, so the peptide could be binding any one of the possible MHCs. Class 1 MHC molecules bind to peptides that are usually 8-14 amino acids long and activate CD8 T cells.

\textbf{Question}: Given the amino acid sequence of the peptide and possible pseudo amino acid sequences of MHC 1, predict whether the peptide

(A) does not bind to any of the MHCs (B) binds to any of the MHCs

Peptide amino acid sequence: QLADETLLKV

Possible MHC pseudosequences: YFAMYGEKVAHTHVDTLYVRYHYYTWAEWAYTWY

\textbf{Answer: (B)}\\\\

\color{black}
\textbf{Instructions}: Answer the following question about miRNA protein interactions.

\textbf{Context}: MicroRNAs (miRNAs) are, small non-coding RNAs with 18–25 nucleotides, which are central regulators at the post-transcriptional level in both animals and plants. Perfect or near-perfect complementary binding of miRNAs and their target mRNA negatively regulates gene expression by accelerating mRNA degradation or suppressing mRNA translation.

\textbf{Question}: Given the miRNA mature sequence and target amino acid sequence, predict whether

(A) the miRNA and target do not interact (B) the miRNA and target interact

miRNA sequence: UUCCUGUCAGCCGUGGGUGCC

Target amino acid sequence: MSVNMDELRHQVMINQFVLAAGCAADQAKQLLQAAHWQFETALSTFFQET-NIPNSHHHHQMMCTPSNTPATPPNFPDALAMFSKLRASEGLQSSNSPMTAAACSPPANFSPFWASSPPSHQAPWIP-PSSPTTFHHLHRPQPTWPPGAQQGGAQQKAMAAMDGQR

\textbf{Answer: (A)}\\\\

\color{black}
\textbf{Instructions}: Answer the following question about clinical trials.

\textbf{Context}: Clinical trial is the most time and cost-consuming step in the drug discovery process. Phase 1 clinical trials test the safety and basic properties of a new drug or treatment in a small group of people for the first time. Optimizing and designing trials with machine learning could drastically lead to the speedup of delivery of life-saving therapeutics to patients. Clinical trial outcome prediction is a machine learning task that aims to forecast the outcome of clinical trials, such as the approval rate of a drug or treatment. It utilizes various clinical trial features, including the drug's molecular structure and patient disease.

\textbf{Question}: Given a drug SMILES string and disease, predict if the phase 1 trial

(A) would not be approved (B) would be approved

Drug SMILES: COC1=NC(N)=NC2=C1N=CN2[C@@H]1O[C@H](CO)[C@@H](O)[C@@H]1O

Disease: Chronic myeloproliferative disease

\textbf{Answer: (A)} \\

\color{black}

\hrule

\end{tabular}
\end{table}

\begin{table}[htbp]
\centering
\small
\caption{Example of prompts for regression and generation tasks.}
\label{tab:example_prompts_regression_generation}
\footnotesize
\begin{tabular}{p{16cm}}
\toprule

\color{black}
\textbf{Instructions}: Answer the following question about drug properties.

\textbf{Context}: The human colon epithelial cancer cell line, Caco-2, is used as an in vitro model to simulate the human intestinal tissue. The experimental result on the rate of drug passing through the Caco-2 cells can approximate the rate at which the drug permeates through the human intestinal tissue.

\textbf{Question}: Given a drug SMILES string, predict its normalized Caco-2 cell effective permeability from 000 to 1000, where 000 is minimum permeability and 1000 is maximum permeability.

Drug SMILES: O=C(O)COC(=O)Cc1ccccc1Nc1c(Cl)cccc1Cl

\textbf{Answer: 788}\\\\

\color{black}
\textbf{Instructions}: Answer the following question about drug responses.

\textbf{Context}: The same drug compound could have various levels of responses in different patients. To design drug for individual or a group with certain characteristics is the central goal of precision medicine. In experiments, IC50s of drugs were measured against cancer cell lines.

\textbf{Question}: Given a drug SMILES string and a cell line description, predict the normalized drug sensitivity from 000 to 1000, where 000 is minimum drug sensitivity and 1000 is maximum drug sensitivity.

Drug SMILES: CN1C=C(C2=CC=CC=C21)/C=C\textbackslash3/C4=C(C=CC=N4)NC3=O

Cell line description: SNU-1, stomach cell sourced from cancer

\textbf{Answer: 615}\\\\

\color{black}

\color{black}
\textbf{Instructions}: Answer the following question about drug target interactions.

\textbf{Context}: Drug-target binding is the physical interaction between a drug and a specific biological molecule, such as a protein or enzyme. This interaction is essential for the drug to exert its pharmacological effect. The strength of the drug-target binding is determined by the binding affinity, which is a measure of how tightly the drug binds to the target. Kd is the dissociation constant of a drug-target complex. It is the concentration of drug at which half of the drug-target complexes have dissociated. A lower Kd value indicates a stronger binding affinity.

\textbf{Question}: Given the target amino acid sequence and compound SMILES string, predict their normalized binding affinity Kd from 000 to 1000, where 000 is minimum Kd and 1000 is maximum Kd.

Drug SMILES: O=S(=O)(O)c1cccc2cccc(Nc3ccccc3)c12

Target amino acid sequence: MATVQQLEGRWRLVDSKGFDEYMKELGVGIALRKMGAMAKPDC-IITCDGKNLTIKTESTLKTTQFSCTLGEKFEETTADGRKTQTVCNFTDGALVQHQEWDGKESTITRKLKDGKLVV-ECVMNNVTCTRIYEKVE

\textbf{Answer: 397}\\\\

\color{black}
\textbf{Instructions}: Answer the following question about reactions.

\textbf{Context}: Retrosynthesis is the process of finding a set of reactants that can synthesize a target molecule, i.e., product, which is a fundamental task in drug manufacturing. The target is recursively transformed into simpler precursor molecules until commercially available "starting" molecules are identified. In a data sample, there is only one product molecule, reactants can be one or multiple molecules.

\textbf{Question}: Given a product SMILES string, predict the reactant SMILES string.

Product SMILES: [CH2:12]1[C:7]2([CH2:6][CH2:5][O:15][CH2:1][CH2:8]2)[CH2:13][CH2:14][O:10][C:11]1=[O:17]

\textbf{Answer: [CH:1]12B[CH:5]([CH2:6][CH2:7][CH2:8]1)CCC2.[O:10]1[CH2:14][CH2:13][CH2:12] [CH2:11]1.[OH-:15].[Na+].[OH:17]O.Cl}\\\\

\hrule

\end{tabular}
\end{table}

\begin{table}[htbp]
\centering
\small
\caption{Example of a 10-shot prompt for a binary classification task. Shots are selected from nearest neighbors in the combined training and validation set (not the test set).}
\label{tab:example_prompts_fewshot}
\footnotesize
\begin{tabular}{p{16cm}}
\toprule

\color{black}
\textbf{Instructions}: Answer the following question about drug properties.

\textbf{Context}: As a membrane separating circulating blood and brain extracellular fluid, the blood-brain barrier (BBB) is the protection layer that blocks most foreign drugs. Thus the ability of a drug to penetrate the barrier to deliver to the site of action forms a crucial challenge in development of drugs for central nervous system. \\\\

\textbf{Question}: Given a drug SMILES string, predict whether it (A) does not cross the BBB (B) crosses the BBB \\

\color{black}
Drug SMILES: CN1C(=O)CN=C(c2ccccc2)c2cc(Cl)ccc21 \\
Answer: (B) \\\\

Drug SMILES: CN1C(=O)CN=C(c2ccccc2F)c2cc(Cl)ccc21 \\
Answer: (B) \\\\

Drug SMILES: CN1C(=S)CN=C(c2ccccc2)c2cc(Cl)ccc21 \\
Answer: (B) \\\\

Drug SMILES: CP(C)(=O)CN1C(=O)CN=C(c2ccccc2)c2cc(Cl)ccc21 \\
Answer: (B) \\\\

Drug SMILES: CN1C(=O)CN=C(c2ccccc2)c2cc([N+](=O)[O-])ccc21 \\ 
Answer: (B) \\\\

Drug SMILES: CCN(CC)CCN1C(=O)CN=C(c2ccccc2F)c2cc(Cl)ccc21 \\
Answer: (B) \\\\

Drug SMILES: O=C1CN=C(c2ccccc2)c2cc(Cl)ccc2N1CC1CC1 \\
Answer: (B) \\\\

Drug SMILES: C\#CCN1C(=O)CN=C(c2ccccc2)c2cc(Cl)ccc21 \\
Answer: (B) \\\\

Drug SMILES: O=C1CN=C(c2ccccc2)c2cc(Cl)ccc2N1CC(F)(F)F \\
Answer: (B) \\\\

Drug SMILES: CCS(=O)(=O)CCN1C(=O)CN=C(c2ccccc2F)c2cc(Cl)ccc21 \\
Answer: (B) \\

\color{black}
Drug SMILES: CN1C(=O)CN=C(C2=CCCCC2)c2cc(Cl)ccc21

\color{black}
\textbf{Answer: (B)}\\

\color{black}
\hrule

\end{tabular}
\end{table}

\begin{table}[htbp]
\centering
\small
\caption{Example of prompts for predicting adverse events in clinical trials. The top prompt only provides drug SMILES strings while the bottom prompt also includes textual information about the clinical trial.}
\label{tab:example_prompts_adverse}
\footnotesize
\begin{tabular}{p{16cm}}
\toprule

From the following information about a clinical trial, predict whether it would have an adverse event. \\\\

Drug: CC[C@H]1[C@@H](COC2=C3C=C(OC)C(=CC3=CC=N2)C(N)=O)NC(=O)[C@H]1F\\.[H][C@@]12CC[C@H](O)[C@@]1(C)CC[C@]1([H])C3=C(CC[C@@]21[H])C=C(O)C=C3 \\\\

\textbf{Answer}: No \\\\\\\\

From the following information about a clinical trial, predict whether it would have an adverse event. \\\\

Title: A Study To Estimate The Effect of PF-06650833 On The Pharmacokinetics (PK) of Oral Contraceptive (OC) \\
Summary: This is a Phase 1, open label, fixed sequence study of the effect of multiple dose PF-06650833 on single dose OC PK in healthy female subjects. \\   
Phase: 1 \\
Disease: Healthy \\
Minimum age: 18 Years \\
Maximum age: 60 Years \\
Healthy volunteers: Accepts Healthy Volunteers \\
Interventions: 400 mg by mouth (PO) Once daily (QD) for 11 days; Single dose of Oral tablet containing 30 ug EE and 150 ug of LN \\
Drug: CC[C@H]1[C@@H](COC2=C3C=C(OC)C(=CC3=CC=N2)C(N)=O)NC(=O)[C@H]1F\\.[H][C@@]12CC[C@H](O)[C@@]1(C)CC[C@]1([H])C3=C(CC[C@@]21[H])C=C(O)C=C3 \\\\

\textbf{Answer}: No \\

\hrule

\end{tabular}
\end{table}

\begin{table}[t]
\centering
\caption{\small{\textbf{Example of \ouragent ChemBench chemical preference question answering.}}}
\label{tab-sup:example-react}
\renewcommand{\arraystretch}{1.}
\footnotesize
\centering
\begin{tabular}{l|p{14cm}} 
\toprule
\rowcolor{ourlightgray} 
\multicolumn{2}{l}{\textbf{Example of Chembench Chemical Preference}} \\ \midrule

Question & {Consider the following two drug candidates: \newline A. N\#Cc1c(NC(=O)c2cc([N+](=O)[O-])ccc2N2CCOCC2)sc2c1CCCC2 \newline B. O=C(C=Cc1ccccn1)c1ccccc1.\newline Which is more preferable for further development?} \\  \midrule
Thought 1 & First, I need to obtain more information about each drug candidate. I will use the SMILES to Description tool to get detailed descriptions of each molecule from their SMILES strings, starting with candidate. \\ 
Action 1 & SMILES to Description \newline
Input SMILES: N\#Cc1c(NC(=O)c2cc([N+](=O)[O-])ccc2N2CCOCC2)sc2c1CCCC2 
         \\  
Observation  1                 & PubChem CID: 3934361 \newline 
Molecular Formula: C20H20N4O4S \newline 
Molecular Weight: 412.5 \newline 
Charge: 0 \newline 
IUPAC Name: N-(3-cyano-4,5,6,7-tetrahydro-1-benzothiophen-2-yl)-2-morpholin-4-yl-5-nitrobenzamide \newline 
XLogP: 3.8 \newline 
Synonyms: MLS000335194, CHEMBL1549645, HMS2597A10, HMS3379H10, AKOS001044982, SMR000249952, SR-01000056848, SR-01000056848-1, Z27367728 
                \\ 
Observation 1 & (Summarized) The provided information describes a single molecule (PubChem CID 3934361) with a cyano-tetrahydro-benzothiophene core, a morpholino-nitrobenzamide substituent, and an XLogP of 3.8, indicating its lipophilicity.
         \\  \midrule
Thought  2 &  Now, I will use the SMILES to Description tool to get detailed descriptions of candidate B from its SMILES string. After that, I will compare the information obtained for both candidates to determine which is more preferable for further development.
         \\  
Action  2  & SMILES to Description \newline
Input SMILES: O=C(C=Cc1ccccn1)c1ccccc1
         \\  
Observation  2  & PubChem CID: 219207 \newline
Molecular Formula: C14H11NO \newline
Molecular Weight: 209.24 \newline
Charge: 0 \newline
IUPAC Name: 1-phenyl-3-pyridin-2-ylprop-2-en-1-one \newline
XLogP: 2.7 \newline
Synonyms: 3-(2-PYRIDYL)-ACRYLOPHENONE, MLS002637493, azachalcone, CHEMBL1717486, DTXSID601279307, HMS3079I05, SMR001547031, \newline 1-Phenyl-3-(2-pyridinyl)-2-propen-1-one
         \\  
Observation  2   & (Summary) The molecule (B) represented by the SMILES O=C(C=Cc1ccccn1)c1ccccc1, also known as 1-phenyl-3-pyridin-2-ylprop-2-en-1-one, has a molecular weight of 209.24 g/mol and a calculated XLogP value of 2.7.
         \\  \midrule
Thought  3   & Based on the descriptions, candidate B (XLogP = 2.7) is less lipophilic than candidate A (XLogP = 3.8). Lower lipophilicity can often be associated with better absorption and distribution properties. I will now use the ClinicalTox tool to assess the clinical toxicity of candidate B.
         \\  
Action  3 & ClinicalTox \newline
Input SMILES: O=C(C=Cc1ccccn1)c1ccccc1
         \\  
Observation  3  & Context: Humans are exposed to a variety of chemicals through food, household products, and medicines, some of which can be toxic, leading to over 30\% of promising pharmaceuticals failing in human trials due to toxicity. Toxic drugs can be identified from clinical trials that failed due to toxicity, while non-toxic drugs can be identified from FDA approval status or from clinical trials that report no toxicity. \newline Prediction returned: O=C(C=Cc1ccccn1)c1ccccc1 is not toxic!
         \\  
Observation  3  & (Summary) Based on the provided information, drug candidate B (O=C(C=Cc1ccccn1)c1ccccc1) is predicted to be non-toxic, suggesting it might be more preferable for further development.\\ \midrule
Final Response   & Candidate B is more preferable for further development. Here's why:  \newline - \textbf{Lipophilicity (XLogP)}: Candidate B has a lower XLogP value (2.7) compared to Candidate A (3.8). Lower lipophilicity is generally favored in drug development. Excessive lipophilicity (high XLogP) can lead to problems like poor solubility, increased risk of off-target binding and toxicity, and difficulties in achieving good absorption and distribution in the body. A value closer to the range of many oral drugs is preferred. \newline - \textbf{Toxicity}: The clinical toxicity tool states that candidate B is predicted to be non-toxic. \\

\bottomrule
\end{tabular}
\end{table}

\begin{table}[htbp]
\centering
\small
\caption{\textbf{Descriptions of tools used by \ouragent.}}
\label{tab-sup:txagent-tools}
\renewcommand{\arraystretch}{1.}
\begin{tabular}{l|p{12cm}}
\toprule
Tool Name & Description \\
\midrule
ToxCast & Uses \ourmodel to predict the toxicity of a given drug (SMILES string) in various ToxCast assays based on the provided context. Returns results indicating whether the drug is toxic or not in each selected assay.  \\  \hline
ClinicalTox & Uses \ourmodel to predict the clinical toxicity of a given drug (SMILES string) for humans. Returns a result indicating whether the drug is predicted to be toxic or not. \\ \hline 
Chat & Allows conversational interaction with \ourmodelconverse. Enables posing therapeutics-related questions and receiving responses.  \\ \hline
Mutagenicity & Uses \ourmodel to predict whether a given drug (SMILES) is mutagenic based on the Ames test. Returns a result indicating if the drug is mutagenic or not. \\ \hline
IC$_{50}$ & Uses \ourmodel to predict the normalized IC$_{50}$ between a drug (SMILES) and a target protein (amino acid sequence). Returns a IC$_{50}$ value, with lower values suggesting potent inhibition. \\ \hline
Phase 1 Trial & Uses \ourmodel to predict the approval outcome of a Phase 1 clinical trial for a drug (SMILES) against a specified disease. Returns a result indicating whether the trial would be approved or not. \\ \hline
Wikipedia Search & Searches Wikipedia for a given text query. Returns the top matching article's title, link, and a short summary. \\ \hline
PubMed Search & Queries PubMed for scientific articles based on a search text. Returns metadata (PMID, title, authors, journal, date, abstract) for the top few articles. \\ \hline
Web Search & Performs a general web search. Returns titles, links, and snippets for the top search results. \\ \hline
HTML Fetch & Fetches the raw HTML content of a given URL. Useful for inspecting webpage details.  \\ \hline
SMILES to Description & Retrieves molecular information from PubChem for a given SMILES string.  Returns properties like PubChem CID, molecular formula, IUPAC name, XLogP, and synonyms. \\ \hline
SMILES Therapy & Retrieves therapeutic information (ChEMBL ID, mechanisms of action, drug indications, ATC classifications) for a drug given its SMILES string. \\ \hline
Molecule Tool & Provides molecule-related functions: searching for compounds by name (returns properties and IDs) and converting between molecular representations (InChI, SMILES, InChIKey, Mol). \\ \hline
Molecule Convert & Converts a molecules representation from one type to another (e.g., SMILES to InChI). \\ \hline
Gene Sequence & Retrieves amino acid sequences for a given gene name and organism. Searches NCBI Nucleotide, fetches records, and translates DNA to protein sequences.  \\ \hline
Gene Description & Retrieves descriptive information about a gene from NCBI Gene, including official symbol, full name, description, and summary.  \\ \hline
BlastP & Runs a BLASTP search against NCBI databases for a given amino acid sequence. Returns hits with gene names, organisms, and accessions. \\ \hline
Protein Description & Provides descriptive information (organism, definition, accession) for a protein, either by name or amino acid sequence. Uses NCBI Protein database or BLASTP. \\  
\bottomrule
\end{tabular}
\end{table}

\clearpage

\section{Additional results}\label{subsec:additional_results}

\subsection{\ourmodelpredict performance}
\label{sec:appendix_predict_results}
\begin{figure}
    \centering
    \includegraphics[width=0.8\textwidth]{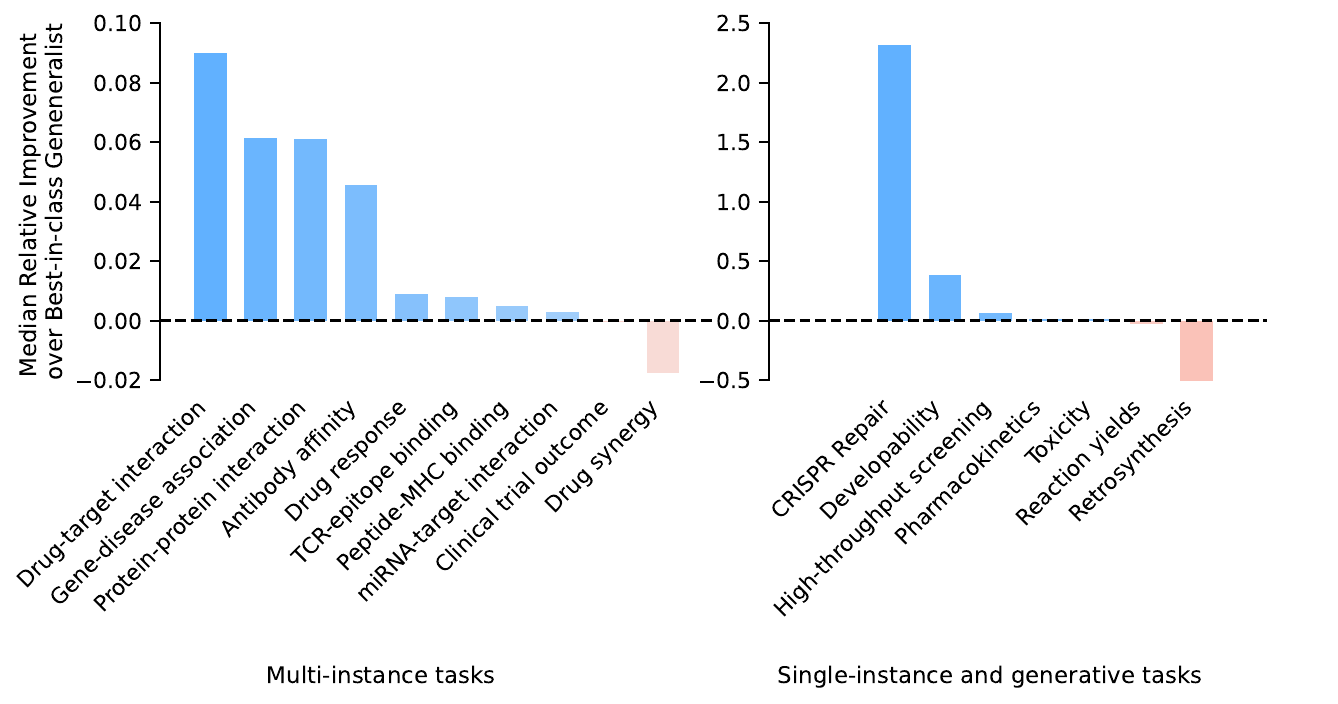} 
    \includegraphics[width=0.8\textwidth]{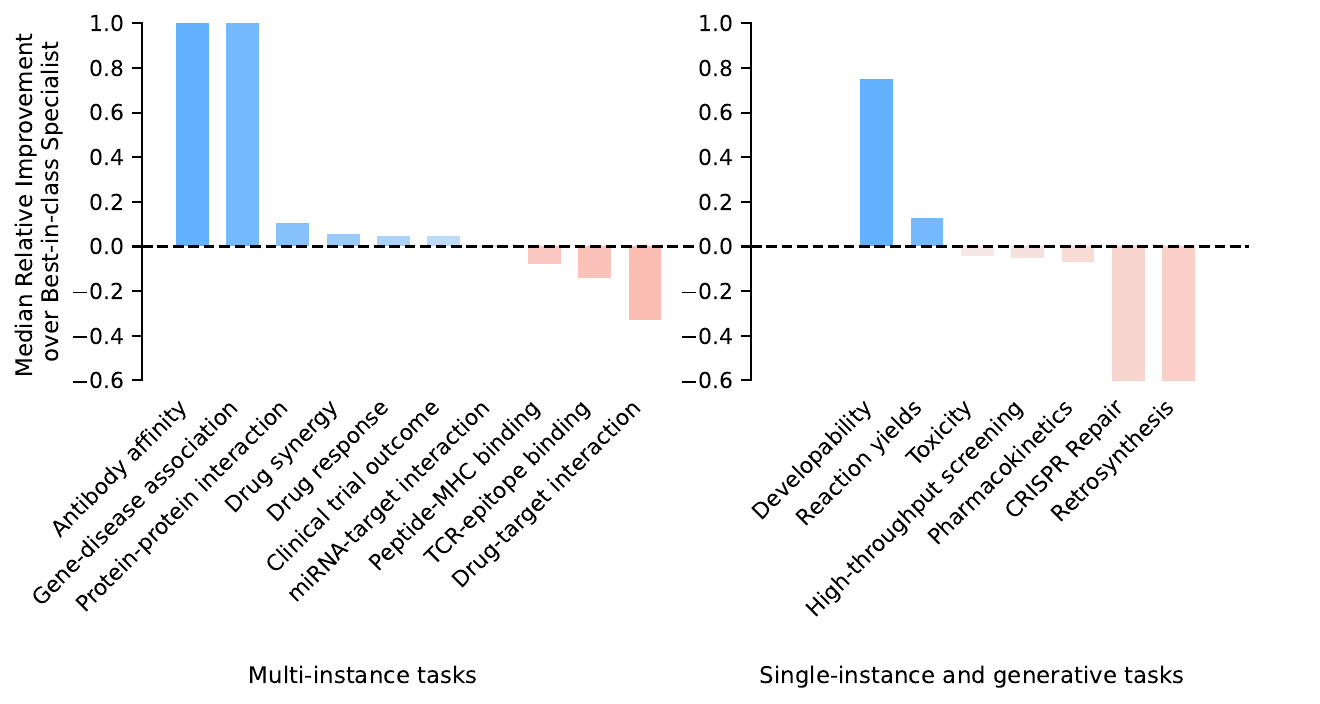}    
    \includegraphics[width=0.8\textwidth]{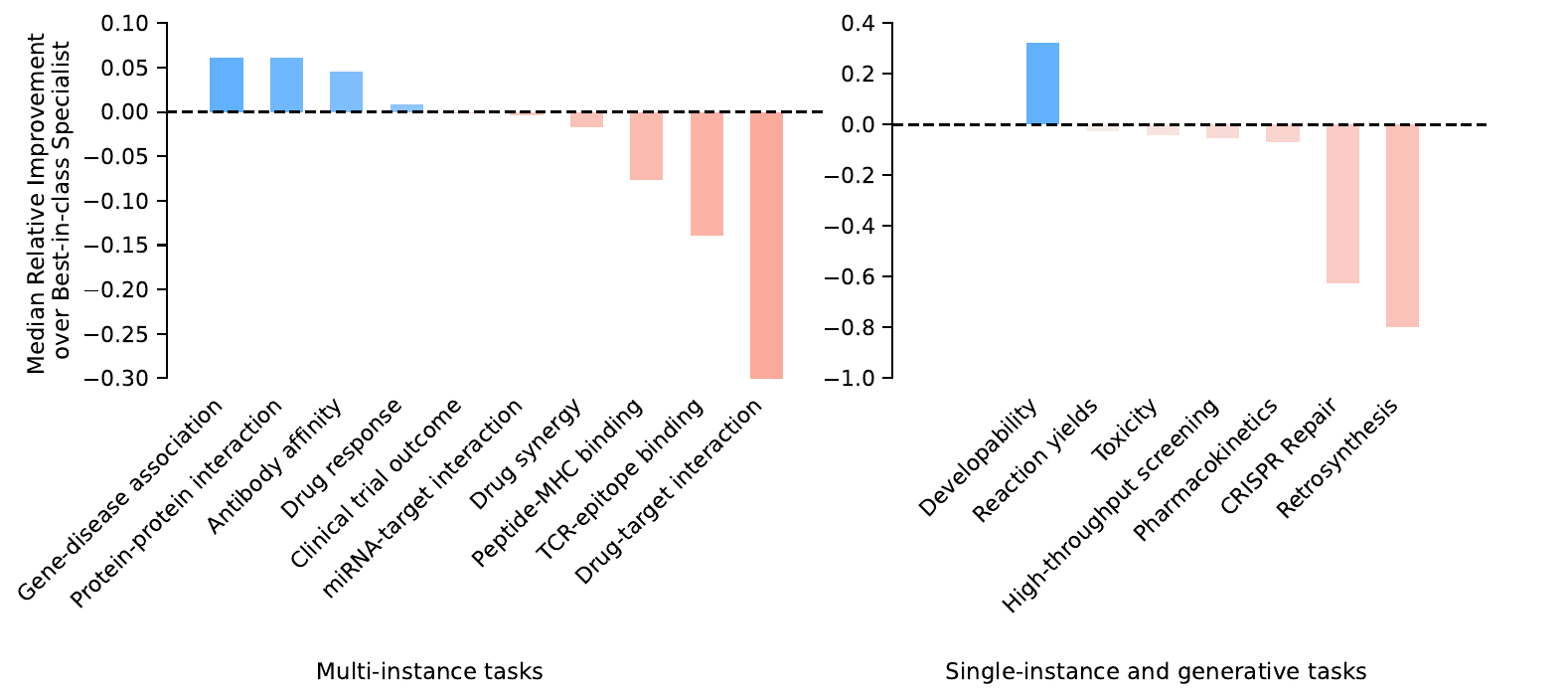}
    \vspace{6pt}
    \caption{\small{\textbf{Performance of \ourmodelpredictlargest compared to generalist and specialist SOTA models} (\textbf{top}) The median relative change in performance of \ourmodelpredictlargest compared to \ouroldmodelm. (\textbf{middle}) The median relative change in performance of \ourmodelpredictlargest compared to specialist SOTA models. (\textbf{bottom}) The median relative change in performance of \ourmodelpredictlargest compared to all SOTA models, including both \ouroldmodelm and specialist models. Multi-instance tasks indicate tasks that involve multiple features, whereas single-instance tasks only involve one feature. The tasks within each task type are defined in \cref{tab-sup:binary_dataset_sizes,tab-sup:regression_generation_dataset_sizes}.}}
    \label{fig-sup:comparison-reletive-mix}
\end{figure}

\begin{figure}
    \centering
    \includegraphics[width=0.5\textwidth]{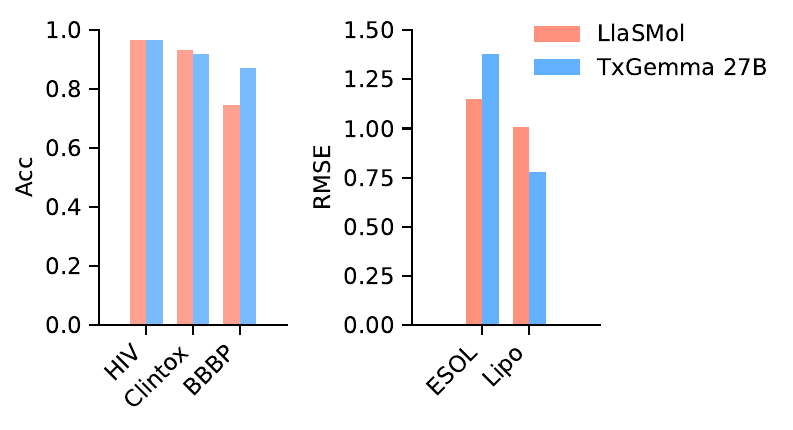}
    \vspace{6pt}
    \caption{\small{\textbf{\ourmodel performs comparably to LlaSMol on small molecule tasks.} Accuracy is reported for binary classification tasks, and RMSE is reported for regression tasks. BBBP corresponds to BBB Martins in TDC tasks, ESOL corresponds to Solubility AqSolDB, and Lipo corresponds to Lipophilicity AstraZeneca.}}
    \label{fig-sup:llasmol}
\end{figure}

\begin{figure}
    \centering
    \includegraphics[width=1.0\textwidth]{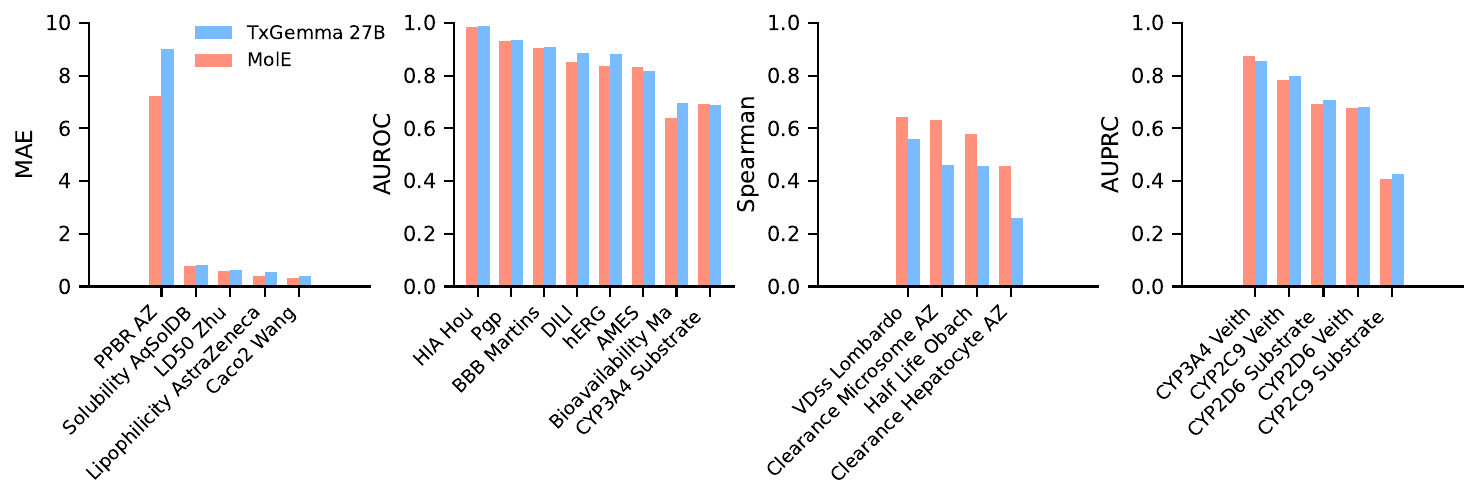}
    \vspace{6pt}
    \caption{\small{\textbf{\ourmodel performs comparably to MolE on small molecule tasks.} Comparison of MolE with \ourmodelpredictlargest on TDC tasks, separated by metric type (MAE, AUROC, Spearman correlation, and AUPRC). \ourmodelpredictlargest performs better than MolE on 10 out of 22 tasks.}}
    \label{fig-sup:mole}
\end{figure}

\begin{figure}
    \centering
    \includegraphics[width=0.6\textwidth]{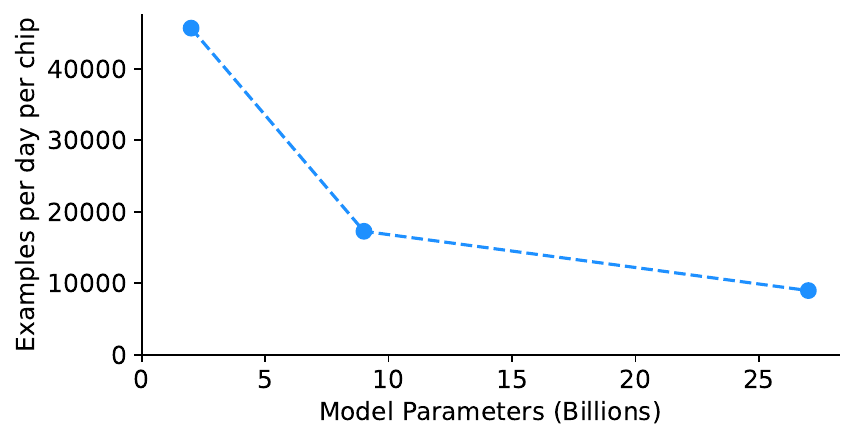}
    \vspace{6pt}
    \caption{\textbf{Inference speed of \ourmodel models.} The number of examples inferenced per day at different model sizes, normalized by the number of TPUv5e chips used for serving. The PPBR AZ task was used for the benchmarking due to its reasonable size.}
    \label{fig:speedtest}
\end{figure}

\begin{figure}
    \centering
    \includegraphics[width=0.49\textwidth]{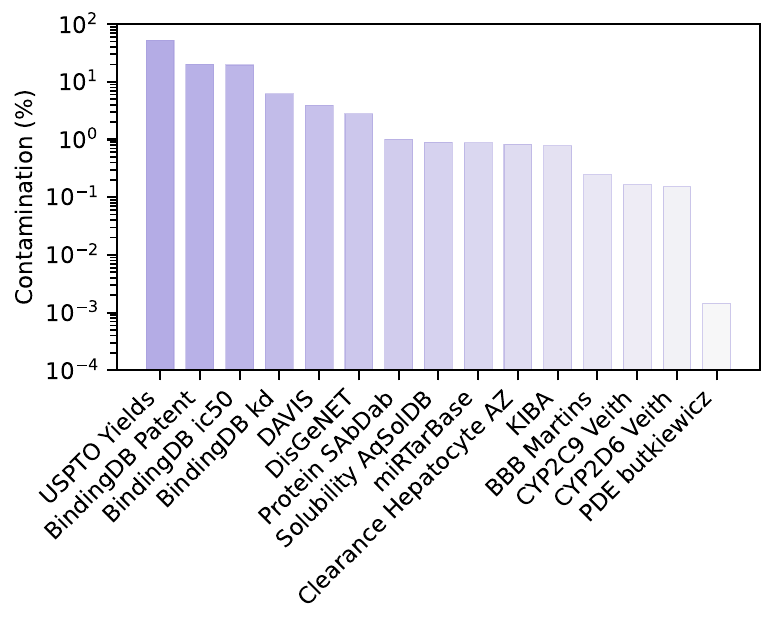} \hfill\\
    \includegraphics[width=0.49\textwidth]{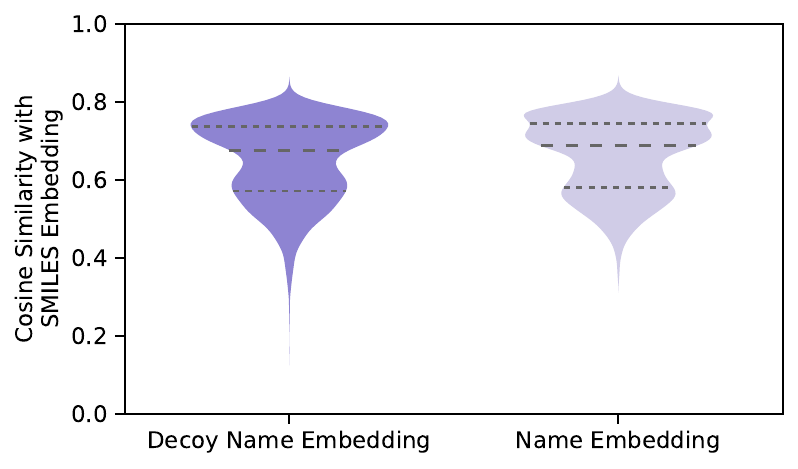}
    \vspace{0pt}
    \caption{\small{\textbf{Contamination analysis.} \textbf{(top)} Out of \ntasks tasks, 23\% had some datapoints in the test set that were found in the \basemodel pretraining data, while 77\% did not. For tasks that had some contaminated datapoints, we plot the percent of the test set that was contaminated. \textbf{(bottom)} Distributions of cosine similarities between SMILES string embeddings and molecular name embeddings. Decoy name embeddings indicate a random different molecule name.}}
    \label{fig:contamination-percent-decoy}
\end{figure}

\begin{figure}
    \centering
    \includegraphics[width=0.85\textwidth]{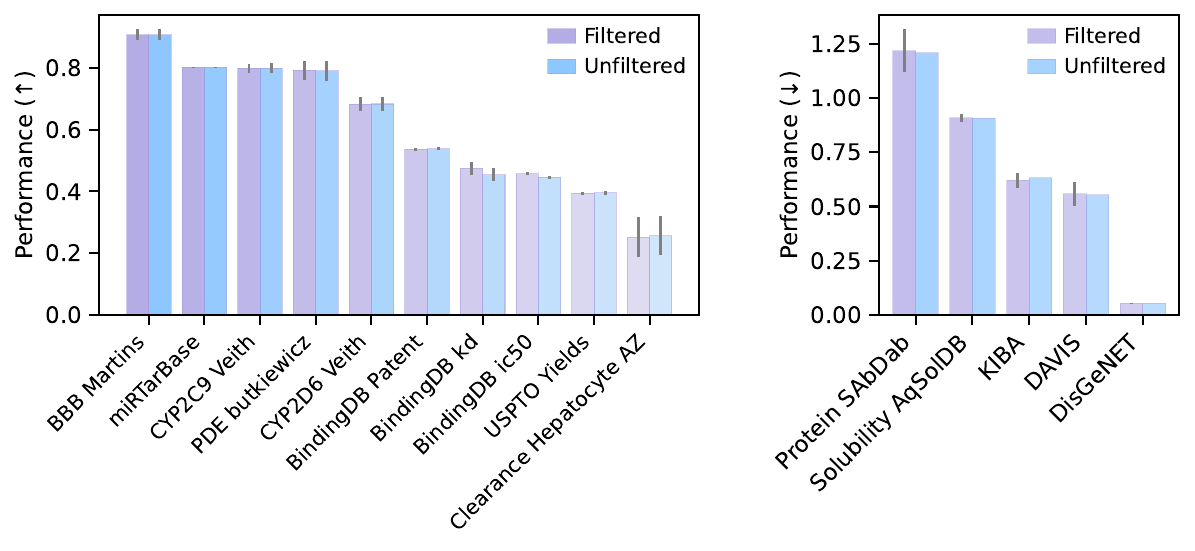}
    \vspace{0pt}
    \caption{\small{\textbf{Model performance after filtering contaminated datapoints.} Performance of \ourmodelpredictlargest on both original unfiltered test sets and filtered test sets in which contaminated datapoints were removed. \textbf{(left)} For these tasks, higher values correspond to better models, and the metrics are defined in \cref{tab:binary_results,tab:regression_generation_results}. Error bars are bootstrapped standard errors. \textbf{(right)} For these tasks, lower values correspond to better models, and the metrics (either MAE or MSE) are defined in \cref{tab:binary_results,tab:regression_generation_results}. Error bars are bootstrapped standard errors.}}
    \label{fig:contamination-performance}
\end{figure}

\begin{figure}
    \centering
    \includegraphics[width=\textwidth]{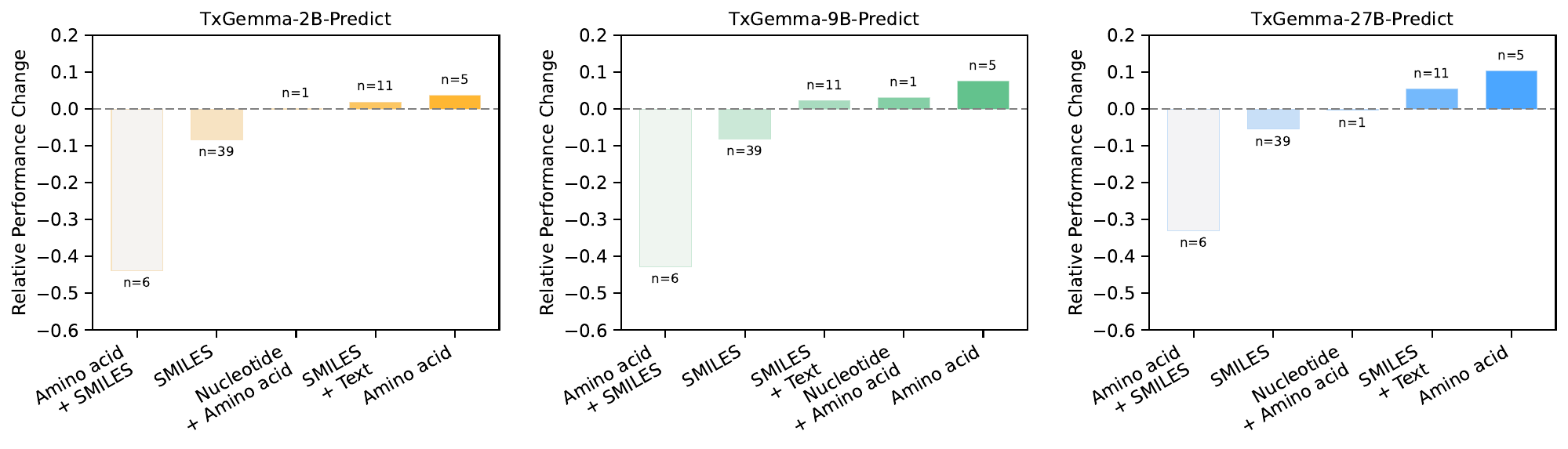}
    \vspace{-6pt}
    \caption{\textbf{Ability to combine SMILES and text is independent of model size.} Median relative change of \ourmodelpredictlargest, \ourmodelpredictnine and \ourmodelpredictsmall performance from SOTA for tasks grouped by feature type. The signs were reversed for MAE and MSE metrics because lower MAE and MSE values correspond to better performances. The number of tasks in each feature type is displayed over each bar. In all models, over 90\% of tasks had a median relative performance change greater than -0.2, and SMILES + Text consistently outperformed SOTA.}
    \label{fig:txgeamma-featuretype}
\end{figure}

\begin{figure}
    \centering
    \includegraphics[width=1\textwidth]{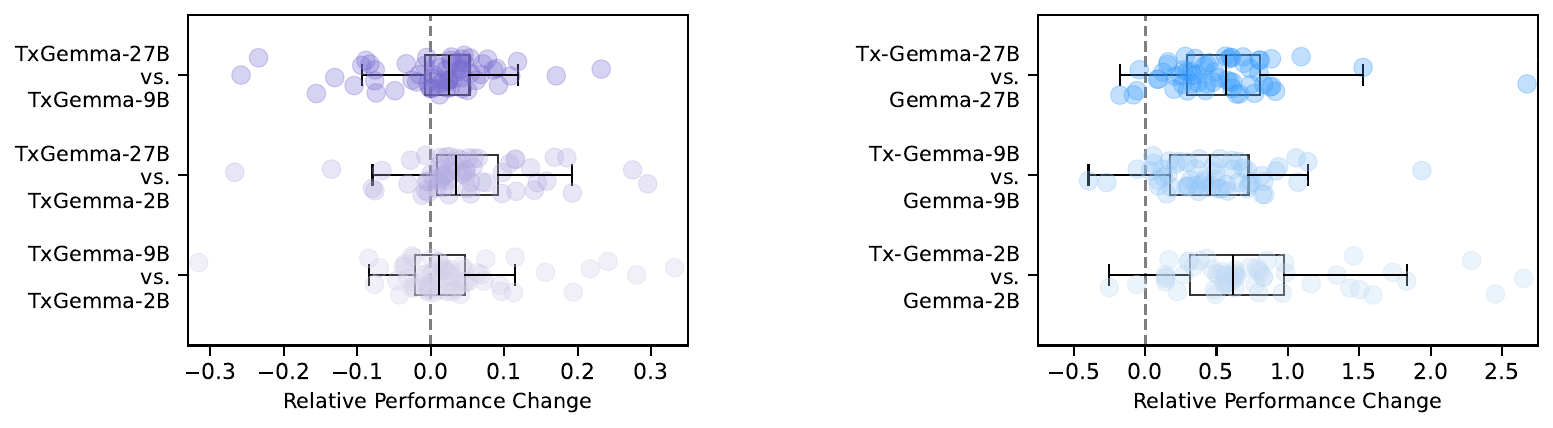}
    \vspace{-6pt}
    \caption{\textbf{Ablations of model sizes and model adaptations.} (\textbf{left}) Relative performance changes for pairwise comparisons of \ourmodelpredict models (\ourmodelpredictsmall, \ourmodelpredictnine, \ourmodelpredictlargest). (\textbf{right}) Relative performance changes of \ourmodel models compared to their respective base models.}
    \label{fig:size_ablation}
\end{figure}

\begin{figure}
    \centering
    \includegraphics[width=0.4\textwidth]{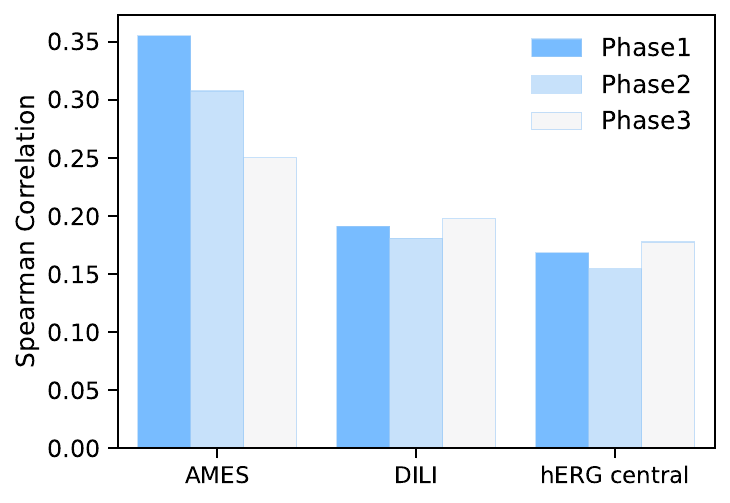}
    \vspace{6pt}
    \caption{\textbf{\ourmodel predictions show correlations between toxicity and clinical trial approval.} Spearman correlation coefficients between toxicity predictions (measured by AMES, DILI, and hERG central) and clinical trial predictions (measured by Phase1, Phase2, and Phase3) on a set of PubChem molecules.}
    \label{fig:toxtop-correlation}
\end{figure}

\cref{fig-sup:comparison-reletive-mix} compares \ourmodelpredictlargest with previous SOTA models, taking into account that \ouroldmodelm achieved SOTA performance on many tasks. We provide detailed results tables for binary classification tasks in \cref{tab:binary_results} (comparing against specialist SOTA and base models) and \cref{tab:binary_results_chat_and_txllm} (comparing against \ourmodelconverse and \ouroldmodel), and for regression and generation tasks in \cref{tab:regression_generation_results} (comparing against specialist SOTA and base models) and \cref{tab:regression_generation_chat_and_txllm} (comparing against \ourmodelconverse and \ouroldmodel). \cref{tab:binary_results_commercial,tab:regression_generation_commercial} list the performances of released \ourmodel models trained only on datasets with commercial licenses. \cref{fig-sup:llasmol,fig-sup:mole} compares \ourmodelpredictlargest with LlaSMol and MolE, models specialized for small molecules, on small molecule tasks. \cref{fig:contamination-percent-decoy} plots the percentage of tasks that contain contaminated datapoints overlapping with the \basemodel pretraining data, the percent of contaminated datapoints for these tasks, and \cref{fig:contamination-performance} shows the results of \ourmodelpredictlargest after filtering contaminated datapoints out. We observe that most tasks have no contamination, and filtering these datapoints out does not negatively impact \ourmodelpredictlargest performance. \cref{fig:txgeamma-featuretype} plots performances for particular feature types across multiple model sizes, showing that the integration of SMILES strings and textual information is consistent. \cref{fig:size_ablation} plots performances over all tasks for comparisons of model size and domain fine-tuning, showing that these variables are significant. \cref{fig:toxtop-correlation} shows that \ourmodelpredictlargest toxicity and clinical trial approval predictions are correlated, likely because toxicity in an important component of trial approval. \cref{fig:speedtest} plots the inference speed, normalized by the number of chips used for serving, for all model sizes.

\subsection{Conversing with \ourmodelpredictlargest and \ourmodelconverselargest} 
\label{sec:appendix_converse_results}
\begin{figure}[!ht]
    \centering
    \includegraphics[width=0.78\textwidth]{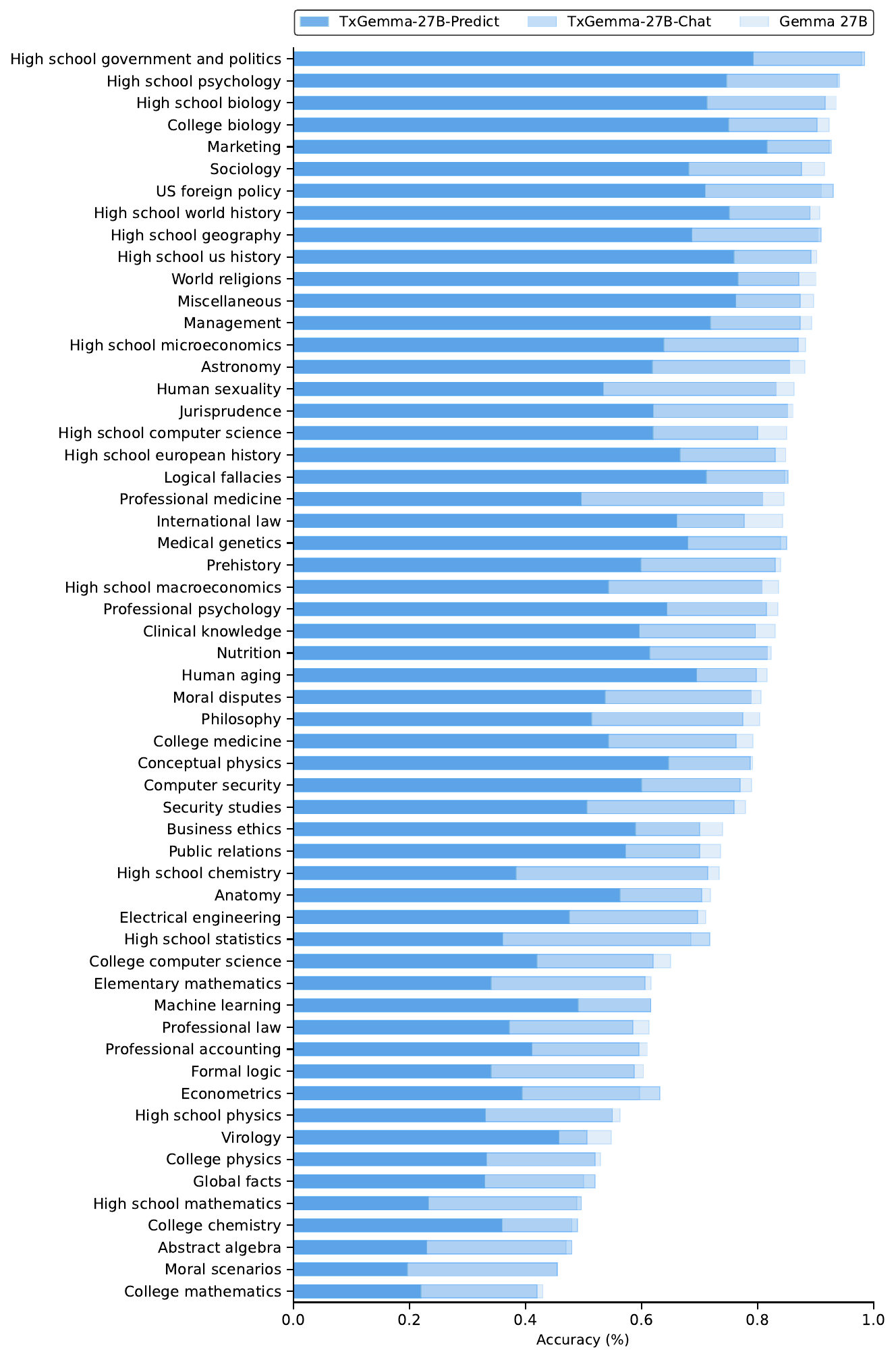}
    \vspace{0pt}
    \caption{\small{\textbf{Performance of \ourmodel and \basemodel models on MMLU.} Score of \ourmodelconverselargest, \ourmodelpredictlargest, and \basemodellargest on each subject of the MMLU benchmark~\cite{hendrycks2020measuring}, a comprehensive suite of 57 diverse tasks spanning mathematics, history, computer science, law, \etc. This benchmark evaluates knowledge, reasoning, and problem-solving abilities across a wide range of academic subjects, providing a robust measure of overall language understanding. \ourmodelconverselargest (73.87\%) performs slightly below \basemodellargest (75.38\%) but significantly better than \ourmodelpredictlargest (53.60\%), suggesting that domain fine-tuning purely on therapeutic data can cause loss of information learned during base model training, but simultaneously training with general instruction-tuning data can compensate for this.}}
    \label{fig-sup:comparison-mmlu}
\end{figure}

\begin{figure}
    \centering
    \includegraphics[width=0.9\textwidth]{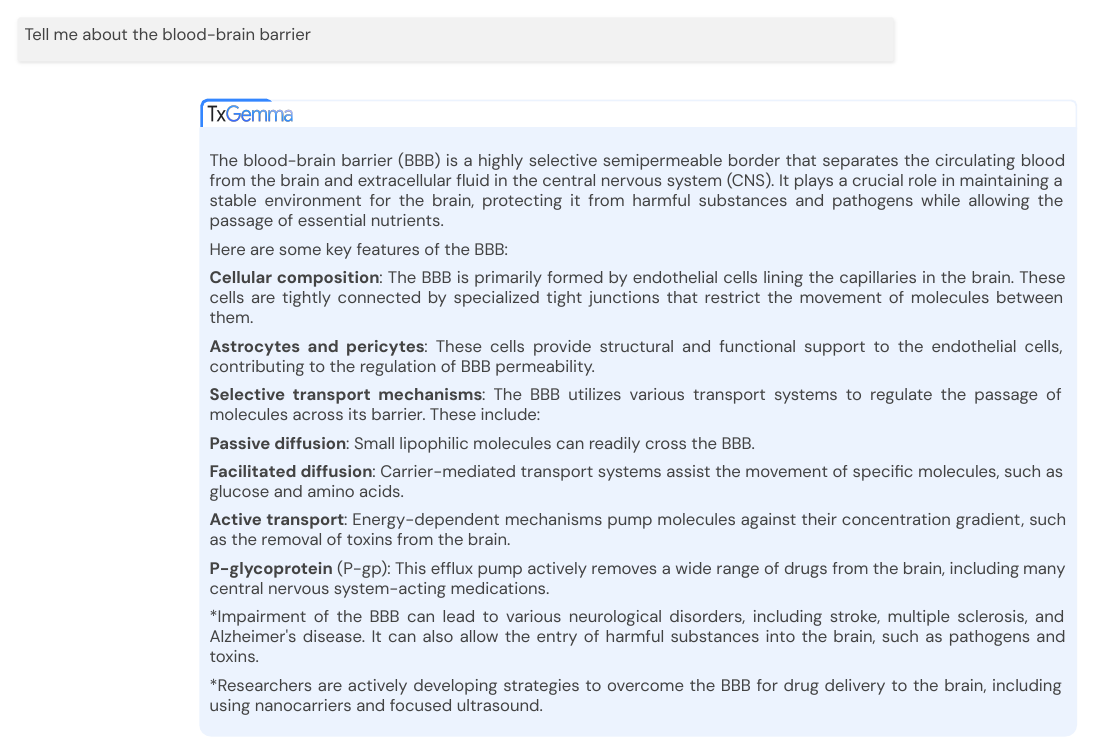}
    \vspace{6pt}
    \caption{\small{\textbf{Models can converse about general topics after fine-tuning only on the processed TDC data.} Example of a dialogue with \ourmodelpredictlargest. When asked a question that is not in a processed TDC data format, the model is able to respond coherently.}}
    \label{fig-sup:27b_nontdc_prompts}
\end{figure}

\begin{figure}
    \centering
    \includegraphics[width=0.75\textwidth]{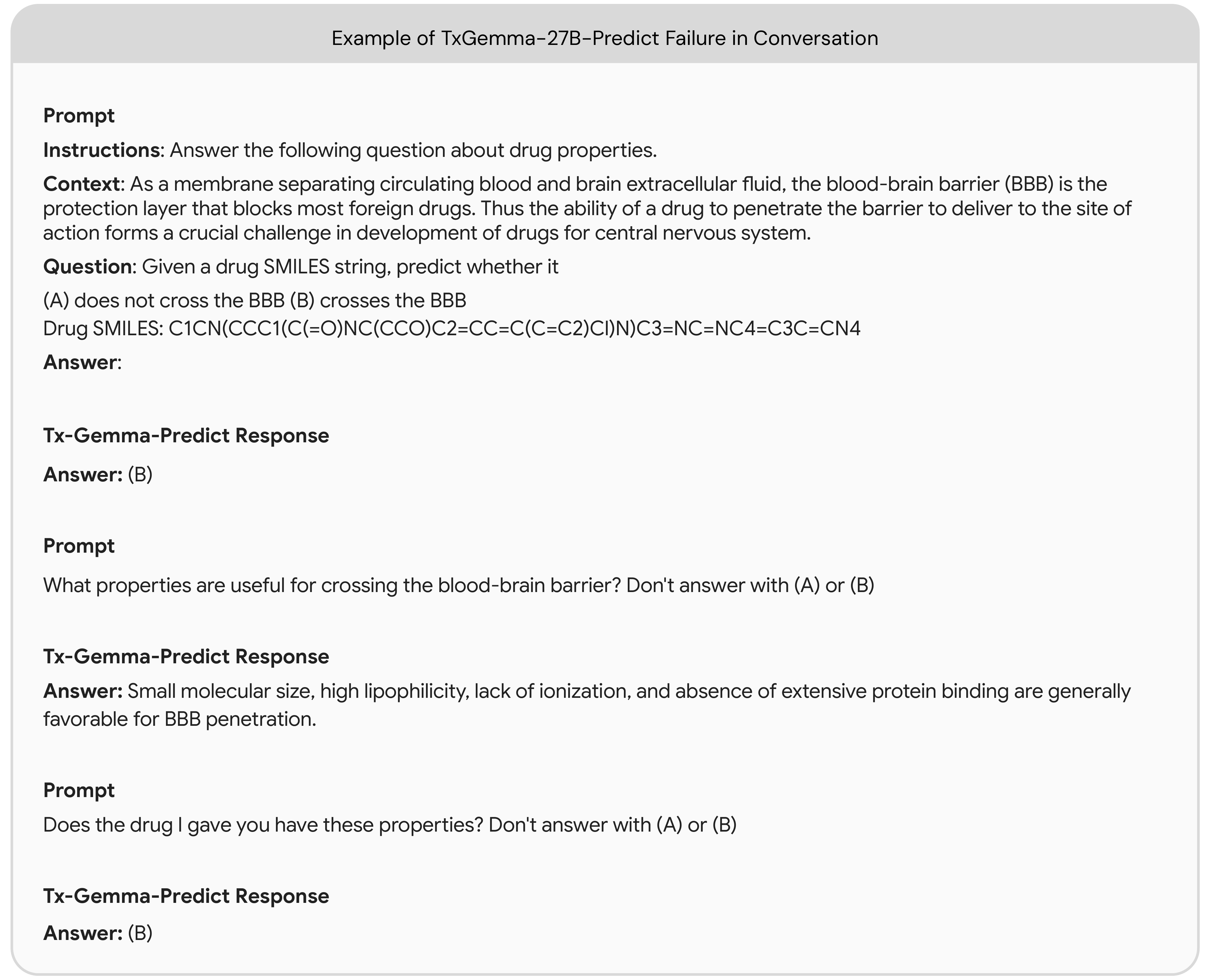}
    \vspace{6pt}
    \caption{\small{\textbf{Predictive model struggles to hold a conversation about processed TDC data prompts after fine-tuning only on this data.} Example of a multi-turn dialogue with \ourmodelpredictlargest. The initial turn is a question in the processed TDC data format. After the initial turn, the model is prompted to describe some properties associated with the problem of crossing the blood-brain barrier. An additional instruction to avoid answering with (A) or (B) is necessary. The model is able to provide some answers in natural language, but another turn leads to an inappropriate multiple choice answer.}}
    \label{fig-sup:27b-predict-conv-failure}
\end{figure}

\begin{figure}
    \centering
    \includegraphics[width=0.95\textwidth]{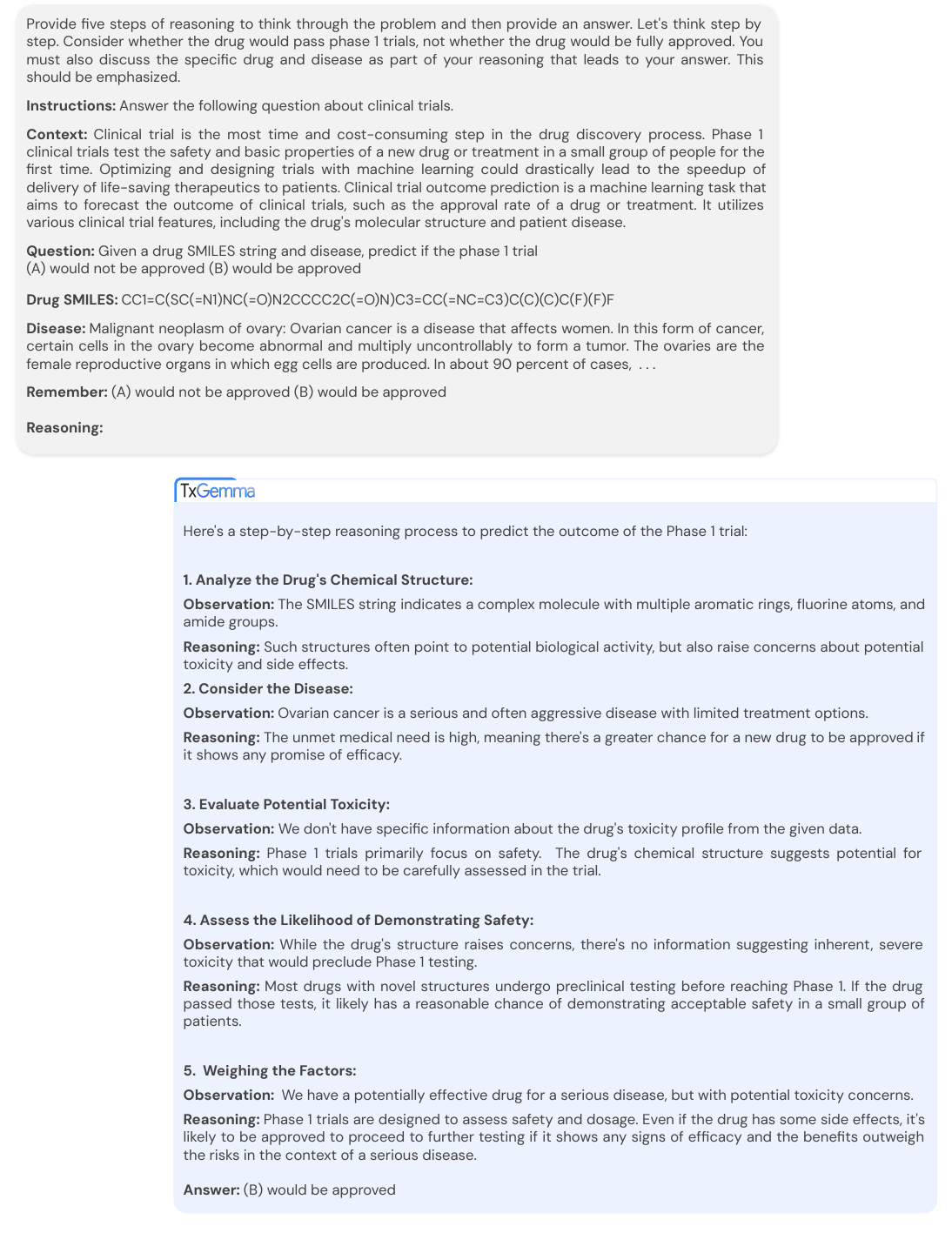}
    \vspace{0pt}
    \caption{\small{\textbf{Enabling reasoning with \ourmodelconverse on challenging tasks.} Example of prompting \ourmodelconverselargest to explain its prediction for predicting clinical trial approval, a relatively unintuitive task. A prefix is added before the prompt to provide instructions for reasoning, and a reminder is added at the end so the model correctly specifies the option corresponding to its desired answer. Lastly, the ``Answer'' text is changed to ``Reasoning:'' to enable reasoning steps. The reasoning provided by the model is not comprehensive but can provide useful insights into the drug-disease interaction.}}
    \label{fig-sup:txgemma-convo-trial}
\end{figure}

\cref{fig-sup:27b_nontdc_prompts} illustrates an example of providing a prompt to \ourmodelpredictlargest that is not in the processed data format. \ourmodelpredictlargest is able to provide a coherent response in a manner similar to the general LLMs. \cref{fig-sup:27b-predict-conv-failure} illustrates an example of first providing a prompt to \ourmodelpredictlargest in the processed format and asking follow-up questions in subsequent turns. In the second turn, instructing the model to not in the processed data format is able to elicit a reasonable but succinct response. However, the third turn leads to the model answering in the processed data format, highlighting the difficulty of multi-turn dialogue after training only on the processed TDC data. \cref{fig-sup:comparison-mmlu} plots the performance of \ourmodelconverselargest on the MMLU benchmark in comparison with both \basemodellargest and \ourmodelpredictlargest. \ourmodelconverselargest performs similarly to \basemodellargest on MMLU while \ourmodelpredictlargest scores much lower. \cref{fig-sup:txgemma-convo-trial} shows an example of using a specific prompting structure with \ourmodelconverselargest to elicit reasoning on a more challenging task of clinical trial approval. If this prompting structure is not used, the model refuses to provide reasoning.

\subsection{\ouragent Tool Use Analysis}\label{sec:appendix_agent_tooluse}

\begin{figure}
    \centering
    \includegraphics[width=0.85\textwidth]{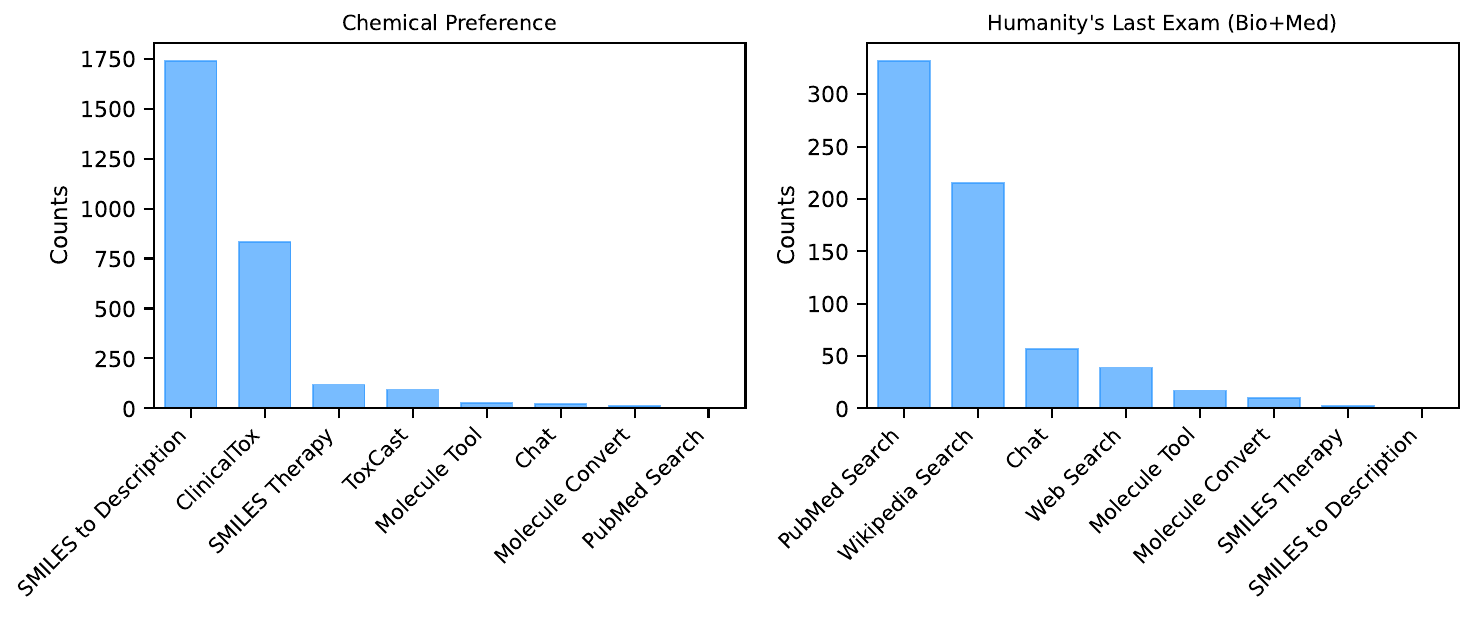}
    \vspace{0pt}
    \caption{\small{\textbf{Breakdown of tool-usage frequency for Chemical Preference dataset and HLE dataset}. \ouragent adapts its tool usage to reason effectively about different tasks. For Chemical Preference, which requires evaluating drug candidates, the system correctly invokes tools for molecular characterization and safety assessment, such as SMILES description and toxicity prediction. For the Bio+Med task, focused on complex biomedical questions, the agent prioritizes PubMed and Wikipedia, demonstrating reliance on broad knowledge retrieval and synthesis.}}
    \label{fig-sup:tool-use-frequency}
\end{figure}

\begin{figure}
    \centering
    \includegraphics[width=1.0\textwidth]{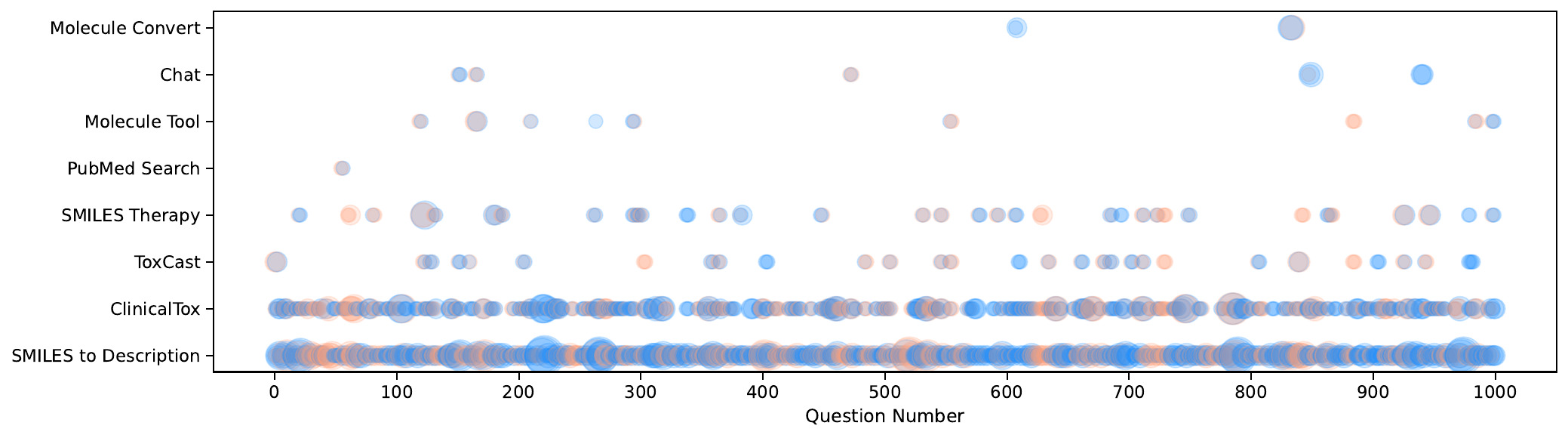}
    \vspace{0pt}
    \caption{\small{\textbf{Breakdown of tool-usage per question in chemical preference dataset}. Marker size represents usage count and corresponds to the number of uses per each tool; blue indicates accuracy increase, light red indicates decrease associated with each tool per question. We observe questions involve up to 8 tool calls. High usage of SMILES description and toxicity prediction correlates with improved performance. This demonstrates \ouragent's adaptive tool selection to meet task requirements and improved performance.}}
    \label{fig-sup:tool-use-per-question}
\end{figure}

\cref{fig-sup:tool-use-frequency} shows the tool usage frequency for different benchmarks, illustrating that \ouragent dynamically adjusts it tool usage to suit the problem. \cref{fig-sup:tool-use-per-question} shows the most frequent tools used per question for chemical preference questions, showing consistent usage of molecule-based tools.

\subsection{Proof-of-concept use of \ourmodel for end-to-end therapeutic development}\label{sec:ovarian_cancer}

\begin{figure}[h]
    \centering
    \includegraphics[width=1\textwidth]{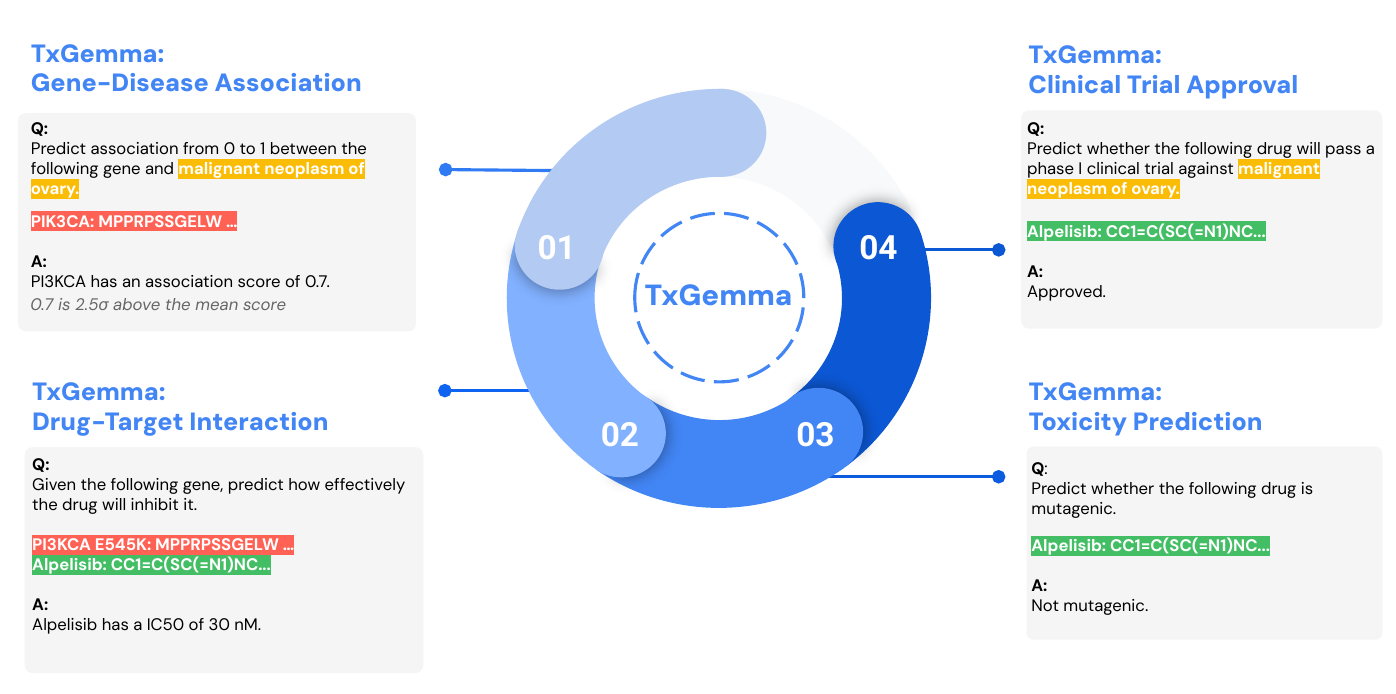}
    \vspace{6pt}
    \caption{\textbf{Proof-of-concept example of applying \ourmodel to end-to-end therapeutic development.} \ourmodel is used to suggest a therapeutic for ovarian cancer by first identifying PIK3CA as an associated gene target from a list of possible genes. Then, from a list of candidate therapeutics, \ourmodel predicts that alpelisib (a molecule previously unseen to \ourmodel that has shown activity against ovarian cancer and is approved for breast cancer) would bind the E545K mutant of PIK3CA, that it would not be toxic/mutagenic, and that it would be approved in a clinical trial. Note that this example serves as a proof-of-concept demonstration and does not account for all aspects of efficacy, toxicity, or trial approval. Rigorous experimental validation of \ourmodel predictions to completely new therapeutics is also a critical step to evaluating \ourmodel and remains an area of future work.}
    \label{fig:end_to_end}
\end{figure}

In \cref{fig:end_to_end}, we illustrate a simplified example of how \ourmodel might be helpful in identifying a drug for ovarian cancer. In this example, we chose to directly prompt \ourmodel, rather than using \ouragent, to strictly isolate potential information leakage introduced by web search, which is outside of our training data. This approach allows us to examine the model's inherent capabilities, though we acknowledge that a full agent-based workflow is a plausible extension. 

We initially use the DisGeNET prompt to identify an ovarian cancer-associated target gene from a short list of genes including PIK3CA, JAK2, RET. \ourmodelpredictlargest predicts that PIK3CA, a gene not found in the training set which is known to be mutated in ovarian cancer \cite{kuo2009frequent}, has an association score of 0.7 with ovarian cancer. This association score is nearly 2.5 standard deviations above the mean score ($\mu=0.37$, $\sigma=0.13$), indicating a strong association. JAK2 and RET share an association score of 0.3 which is below the mean score. We then used \ourmodelpredictlargest to select a potential therapeutic from a molecule shortlist, prioritizing predicted IC$_{50}$ against the E545K mutant (an oncogenic mutation~\cite{leontiadou2018insights}), toxicity, and clinical trial success. Our manually curated shortlist of drugs, unseen to the model during training, include two existing cancer therapies including alpelisib and afatinib and a novel molecule which we randomly generated.  Both afatinib (1.02 $\mu$M IC$_{50}$) and the novel molecule (10.2 $\mu$M IC$_{50}$) exhibit high predicted IC$_{50}$ values, suggesting weak inhibition. However, alpelisib has a predicted IC$_{50}$ of 30~nM, suggestive of potent inhibition and relatively close to the experimental value of 5~nM suggested by \citet{chen2023p110alpha,fritsch2014characterization}. \ourmodelpredictlargest also predicts that alpelisib is not mutagenic and would pass a phase 1 clinical trial for ovarian cancer. This iterative evaluation also corroborated by existing evidence: alpelisib is approved for breast cancer \cite{narayan2021fda} and has shown activity in ovarian cancer~\cite{passarelli2024alpelisib,thibault2025pi3kalpha,hu2020dual}.

This workflow demonstrates a proof-of-concept for \ourmodel's application in automating and optimizing therapeutic selection. We anticipate an agentic system capable of generating comprehensive lists of potential therapies and gene-disease associations paired with \ourmodel would enable rapid prioritization and filtering, helping in reducing the candidate pool and accelerating the transition to preclinical studies.  However, it's crucial to acknowledge the limitations of this demonstration. Clinical trial predictions are limited to Phase 1 success, and mutagenicity predictions do not encompass all aspects of small molecule toxicity. Future work should include experimental validation of \ourmodel predictions and consideration of additional toxicity factors, such as hematologic toxicity, which were not included in our data.

\newcolumntype{P}[1]{>{\centering\arraybackslash}p{#1}}
\begin{table}[htbp]
\centering
\scriptsize
\caption{\textbf{Model performance on binary classification tasks.} \ourmodelpredict and \basemodel performances compared with specialist SOTA for each binary classification task, along with the metric type.}
\label{tab:binary_results}
\renewcommand{\arraystretch}{1.1}
\centerline{
\footnotesize
\begin{tabular}{p{4.75cm}|P{1.2cm}P{1.4cm}P{1.cm}P{1.cm}P{1.cm}P{1.5cm}P{1.5cm}P{1.6cm}}
\toprule
Task Name & Metric & Specialist SOTA & \basemodelsmall & \basemodelnine & \basemodellargest & \ourmodelpredictsmall & \ourmodelpredictnine & \ourmodelpredictlargest \\
\midrule
AMES & AUROC & 0.871 \cite{turon2023first} & 0.487 & 0.605 & 0.508 & 0.796 & 0.798 & 0.816 \\
BBB Martins & AUROC & 0.915 \cite{fonteno2023predicting} & 0.250 & 0.645 & 0.546 & 0.864 & 0.874 & 0.907 \\
Bioavailability Ma & AUROC & 0.748 \cite{bera2022simgcn} & 0.479 & 0.584 & 0.579 & 0.715 & 0.655 & 0.696 \\
CYP1A2 Veith & AUPRC & 0.900 \cite{plonka2021cyplebrity} & 0.388 & 0.533 & 0.562 & 0.910 & 0.916 & 0.922 \\
CYP2C19 Veith & AUROC & 0.890 \cite{plonka2021cyplebrity} & 0.456 & 0.595 & 0.619 & 0.905 & 0.906 & 0.899 \\
CYP2C9 Substrate CarbonMangels & AUPRC & 0.441 \cite{turon2023first} & 0.293 & 0.336 & 0.367 & 0.457 & 0.468 & 0.427 \\
CYP2C9 Veith & AUPRC & 0.839 \cite{hu2019strategies} & 0.283 & 0.374 & 0.417 & 0.801 & 0.799 & 0.798 \\
CYP2D6 Substrate CarbonMangels & AUPRC & 0.736 \cite{hu2019strategies} & 0.233 & 0.329 & 0.386 & 0.605 & 0.603 & 0.706 \\
CYP2D6 Veith & AUPRC & 0.739 \cite{hu2019strategies} & 0.145 & 0.166 & 0.185 & 0.637 & 0.664 & 0.681 \\
CYP3A4 Substrate CarbonMangels & AUROC & 0.662 \cite{huang2020deeppurpose} & 0.514 & 0.585 & 0.596 & 0.669 & 0.622 & 0.690 \\
CYP3A4 Veith & AUPRC & 0.904 \cite{hu2019strategies} & 0.427 & 0.531 & 0.535 & 0.844 & 0.839 & 0.854 \\
Carcinogens Lagunin & Accuracy & 0.770 \cite{lagunin2009computer} & 0.250 & 0.286 & 0.339 & 0.821 & 0.839 & 0.857 \\
ClinTox & AUROC & 0.948 \cite{li2021trimnet} & 0.437 & 0.482 & 0.424 & 0.810 & 0.831 & 0.888 \\
DILI & AUROC & 0.925 \cite{turon2023first} & 0.320 & 0.651 & 0.627 & 0.875 & 0.848 & 0.887 \\
HIA Hou & AUROC & 0.988 \cite{huang2022unified} & 0.257 & 0.932 & 0.783 & 0.937 & 0.967 & 0.988 \\
HIV & AUROC & 0.851 \cite{li2017learning} & 0.491 & 0.495 & 0.537 & 0.737 & 0.734 & 0.764 \\
HuRI & AUPRC & 0.724 \cite{raimondi2021novel} & 0.496 & 0.484 & 0.526 & 0.751 & 0.779 & 0.799 \\
MHC1 IEDB IMGT Nielsen & AUROC & 0.986 \cite{gfeller2023improved} & 0.498 & 0.504 & 0.517 & 0.910 & 0.927 & 0.929 \\
MHC2 IEDB Jensen & AUROC & 0.940 \cite{motmaen2023peptide} & 0.498 & 0.526 & 0.544 & 0.812 & 0.850 & 0.851 \\
PAMPA NCATS & AUROC & 0.900 \cite{siramshetty2021validating} & 0.465 & 0.583 & 0.544 & 0.642 & 0.671 & 0.705 \\
Pgp Broccatelli & AUROC & 0.935 \cite{turon2023first} & 0.416 & 0.670 & 0.497 & 0.900 & 0.911 & 0.936 \\
SARSCOV2 3CLPro Diamond & AUROC & 0.800 \cite{haneczok2021machine} & 0.301 & 0.388 & 0.477 & 0.733 & 0.708 & 0.769 \\
SARSCoV2 Vitro Touret & AUROC & 0.640 \cite{lie2021covid} & 0.568 & 0.611 & 0.479 & 0.650 & 0.668 & 0.598 \\
SAbDab Chen & AUPRC & 0.510 \cite{chen2020predicting} & 0.532 & 0.696 & 0.701 & 0.676 & 0.807 & 0.767 \\
Skin Reaction & AUROC & 0.840 \cite{alves2015predicting} & 0.429 & 0.546 & 0.493 & 0.671 & 0.648 & 0.708 \\
Tox21 & AUROC & 0.961 \cite{shermukhamedov2023structure} & 0.358 & 0.436 & 0.497 & 0.881 & 0.896 & 0.893 \\
ToxCast & AUROC & 0.777 \cite{li2021trimnet} & 0.485 & 0.512 & 0.558 & 0.784 & 0.767 & 0.800 \\
butkiewicz & AUROC & 0.840 \cite{vu2019bcl} & 0.457 & 0.491 & 0.491 & 0.791 & 0.772 & 0.831 \\
hERG & AUROC & 0.874 \cite{bera2022simgcn} & 0.538 & 0.639 & 0.500 & 0.876 & 0.881 & 0.884 \\
hERG Karim & Accuracy & 0.770 \cite{karim2021cardiotox} & 0.529 & 0.532 & 0.522 & 0.778 & 0.794 & 0.774 \\
herg central & AUROC & 0.860 \cite{korotcov2017comparison} & 0.481 & 0.511 & 0.517 & 0.880 & 0.861 & 0.896 \\
miRTarBase & Accuracy & 0.804 \cite{wong2020mipdh} & 0.498 & 0.501 & 0.498 & 0.805 & 0.829 & 0.801 \\
phase1 & AUROC & 0.576 \cite{fu2022hint} & 0.562 & 0.562 & 0.553 & 0.642 & 0.635 & 0.622 \\
phase2 & AUROC & 0.645 \cite{fu2022hint} & 0.543 & 0.571 & 0.531 & 0.665 & 0.668 & 0.676 \\
phase3 & AUROC & 0.723 \cite{fu2022hint} & 0.559 & 0.567 & 0.559 & 0.731 & 0.729 & 0.739 \\
weber & AUROC & 0.870 \cite{weber2021titan} & 0.466 & 0.586 & 0.469 & 0.730 & 0.727 & 0.749 \\
\bottomrule
\end{tabular}
}
\end{table}
\begin{table}[htbp]
\centering
\small
\caption{\textbf{Model performance on regression and generation tasks.} \ourmodelpredict and \basemodel performances compared with specialist SOTA for each regression and generation task, along with the metric type. Tasks for which we did not find a specialist SOTA value are indicated with N/A.}
\label{tab:regression_generation_results}
\renewcommand{\arraystretch}{1.1}
\centerline{
\footnotesize
\begin{tabular}{p{3.5cm}|P{1.2cm}P{1.4cm}P{1.cm}P{1.cm}P{1.cm}P{1.5cm}P{1.5cm}P{1.6cm}}
\toprule
Task Name & Metric & Specialist SOTA & \basemodelsmall & \basemodelnine & \basemodellargest & \ourmodelpredictsmall & \ourmodelpredictnine & \ourmodelpredictlargest \\
\midrule
BindingDB Patent & PCC & 0.588 \cite{lam2023otter} & -0.066 & -0.039 & 0.030 & 0.422 & 0.524 & 0.538 \\
BindingDB ic50 & Spearman & 0.637 \cite{kinnings2011machine} & 0.001 & 0.002 & 0.044 & 0.399 & 0.398 & 0.445 \\
BindingDB kd & PCC & 0.712 \cite{kalemati2023bicomp} & 0.197 & -0.009 & 0.119 & 0.352 & 0.370 & 0.456 \\
BindingDB ki & PCC & 0.840 \cite{wei2021deeppla} & -0.018 & -0.053 & -0.027 & 0.661 & 0.737 & 0.676 \\
Buchwald Hartwig & PCC & 0.786 \cite{probst2022reaction} & 0.528 & 0.636 & 0.684 & 0.861 & 0.915 & 0.910 \\
Caco2 Wang & MAE & 0.285 \cite{huang2022unified} & 1.057 & 0.533 & 0.618 & 0.476 & 0.373 & 0.401 \\
Clearance Hepatocyte AZ & Spearman & 0.440 \cite{rivera2024silico} & 0.141 & 0.163 & 0.214 & 0.353 & 0.338 & 0.259 \\
Clearance Microsome AZ & Spearman & 0.625 \cite{huang2022unified} & 0.239 & 0.325 & 0.294 & 0.468 & 0.623 & 0.462 \\
DAVIS & MSE & 0.219 \cite{pei2023breaking} & 2.705 & 9.054 & 4.473 & 0.601 & 0.587 & 0.555 \\
DisGeNET & MAE & N/A & 0.294 & 0.295 & 0.277 & 0.057 & 0.054 & 0.054 \\
DrugComb Bliss & MAE & 4.560 \cite{xia2018predicting} & 8.213 & 7.413 & 6.456 & 4.230 & 4.337 & 4.156 \\
DrugComb CSS & MAE & 16.858 \cite{xia2018predicting} & 36.847 & 33.837 & 22.614 & 15.752 & 16.480 & 15.000 \\
DrugComb HSA & MAE & 4.453 \cite{xia2018predicting} & 7.458 & 7.365 & 6.670 & 4.231 & 4.335 & 4.209 \\
DrugComb Loewe & MAE & 9.184 \cite{xia2018predicting} & 13.873 & 13.369 & 14.731 & 17.342 & 18.665 & 17.336 \\
DrugComb ZIP & MAE & 4.027 \cite{xia2018predicting} & 8.588 & 6.226 & 5.404 & 3.950 & 3.904 & 3.807 \\
GDSC1 & PCC & 0.860 \cite{lind2019predicting} & -0.041 & 0.073 & 0.093 & 0.876 & 0.545 & 0.892 \\
GDSC2 & PCC & 0.860 \cite{lind2019predicting} & -0.043 & -0.037 & 0.086 & 0.824 & 0.539 & 0.912 \\
Half Life Obach & Spearman & 0.547 \cite{euclia2023half} & 0.288 & 0.284 & 0.485 & 0.386 & 0.494 & 0.458 \\
KIBA & MSE & 0.154 \cite{pei2023breaking} & 2.887 & 1.925 & 2.016 & 0.588 & 0.548 & 0.633 \\
LD50 Zhu & MAE & 0.552 \cite{huang2022unified} & 1.971 & 0.896 & 0.874 & 0.710 & 0.630 & 0.628 \\
Leenay & Spearman & 0.740 \cite{leenay2019large} & 0.085 & 0.091 & 0.146 & 0.097 & 0.067 & 0.276 \\
Lipophilicity AstraZeneca & MAE & 0.467 \cite{yang2019analyzing} & 1.506 & 1.207 & 1.032 & 0.610 & 0.565 & 0.539 \\
OncoPolyPharmacology & PCC & 0.730 \cite{preuer2018deepsynergy} & -0.040 & 0.064 & 0.072 & 0.473 & 0.518 & 0.540 \\
PPBR AZ & MAE & 7.788 \cite{yang2019analyzing} & 10.836 & 9.768 & 9.879 & 9.266 & 8.889 & 9.029 \\
Protein SAbDab & MAE & N/A & 1.280 & 1.170 & 1.163 & 1.066 & 1.106 & 1.210 \\
Solubility AqSolDB & MAE & 0.761 \cite{yang2019analyzing} & 4.214 & 2.549 & 3.096 & 0.961 & 0.868 & 0.821 \\
TAP & MAE & N/A & 5.008 & 4.241 & 3.958 & 5.301 & 4.473 & 4.280 \\
USPTO & Accuracy & 0.415 \cite{zheng2019predicting} & 0.000 & 0.001 & 0.000 & 0.287 & 0.097 & 0.084 \\
USPTO Yields & PCC & 0.361 \cite{probst2022reaction} & -0.015 & 0.026 & 0.064 & 0.011 & 0.031 & 0.395 \\
VDss Lombardo & Spearman & 0.627 \cite{boral2022accountable} & 0.100 & 0.413 & 0.354 & 0.564 & 0.607 & 0.560 \\
\bottomrule
\end{tabular}
}
\end{table}

\newcolumntype{P}[1]{>{\centering\arraybackslash}p{#1}}
\begin{table}[htbp]
\centering
\scriptsize
\caption{\textbf{Model performance on binary classification tasks.} \ourmodelpredict, \ourmodelconverse, and \ouroldmodel performances for each binary classification task, along with the metric type.}
\label{tab:binary_results_chat_and_txllm}
\renewcommand{\arraystretch}{1.1}
\centerline{
\footnotesize
\begin{tabular}{p{4.75cm}|P{1.4cm}P{1.5cm}P{1.6cm}P{1.5cm}P{1.5cm}P{1.4cm}P{1.42cm}}
\toprule
Task Name & Metric & \ourmodelpredictnine & \ourmodelpredictlargest & \ourmodelconversenine & \ourmodelconverselargest & \ouroldmodels & \ouroldmodelm \\
\midrule
AMES & AUROC & 0.798 & 0.816 & 0.721 & 0.733  & 0.785 & 0.786 \\
BBB Martins & AUROC & 0.874 & 0.907 & 0.811 & 0.861  & 0.805 & 0.882 \\
Bioavailability Ma & AUROC & 0.655 & 0.696 & 0.620 & 0.659  & 0.605 & 0.702 \\
CYP1A2 Veith & AUPRC & 0.916 & 0.922 & 0.839 & 0.823  & 0.906 & 0.914 \\
CYP2C19 Veith & AUROC & 0.906 & 0.899 & 0.837 & 0.828  & 0.877 & 0.895 \\
CYP2C9 Substrate CarbonMangels & AUPRC & 0.468 & 0.427 & 0.382 & 0.427  & 0.403 & 0.436 \\
CYP2C9 Veith & AUPRC & 0.799 & 0.798 & 0.667 & 0.682  & 0.750 & 0.788 \\
CYP2D6 Substrate CarbonMangels & AUPRC & 0.603 & 0.706 & 0.549 & 0.700  & 0.643 & 0.600 \\
CYP2D6 Veith & AUPRC & 0.664 & 0.681 & 0.504 & 0.435  & 0.605 & 0.659 \\
CYP3A4 Substrate CarbonMangels & AUROC & 0.622 & 0.690 & 0.642 & 0.666  & 0.637 & 0.647 \\
CYP3A4 Veith & AUPRC & 0.839 & 0.854 & 0.749 & 0.750  & 0.800 & 0.840 \\
Carcinogens Lagunin & Accuracy & 0.839 & 0.857 & 0.893 & 0.911  & 0.857 & 0.786 \\
ClinTox & AUROC & 0.831 & 0.888 & 0.711 & 0.637  & 0.818 & 0.863 \\
DILI & AUROC & 0.848 & 0.887 & 0.688 & 0.766  & 0.727 & 0.882 \\
HIA Hou & AUROC & 0.967 & 0.988 & 0.872 & 0.897  & 0.942 & 0.990 \\
HIV$^*$ & AUROC & 0.734 & 0.764 & 0.612 & 0.582  & 0.686 & 0.732 \\
HuRI & AUPRC & 0.779 & 0.799 & 0.628 & 0.621  & 0.705 & 0.753 \\
MHC1 IEDB IMGT Nielsen & AUROC & 0.927 & 0.929 & 0.875 & 0.825  & 0.913 & 0.907 \\
MHC2 IEDB Jensen & AUROC & 0.850 & 0.851 & 0.724 & 0.683  & 0.781 & 0.863 \\
PAMPA NCATS & AUROC & 0.671 & 0.705 & 0.735 & 0.664  & 0.646 & 0.668 \\
Pgp Broccatelli & AUROC & 0.911 & 0.936 & 0.899 & 0.912  & 0.909 & 0.939 \\
SARSCOV2 3CLPro Diamond & AUROC & 0.708 & 0.769 & 0.699 & 0.722  & 0.755 & 0.712 \\
SARSCoV2 Vitro Touret & AUROC & 0.668 & 0.598 & 0.503 & 0.506  & 0.512 & 0.601 \\
SAbDab Chen & AUPRC & 0.807 & 0.767 & 0.702 & 0.719  & 0.390 & 0.473 \\
Skin Reaction & AUROC & 0.648 & 0.708 & 0.638 & 0.543  & 0.564 & 0.615 \\
Tox21 & AUROC & 0.896 & 0.893 & 0.807 & 0.797  & 0.858 & 0.882 \\
ToxCast & AUROC & 0.767 & 0.800 & 0.754 & 0.734  & 0.779 & 0.792 \\
butkiewicz & AUROC & 0.772 & 0.831 & 0.629 & 0.619  & 0.574 & 0.566 \\
hERG & AUROC & 0.881 & 0.884 & 0.830 & 0.832  & 0.879 & 0.909 \\
hERG Karim & Accuracy & 0.794 & 0.774 & 0.657 & 0.668  & 0.724 & 0.745 \\
herg central & AUROC & 0.861 & 0.896 & 0.830 & 0.807  & 0.880 & 0.888 \\
miRTarBase & Accuracy & 0.829 & 0.801 & 0.679 & 0.644  & 0.765 & 0.799 \\
phase1 & AUROC & 0.635 & 0.622 & 0.576 & 0.557  & 0.624 & 0.667 \\
phase2 & AUROC & 0.668 & 0.676 & 0.638 & 0.626  & 0.639 & 0.676 \\
phase3 & AUROC & 0.729 & 0.739 & 0.683 & 0.668  & 0.701 & 0.728 \\
weber & AUROC & 0.727 & 0.749 & 0.672 & 0.643   & 0.738 & 0.743 \\
\bottomrule
\end{tabular}
}
{\raggedright
\vspace{0.05in}
\scriptsize{
$*$ To predict whether compounds have Anti-HIV properties. \\ 
}}
\end{table}
\begin{table}[htbp]
\centering
\small
\caption{\textbf{Model performance on regression and generation tasks.} \ourmodelpredict, \ourmodelconverse, and \ouroldmodel performances for each regression and generation task, along with the metric type.}
\label{tab:regression_generation_chat_and_txllm}
\renewcommand{\arraystretch}{1.1}
\centerline{
\footnotesize
\begin{tabular}{p{3.5cm}|P{1.2cm}P{1.5cm}P{1.6cm}P{1.5cm}P{1.5cm}P{1.4cm}P{1.42cm}}
\toprule
Task Name & Metric & \ourmodelpredictnine & \ourmodelpredictlargest & \ourmodelconversenine & \ourmodelconverselargest & \ouroldmodels & \ouroldmodelm \\
\midrule
BindingDB Patent & PCC & 0.524 & 0.538 & 0.452 & 0.220   & 0.474 & 0.531 \\
BindingDB ic50 & Spearman & 0.398 & 0.445 & 0.412 & 0.362   & 0.326 & 0.311 \\
BindingDB kd & PCC & 0.370 & 0.456 & 0.162 & 0.159   & 0.317 & 0.391 \\
BindingDB ki & PCC & 0.737 & 0.676 & 0.448 & 0.211   & 0.565 & 0.726 \\
Buchwald Hartwig & PCC & 0.915 & 0.910 & 0.255 & 0.757   & 0.682 & 0.905 \\
Caco2 Wang & MAE & 0.373 & 0.401 & 0.643 & 0.398   & 0.621 & 0.432 \\
Clearance Hepatocyte AZ & Spearman & 0.338 & 0.259 & 0.197 & 0.150   & 0.256 & 0.385 \\
Clearance Microsome AZ & Spearman & 0.623 & 0.462 & 0.345 & 0.420  & 0.385 & 0.413 \\
DAVIS & MSE & 0.587 & 0.555 & 0.608 & 0.561   & 0.564 & 0.704 \\
DisGeNET & MAE & 0.054 & 0.054 & 0.066 & 0.064   & 0.059 & 0.057 \\
DrugComb Bliss & MAE & 4.337 & 4.156 & 4.502 & 4.511   & 4.425 & 4.104 \\
DrugComb CSS & MAE & 16.480 & 15.000 & 16.384 & 16.900   & 14.740 & 14.057 \\
DrugComb HSA & MAE & 4.335 & 4.209 & 4.497 & 4.520   & 4.311 & 4.118 \\
DrugComb Loewe & MAE & 18.665 & 17.336 & 16.994 & 16.914   & 17.428 & 17.381 \\
DrugComb ZIP & MAE & 3.904 & 3.807 & 4.139 & 4.141   & 4.047 & 3.777 \\
GDSC1 & PCC & 0.545 & 0.892 & 0.861 & 0.802   & 0.876 & 0.887 \\
GDSC2 & PCC & 0.539 & 0.912 & 0.864 & 0.823  & 0.896 & 0.900 \\
Half Life Obach & Spearman & 0.494 & 0.458 & 0.330 & 0.414   & 0.525 & 0.448 \\
KIBA & MSE & 0.548 & 0.633 & 0.705 & 0.852  & 0.709 & 0.548 \\
LD50 Zhu & MAE & 0.630 & 0.628 & 0.740 & 0.705  & 0.808 & 0.618 \\
Leenay & Spearman & 0.067 & 0.276 & 0.128 & 0.095   & 0.048 & 0.083 \\
Lipophilicity AstraZeneca & MAE & 0.565 & 0.539 & 0.985 & 0.842   & 0.779 & 0.587 \\
OncoPolyPharmacology & PCC & 0.518 & 0.540 & 0.359 & 0.193   & 0.418 & 0.552 \\
PPBR AZ & MAE & 8.889 & 9.029 & 11.367 & 10.895  & 11.138 & 9.108 \\
Protein SAbDab & MAE & 1.106 & 1.210 & 1.268 & 1.116  & 1.432 & 1.268 \\
Solubility AqSolDB & MAE & 0.868 & 0.821 & 1.159 & 1.133  & 0.931 & 0.987 \\
TAP & MAE & 4.473 & 4.280 & 4.859 & 4.083  & 5.075 & 4.983 \\
USPTO & Accuracy & 0.097 & 0.084 & 0.086 & 0.091  & 0.220 & 0.239 \\
USPTO Yields & PCC & 0.031 & 0.395 & 0.003 & 0.026  & 0.042 & 0.070 \\
VDss Lombardo & Spearman & 0.607 & 0.560 & 0.396 & 0.407  & 0.497 & 0.609 \\
\bottomrule
\end{tabular}
}
\end{table}

\newcolumntype{P}[1]{>{\centering\arraybackslash}p{#1}}
\begin{table}[htbp]
\centering
\scriptsize
\caption{\textbf{Model performance on binary classification tasks for models trained only on datasets with commercial licenses.} \ourmodelpredict and \ourmodelconverse performances for each binary classification task, along with the metric type.}
\label{tab:binary_results_commercial}
\renewcommand{\arraystretch}{1.1}
\centerline{
\footnotesize
\begin{tabular}{p{4.75cm}|P{1.4cm}P{1.5cm}P{1.5cm}P{1.6cm}P{1.5cm}P{1.4cm}P{1.42cm}}
\toprule
Task Name & Metric & \ourmodelpredictsmall & \ourmodelpredictnine & \ourmodelpredictlargest & \ourmodelconversenine & \ourmodelconverselargest \\
\midrule
AMES & AUROC & 0.812 & 0.803 & 0.826 & 0.723 & 0.729 \\
BBB Martins & AUROC & 0.883 & 0.849 & 0.899 & 0.832 & 0.848 \\
Bioavailability Ma & AUROC & 0.688 & 0.688 & 0.724 & 0.666 & 0.625 \\
CYP1A2 Veith & AUPRC & 0.911 & 0.914 & 0.916 & 0.862 & 0.817 \\
CYP2C19 Veith & AUROC & 0.905 & 0.897 & 0.897 & 0.844 & 0.823 \\
CYP2C9 Substrate CarbonMangels & AUPRC & 0.417 & 0.390 & 0.460 & 0.414 & 0.375 \\
CYP2C9 Veith & AUPRC & 0.787 & 0.800 & 0.793 & 0.700 & 0.685 \\
CYP2D6 Substrate CarbonMangels & AUPRC & 0.626 & 0.697 & 0.706 & 0.653 & 0.704 \\
CYP2D6 Veith & AUPRC & 0.666 & 0.662 & 0.677 & 0.517 & 0.422 \\
CYP3A4 Substrate CarbonMangels & AUROC & 0.638 & 0.680 & 0.692 & 0.644 & 0.653 \\
CYP3A4 Veith & AUPRC & 0.842 & 0.839 & 0.852 & 0.760 & 0.747 \\
Carcinogens Lagunin & Accuracy & 0.911 & 0.857 & 0.875 & 0.893 & 0.929 \\
ClinTox & AUROC & 0.917 & 0.815 & 0.884 & 0.716 & 0.595 \\
DILI & AUROC & 0.829 & 0.823 & 0.927 & 0.675 & 0.797 \\
HIA Hou & AUROC & 0.984 & 0.954 & 0.990 & 0.906 & 0.927 \\
HIV & AUROC & 0.781 & 0.730 & 0.768 & 0.641 & 0.589 \\
HuRI & AUPRC & 0.735 & 0.767 & 0.797 & 0.685 & 0.620 \\
MHC1 IEDB IMGT Nielsen & AUROC & 0.930 & 0.929 & 0.933 & 0.887 & 0.826 \\
MHC2 IEDB Jensen & AUROC & 0.855 & 0.852 & 0.855 & 0.733 & 0.682 \\
PAMPA NCATS & AUROC & 0.694 & 0.630 & 0.724 & 0.684 & 0.659 \\
Pgp Broccatelli & AUROC & 0.922 & 0.932 & 0.941 & 0.873 & 0.920 \\
SARSCOV2 3CLPro Diamond & AUROC & 0.748 & 0.799 & 0.676 & 0.716 & 0.712 \\
SARSCoV2 Vitro Touret & AUROC & 0.659 & 0.622 & 0.597 & 0.527 & 0.516 \\
SAbDab Chen & AUPRC & 0.726 & 0.745 & 0.793 & 0.523 & 0.731 \\
Skin Reaction & AUROC & 0.691 & 0.624 & 0.733 & 0.621 & 0.571 \\
Tox21 & AUROC & 0.897 & 0.893 & 0.890 & 0.818 & 0.797 \\
ToxCast & AUROC & 0.787 & 0.766 & 0.797 & 0.754 & 0.735 \\
butkiewicz & AUROC & 0.811 & 0.775 & 0.826 & 0.681 & 0.606 \\
hERG & AUROC & 0.902 & 0.890 & 0.894 & 0.855 & 0.829 \\
hERG Karim & Accuracy & 0.778 & 0.796 & 0.772 & 0.649 & 0.673 \\
herg central & AUROC & 0.890 & 0.860 & 0.892 & 0.842 & 0.805 \\
miRTarBase & Accuracy & 0.818 & 0.834 & 0.802 & 0.672 & 0.649 \\
weber & AUROC & 0.750 & 0.697 & 0.749 & 0.692 & 0.645 \\
\bottomrule
\end{tabular}
}
{\raggedright
\vspace{0.05in}
\scriptsize{
$*$ To predict whether compounds have Anti-HIV properties. \\ 
}}
\end{table}
\begin{table}[htbp]
\centering
\small
\caption{\textbf{Model performance on regression and generation tasks for models trained only on datasets with commercial licenses.} \ourmodelpredict and \ourmodelconverse performances for each regression or generation task, along with the metric type.}
\label{tab:regression_generation_commercial}
\renewcommand{\arraystretch}{1.1}
\centerline{
\footnotesize
\begin{tabular}{p{3.5cm}|P{1.4cm}P{1.5cm}P{1.5cm}P{1.6cm}P{1.5cm}P{1.4cm}P{1.42cm}}
\toprule
Task Name & Metric & \ourmodelpredictsmall & \ourmodelpredictnine & \ourmodelpredictlargest & \ourmodelconversenine & \ourmodelconverselargest \\
\midrule
BindingDB Patent & PCC & 0.556 & 0.376 & 0.537 & 0.438 & 0.118 \\
BindingDB ic50 & Spearman & 0.425 & 0.313 & 0.465 & 0.443 & 0.361 \\
BindingDB kd & PCC & 0.490 & 0.393 & 0.289 & 0.207 & 0.156 \\
BindingDB ki & PCC & 0.728 & 0.712 & 0.670 & 0.387 & 0.218 \\
Buchwald Hartwig & PCC & 0.920 & 0.918 & 0.903 & 0.574 & 0.818 \\
Caco2 Wang & MAE & 0.619 & 0.491 & 0.479 & 0.588 & 0.383 \\
Clearance Hepatocyte AZ & Spearman & 0.292 & 0.378 & 0.350 & 0.166 & 0.190 \\
Clearance Microsome AZ & Spearman & 0.521 & 0.524 & 0.510 & 0.394 & 0.395 \\
DAVIS & MSE & 0.576 & 0.564 & 0.575 & 0.561 & 0.561 \\
DrugComb Bliss & MAE & 4.088 & 4.286 & 4.157 & 4.454 & 4.519 \\
DrugComb CSS & MAE & 14.568 & 15.370 & 14.925 & 15.960 & 16.649 \\
DrugComb HSA & MAE & 4.063 & 4.282 & 4.178 & 4.486 & 4.529 \\
DrugComb Loewe & MAE & 17.313 & 17.862 & 17.327 & 17.190 & 16.873 \\
DrugComb ZIP & MAE & 3.737 & 3.848 & 3.823 & 4.093 & 4.132 \\
Half Life Obach & Spearman & 0.423 & 0.348 & 0.491 & 0.269 & 0.393 \\
KIBA & MSE & 0.562 & 0.525 & 0.554 & 0.830 & 0.858 \\
LD50 Zhu & MAE & 0.698 & 0.718 & 0.677 & 0.724 & 0.721 \\
Leenay & Spearman & 0.114 & 0.089 & 0.259 & 0.078 & 0.183 \\
Lipophilicity AstraZeneca & MAE & 0.571 & 0.667 & 0.613 & 0.834 & 0.837 \\
OncoPolyPharmacology & PCC & 0.556 & 0.437 & 0.531 & 0.388 & 0.148 \\
PPBR AZ & MAE & 8.813 & 9.177 & 8.792 & 11.004 & 11.025 \\
Protein SAbDab & MAE & 1.117 & 1.022 & 1.072 & 1.348 & 1.173 \\
Solubility AqSolDB & MAE & 0.911 & 1.185 & 0.802 & 1.160 & 1.135 \\
TAP & MAE & 5.498 & 4.839 & 4.088 & 4.611 & 4.444 \\
USPTO & Accuracy & 0.316 & 0.041 & 0.281 & 0.145 & 0.090 \\
USPTO Yields & PCC & 0.471 & 0.002 & 0.350 & 0.114 & 0.002 \\
VDss Lombardo & Spearman & 0.594 & 0.538 & 0.591 & 0.410 & 0.487 \\
\bottomrule
\end{tabular}
}
\end{table}

\newpage

\newpage
\setlength\bibitemsep{3pt}
\clearpage
\printbibliography

@article{fritsch2014characterization,
  title={Characterization of the novel and specific PI3K$\alpha$ inhibitor NVP-BYL719 and development of the patient stratification strategy for clinical trials},
  author={Fritsch, Christine and Huang, Alan and Chatenay-Rivauday, Christian and Schnell, Christian and Reddy, Anupama and Liu, Manway and Kauffmann, Audrey and Guthy, Daniel and Erdmann, Dirk and De Pover, Alain and others},
  journal={Molecular cancer therapeutics},
  volume={13},
  number={5},
  pages={1117--1129},
  year={2014},
  publisher={American Association for Cancer Research}
}

@article{gao2025txagent,
  title={TxAgent: An AI Agent for Therapeutic Reasoning Across a Universe of Tools},
  author={Gao, Shanghua and Zhu, Richard and Kong, Zhenglun and Noori, Ayush and Su, Xiaorui and Ginder, Curtis and Tsiligkaridis, Theodoros and Zitnik, Marinka},
  journal={arXiv preprint arXiv:2503.10970},
  year={2025}
}

@article{gottweis2025towards,
  title={Towards an AI co-scientist},
  author={Gottweis, Juraj and Weng, Wei-Hung and Daryin, Alexander and Tu, Tao and Palepu, Anil and Sirkovic, Petar and Myaskovsky, Artiom and Weissenberger, Felix and Rong, Keran and Tanno, Ryutaro and others},
  journal={arXiv preprint arXiv:2502.18864},
  year={2025}
}

@article{telenti2024large,
  title={Large language models for science and medicine},
  author={Telenti, Amalio and Auli, Michael and Hie, Brian L and Maher, Cyrus and Saria, Suchi and Ioannidis, John PA},
  journal={European journal of clinical investigation},
  volume={54},
  number={6},
  pages={e14183},
  year={2024},
  publisher={Wiley Online Library}
}

@article{kumar2012fragment,
  title={Fragment based drug design: from experimental to computational approaches},
  author={Kumar, Ashutosh and Voet, Arnout and Zhang, Kam YJ},
  journal={Current medicinal chemistry},
  volume={19},
  number={30},
  pages={5128--5147},
  year={2012},
  publisher={Bentham Science Publishers direct}
}

@article{hu2020dual,
  title={Dual PI3K/mTOR inhibitor PKI-402 suppresses the growth of ovarian cancer cells by degradation of Mcl-1 through autophagy},
  author={Hu, Xiaoqing and Xia, Meihui and Wang, Jiabin and Yu, Huimei and Chai, Jiannan and Zhang, Zejun and Sun, Yupei and Su, Jing and Sun, Liankun},
  journal={Biomedicine \& Pharmacotherapy},
  volume={129},
  pages={110397},
  year={2020},
  publisher={Elsevier}
}

@article{thibault2025pi3kalpha,
  title={PI3K$\alpha$-specific inhibitor BYL-719 synergizes with cisplatin in vitro in PIK3CA-mutated ovarian cancer cells},
  author={Thibault, Beno{\^\i}t and Thole, Adrien and D’Angelo, Romina and Basset, C{\'e}line and Guillermet-Guibert, Julie},
  journal={Scientific Reports},
  volume={15},
  number={1},
  pages={6265},
  year={2025},
  publisher={Nature Publishing Group UK London}
}

@article{skarlinski2024language,
  title={Language agents achieve superhuman synthesis of scientific knowledge},
  author={Skarlinski, Michael D and Cox, Sam and Laurent, Jon M and Braza, James D and Hinks, Michaela and Hammerling, Michael J and Ponnapati, Manvitha and Rodriques, Samuel G and White, Andrew D},
  journal={arXiv preprint arXiv:2409.13740},
  year={2024}
}

@article{leontiadou2018insights,
  title={Insights into the mechanism of the PIK3CA E545K activating mutation using MD simulations},
  author={Leontiadou, Hari and Galdadas, Ioannis and Athanasiou, Christina and Cournia, Zoe},
  journal={Scientific reports},
  volume={8},
  number={1},
  pages={15544},
  year={2018},
  publisher={Nature Publishing Group UK London}
}

@article{passarelli2024alpelisib,
  title={Alpelisib for PIK3CA-mutated advanced gynecological cancers: first clues of clinical activity},
  author={Passarelli, Anna and Carbone, Vittoria and Pignata, Sandro and Mazzeo, Roberta and Lorusso, Domenica and Scambia, Giovanni and Canova, Stefania and Di Palma, Teresa and Tasca, Giulia and Mantiero, Mara and others},
  journal={Gynecologic Oncology},
  volume={183},
  pages={61--67},
  year={2024},
  publisher={Elsevier}
}

@article{narayan2021fda,
  title={FDA approval summary: alpelisib plus fulvestrant for patients with HR-positive, HER2-negative, PIK3CA-mutated, advanced or metastatic breast cancer},
  author={Narayan, Preeti and Prowell, Tatiana M and Gao, Jennifer J and Fernandes, Laura L and Li, Emily and Jiang, Xiling and Qiu, Junshan and Fan, Jianghong and Song, Pengfei and Yu, Jingyu and others},
  journal={Clinical Cancer Research},
  volume={27},
  number={7},
  pages={1842--1849},
  year={2021},
  publisher={American Association for Cancer Research}
}

@article{kuo2009frequent,
  title={Frequent activating mutations of PIK3CA in ovarian clear cell carcinoma},
  author={Kuo, Kuan-Ting and Mao, Tsui-Lien and Jones, Si{\^a}n and Veras, Emanuela and Ayhan, Ayse and Wang, Tian-Li and Glas, Ruth and Slamon, Dennis and Velculescu, Victor E and Kuman, Robert J and others},
  journal={The American journal of pathology},
  volume={174},
  number={5},
  pages={1597--1601},
  year={2009},
  publisher={Elsevier}
}

@article{chen2023p110alpha,
  title={P110$\alpha$ inhibitor alpelisib exhibits a synergistic effect with pyrotinib and reverses pyrotinib resistant in HER2+ breast cancer},
  author={Chen, Hao and Si, Yuhao and Wen, Jialiang and Hu, Chunlei and Xia, Erjie and Wang, Yinghao and Wang, Ouchen},
  journal={Neoplasia},
  volume={43},
  pages={100913},
  year={2023},
  publisher={Elsevier}
}

@article{xia2025naturelm,
  title={NatureLM: Deciphering the Language of Nature for Scientific Discovery},
  author={Xia, Yingce and Jin, Peiran and Xie, Shufang and He, Liang and Cao, Chuan and Luo, Renqian and Liu, Guoqing and Wang, Yue and Liu, Zequn and Chen, Yuan-Jyue and others},
  journal={arXiv preprint arXiv:2502.07527},
  year={2025}
}

@article{chen2024trialbench,
  title={{TrialBench}: Multi-modal artificial intelligence-ready clinical trial datasets},
  author={Chen, Jintai and Hu, Yaojun and Wang, Yue and Lu, Yingzhou and Cao, Xu and Lin, Miao and Xu, Hongxia and Wu, Jian and Xiao, Cao and Sun, Jimeng and others},
  journal={arXiv preprint arXiv:2407.00631},
  year={2024}
}

@article{schmidgall2025agent,
  title={Agent Laboratory: Using LLM Agents as Research Assistants},
  author={Schmidgall, Samuel and Su, Yusheng and Wang, Ze and Sun, Ximeng and Wu, Jialian and Yu, Xiaodong and Liu, Jiang and Liu, Zicheng and Barsoum, Emad},
  journal={arXiv preprint arXiv:2501.04227},
  year={2025}
}

@article{parisi2022talm,
  title={Talm: Tool augmented language models},
  author={Parisi, Aaron and Zhao, Yao and Fiedel, Noah},
  journal={arXiv preprint arXiv:2205.12255},
  year={2022}
}

@article{yao2024tree,
  title={Tree of thoughts: Deliberate problem solving with large language models},
  author={Yao, Shunyu and Yu, Dian and Zhao, Jeffrey and Shafran, Izhak and Griffiths, Tom and Cao, Yuan and Narasimhan, Karthik},
  journal={Advances in Neural Information Processing Systems},
  volume={36},
  year={2024}
}

@article{wang2023describe,
  title={Describe, explain, plan and select: Interactive planning with large language models enables open-world multi-task agents},
  author={Wang, Zihao and Cai, Shaofei and Chen, Guanzhou and Liu, Anji and Ma, Xiaojian and Liang, Yitao},
  journal={arXiv preprint arXiv:2302.01560},
  year={2023}
}

@inproceedings{song2023llm,
  title={Llm-planner: Few-shot grounded planning for embodied agents with large language models},
  author={Song, Chan Hee and Wu, Jiaman and Washington, Clayton and Sadler, Brian M and Chao, Wei-Lun and Su, Yu},
  booktitle={Proceedings of the IEEE/CVF International Conference on Computer Vision},
  pages={2998--3009},
  year={2023}
}

@inproceedings{huang2022language,
  title={Language models as zero-shot planners: Extracting actionable knowledge for embodied agents},
  author={Huang, Wenlong and Abbeel, Pieter and Pathak, Deepak and Mordatch, Igor},
  booktitle={International conference on machine learning},
  pages={9118--9147},
  year={2022},
  organization={PMLR}
}

@article{hao2023reasoning,
  title={Reasoning with language model is planning with world model},
  author={Hao, Shibo and Gu, Yi and Ma, Haodi and Hong, Joshua Jiahua and Wang, Zhen and Wang, Daisy Zhe and Hu, Zhiting},
  journal={arXiv preprint arXiv:2305.14992},
  year={2023}
}

@inproceedings{rein2024gpqa,
  title={Gpqa: A graduate-level google-proof q\&a benchmark},
  author={Rein, David and Hou, Betty Li and Stickland, Asa Cooper and Petty, Jackson and Pang, Richard Yuanzhe and Dirani, Julien and Michael, Julian and Bowman, Samuel R},
  booktitle={First Conference on Language Modeling},
  year={2024}
}

@article{wang2024survey,
  title={A survey on large language model based autonomous agents},
  author={Wang, Lei and Ma, Chen and Feng, Xueyang and Zhang, Zeyu and Yang, Hao and Zhang, Jingsen and Chen, Zhiyuan and Tang, Jiakai and Chen, Xu and Lin, Yankai and others},
  journal={Frontiers of Computer Science},
  volume={18},
  number={6},
  pages={186345},
  year={2024},
  publisher={Springer}
}

@article{boiko2023autonomous,
  title={Autonomous chemical research with large language models},
  author={Boiko, Daniil A and MacKnight, Robert and Kline, Ben and Gomes, Gabe},
  journal={Nature},
  volume={624},
  number={7992},
  pages={570--578},
  year={2023},
  publisher={Nature Publishing Group UK London}
}

@article{lu2024ai,
  title={The ai scientist: Towards fully automated open-ended scientific discovery},
  author={Lu, Chris and Lu, Cong and Lange, Robert Tjarko and Foerster, Jakob and Clune, Jeff and Ha, David},
  journal={arXiv preprint arXiv:2408.06292},
  year={2024}
}

@article{qian2023experiential,
  title={Experiential co-learning of software-developing agents},
  author={Qian, Chen and Dang, Yufan and Li, Jiahao and Liu, Wei and Chen, Weize and Yang, Cheng and Liu, Zhiyuan and Sun, Maosong},
  journal={arXiv preprint arXiv:2312.17025},
  year={2023}
}

@article{yang2024swe,
  title={Swe-agent: Agent-computer interfaces enable automated software engineering},
  author={Yang, John and Jimenez, Carlos E and Wettig, Alexander and Lieret, Kilian and Yao, Shunyu and Narasimhan, Karthik and Press, Ofir},
  journal={arXiv preprint arXiv:2405.15793},
  year={2024}
}

@article{talebirad2023multi,
  title={Multi-agent collaboration: Harnessing the power of intelligent llm agents},
  author={Talebirad, Yashar and Nadiri, Amirhossein},
  journal={arXiv preprint arXiv:2306.03314},
  year={2023}
}

@article{hong2023metagpt,
  title={Metagpt: Meta programming for multi-agent collaborative framework},
  author={Hong, Sirui and Zheng, Xiawu and Chen, Jonathan and Cheng, Yuheng and Wang, Jinlin and Zhang, Ceyao and Wang, Zili and Yau, Steven Ka Shing and Lin, Zijuan and Zhou, Liyang and others},
  journal={arXiv preprint arXiv:2308.00352},
  year={2023}
}

@article{qian2023communicative,
  title={Communicative agents for software development},
  author={Qian, Chen and Cong, Xin and Yang, Cheng and Chen, Weize and Su, Yusheng and Xu, Juyuan and Liu, Zhiyuan and Sun, Maosong},
  journal={arXiv preprint arXiv:2307.07924},
  volume={6},
  number={3},
  year={2023}
}

@misc{OpenAIReasoningLLMs,
  title = {Learning to Reason with LLMs},
  author = {{OpenAI}},
  howpublished = {\url{https://openai.com/index/learning-to-reason-with-llms/}},
  year = {2024}, 
  note = {Accessed: \today}
}

@article{shanahan2023role,
  title={Role play with large language models},
  author={Shanahan, Murray and McDonell, Kyle and Reynolds, Laria},
  journal={Nature},
  volume={623},
  number={7987},
  pages={493--498},
  year={2023},
  publisher={Nature Publishing Group UK London}
}

@article{cai2023large,
  title={Large language models as tool makers},
  author={Cai, Tianle and Wang, Xuezhi and Ma, Tengyu and Chen, Xinyun and Zhou, Denny},
  journal={arXiv preprint arXiv:2305.17126},
  year={2023}
}

@article{qin2024tool,
  title={Tool learning with foundation models},
  author={Qin, Yujia and Hu, Shengding and Lin, Yankai and Chen, Weize and Ding, Ning and Cui, Ganqu and Zeng, Zheni and Zhou, Xuanhe and Huang, Yufei and Xiao, Chaojun and others},
  journal={ACM Computing Surveys},
  volume={57},
  number={4},
  pages={1--40},
  year={2024},
  publisher={ACM New York, NY}
}

@article{schick2023toolformer,
  title={Toolformer: Language models can teach themselves to use tools},
  author={Schick, Timo and Dwivedi-Yu, Jane and Dess{\`\i}, Roberto and Raileanu, Roberta and Lomeli, Maria and Hambro, Eric and Zettlemoyer, Luke and Cancedda, Nicola and Scialom, Thomas},
  journal={Advances in Neural Information Processing Systems},
  volume={36},
  pages={68539--68551},
  year={2023}
}

@article{shinn2024reflexion,
  title={Reflexion: Language agents with verbal reinforcement learning},
  author={Shinn, Noah and Cassano, Federico and Gopinath, Ashwin and Narasimhan, Karthik and Yao, Shunyu},
  journal={Advances in Neural Information Processing Systems},
  volume={36},
  year={2024}
}

@article{swanson2024virtual,
  title={The virtual lab: Ai agents design new sars-cov-2 nanobodies with experimental validation},
  author={Swanson, Kyle and Wu, Wesley and Bulaong, Nash L and Pak, John E and Zou, James},
  journal={bioRxiv},
  pages={2024--11},
  year={2024},
  publisher={Cold Spring Harbor Laboratory}
}

@article{phan2025humanity,
  title={Humanity's Last Exam},
  author={Phan, Long and Gatti, Alice and Han, Ziwen and Li, Nathaniel and Hu, Josephina and Zhang, Hugh and Shi, Sean and Choi, Michael and Agrawal, Anish and Chopra, Arnav and others},
  journal={arXiv preprint arXiv:2501.14249},
  year={2025}
}

@article{yao2022react,
  title={React: Synergizing reasoning and acting in language models},
  author={Yao, Shunyu and Zhao, Jeffrey and Yu, Dian and Du, Nan and Shafran, Izhak and Narasimhan, Karthik and Cao, Yuan},
  journal={arXiv preprint arXiv:2210.03629},
  year={2022}
}

@article{mirza2024large,
  title = {Are large language models superhuman chemists?},
  author = {Mirza, Adrian and Alampara, Nawaf and Kunchapu, Sreekanth and Rios-Garcia, Martino and Emoekabu, Benedict and Krishnan, Aswanth and Gupta, Tanya and Schilling-Wilhelmi, Mara and Okereke, Macjonathan and Aneesh, Anagha and others},
  journal = {arXiv preprint arXiv:2404.01475},
  year = {2024}
}

@article{mendez2024mole,
  title={{MolE}: a foundation model for molecular graphs using disentangled attention},
  author={Mendez-Lucio, Oscar and Nicolaou, Christos A and Earnshaw, Berton},
  journal={Nature Communications},
  volume={15},
  number={1},
  pages={9431},
  year={2024},
  publisher={Nature Publishing Group UK London}
}

@article{bubeck2023sparks,
  title={Sparks of artificial general intelligence: Early experiments with {GPT-4}},
  author={Bubeck, Sebastien and Chandrasekaran, Varun and Eldan, Ronen and Gehrke, Johannes and Horvitz, Eric and Kamar, Ece and Lee, Peter and Lee, Yin Tat and Li, Yuanzhi and Lundberg, Scott and others},
  journal={arXiv preprint arXiv:2303.12712},
  year={2023}
}

@article{taylor2022galactica,
  title={Galactica: A large language model for science},
  author={Taylor, Ross and Kardas, Marcin and Cucurull, Guillem and Scialom, Thomas and Hartshorn, Anthony and Saravia, Elvis and Poulton, Andrew and Kerkez, Viktor and Stojnic, Robert},
  journal={arXiv preprint arXiv:2211.09085},
  year={2022}
}

@article{aleixo2023catastrophic,
  title={Catastrophic forgetting in deep learning: A comprehensive taxonomy},
  author={Aleixo, Everton L and Colonna, Juan G and Cristo, Marco and Fernandes, Everlandio},
  journal={arXiv preprint arXiv:2312.10549},
  year={2023}
}

@article{luo2024large,
  title={Large language models surpass human experts in predicting neuroscience results},
  author={Luo, Xiaoliang and Rechardt, Akilles and Sun, Guangzhi and Nejad, Kevin K and Y{\'a}{\~n}ez, Felipe and Yilmaz, Bati and Lee, Kangjoo and Cohen, Alexandra O and Borghesani, Valentina and Pashkov, Anton and others},
  journal={Nature human behaviour},
  pages={1--11},
  year={2024},
  publisher={Nature Publishing Group UK London}
}

@article{schaar2024nicheformer,
  title={Nicheformer: a foundation model for single-cell and spatial omics},
  author={Schaar, Anna C and Tejada-Lapuerta, Alejandro and Palla, Giovanni and Gutgesell, Robert and Halle, Lennard and Minaeva, Mariia and Vornholz, Larsen and Dony, Leander and Drummer, Francesca and Bahrami, Mojtaba and others},
  journal={bioRxiv},
  pages={2024--04},
  year={2024},
  publisher={Cold Spring Harbor Laboratory}
}

@article{cell2sentence,
  title={{Cell2Sentence}: teaching large language models the language of biology},
  author={Levine, Daniel and Rizvi, Syed Asad and L{\'e}vy, Sacha and Pallikkavaliyaveetil, Nazreen and Zhang, David and Chen, Xingyu and Ghadermarzi, Sina and Wu, Ruiming and Zheng, Zihe and Vrkic, Ivan and others},
  journal={BioRxiv},
  pages={2023--09},
  year={2023},
  publisher={Cold Spring Harbor Laboratory}
}

@article{cornman2024omg,
  title={The {OMG} dataset: An Open {MetaGenomic} corpus for mixed-modality genomic language modeling},
  author={Cornman, Andre and West-Roberts, Jacob and Camargo, Antonio Pedro and Roux, Simon and Beracochea, Martin and Mirdita, Milot and Ovchinnikov, Sergey and Hwang, Yunha},
  journal={bioRxiv},
  pages={2024--08},
  year={2024},
  publisher={Cold Spring Harbor Laboratory}
}

@article{dalla2024nucleotide,
  title={Nucleotide Transformer: building and evaluating robust foundation models for human genomics},
  author={Dalla-Torre, Hugo and Gonzalez, Liam and Mendoza-Revilla, Javier and Lopez Carranza, Nicolas and Grzywaczewski, Adam Henryk and Oteri, Francesco and Dallago, Christian and Trop, Evan and de Almeida, Bernardo P and Sirelkhatim, Hassan and others},
  journal={Nature Methods},
  pages={1--11},
  year={2024},
  publisher={Nature Publishing Group US New York}
}

@article{nguyen2024sequence,
  title={Sequence modeling and design from molecular to genome scale with {Evo}},
  author={Nguyen, Eric and Poli, Michael and Durrant, Matthew G and Kang, Brian and Katrekar, Dhruva and Li, David B and Bartie, Liam J and Thomas, Armin W and King, Samuel H and Brixi, Garyk and others},
  journal={Science},
  volume={386},
  number={6723},
  pages={eado9336},
  year={2024},
  publisher={American Association for the Advancement of Science}
}

@misc{anonymous2024parameter,
  title={Parameter Efficient Graph Encoding for Large Language Models},
  author={Anonymous},
  year={2025},
  url={https://openreview.net/forum?id=RbcXV63ZJk}
}

@article{zhang2024instruction,
  title={Instruction tuning for large language models: A survey},
  author={Zhang, Shengyu and Dong, Linfeng and Li, Xiaoya and Zhang, Sen and Sun, Xiaofei and Wang, Shuhe and Li, Jiwei and Hu, Runyi and Zhang, Tianwei and Wu, Fei and others},
  journal={arXiv preprint arXiv:2308.10792},
  year={2023}
}

@article{liu2019text,
  title={Text summarization with pretrained encoders},
  author={Liu, Yang and Lapata, Mirella},
  journal={arXiv preprint arXiv:1908.08345},
  year={2019}
}

@inproceedings{kenton2019bert,
  title={{BERT}: Pre-training of deep bidirectional transformers for language understanding},
  author={Kenton, Jacob Devlin Ming-Wei Chang and Toutanova, Lee Kristina},
  booktitle={Proceedings of naacL-HLT},
  volume={1},
  number={2},
  year={2019},
  organization={Minneapolis, Minnesota}
}

@article{hendrycks2020measuring,
  title={Measuring massive multitask language understanding},
  author={Hendrycks, Dan and Burns, Collin and Basart, Steven and Zou, Andy and Mazeika, Mantas and Song, Dawn and Steinhardt, Jacob},
  journal={arXiv preprint arXiv:2009.03300},
  year={2020}
}

@article{chaves2024txllm,
  title={Tx-LLM: A Large Language Model for Therapeutics},
  author={Chaves, Juan Manuel Zambrano and Wang, Eric and Tu, Tao and Vaishnav, Eeshit Dhaval and Lee, Byron and Mahdavi, S Sara and Semturs, Christopher and Fleet, David and Natarajan, Vivek and Azizi, Shekoofeh},
  journal={arXiv preprint arXiv:2406.06316},
  year={2024}
}

@article{zambaldi2024novo,
  title={De novo design of high-affinity protein binders with AlphaProteo},
  author={Zambaldi, Vinicius and La, David and Chu, Alexander E and Patani, Harshnira and Danson, Amy E and Kwan, Tristan OC and Frerix, Thomas and Schneider, Rosalia G and Saxton, David and Thillaisundaram, Ashok and others},
  journal={arXiv preprint arXiv:2409.08022},
  year={2024}
}

@article{rogers2010extended,
  title={Extended-connectivity fingerprints},
  author={Rogers, David and Hahn, Mathew},
  journal={Journal of chemical information and modeling},
  volume={50},
  number={5},
  pages={742--754},
  year={2010},
  publisher={ACS Publications}
}

@article{huang2024foundation,
  title={A foundation model for clinician-centered drug repurposing},
  author={Huang, Kexin and Chandak, Payal and Wang, Qianwen and Havaldar, Shreyas and Vaid, Akhil and Leskovec, Jure and Nadkarni, Girish N and Glicksberg, Benjamin S and Gehlenborg, Nils and Zitnik, Marinka},
  journal={Nature Medicine},
  pages={1--13},
  year={2024},
  publisher={Nature Publishing Group US New York}
}

@article{team2025gemma3,
  title={Gemma 3 technical report},
  author={Team, Gemma},
  journal={Google DeepMind},
  year={2025}
}

@article{team2024gemma2,
  title={Gemma 2: Improving open language models at a practical size},
  author={Team, Gemma and Riviere, Morgane and Pathak, Shreya and Sessa, Pier Giuseppe and Hardin, Cassidy and Bhupatiraju, Surya and Hussenot, L{\'e}onard and Mesnard, Thomas and Shahriari, Bobak and Ram{\'e}, Alexandre and others},
  journal={arXiv preprint arXiv:2408.00118},
  year={2024}
}

@article{team2024gemma,
  title={Gemma: Open models based on gemini research and technology},
  author={Team, Gemma and Mesnard, Thomas and Hardin, Cassidy and Dadashi, Robert and Bhupatiraju, Surya and Pathak, Shreya and Sifre, Laurent and Rivi{\`e}re, Morgane and Kale, Mihir Sanjay and Love, Juliette and others},
  journal={arXiv preprint arXiv:2403.08295},
  year={2024}
}

@inproceedings{seidl2023enhancing,
  title={Enhancing activity prediction models in drug discovery with the ability to understand human language},
  author={Seidl, Philipp and Vall, Andreu and Hochreiter, Sepp and Klambauer, G{\"u}nter},
  booktitle={International Conference on Machine Learning},
  pages={30458--30490},
  year={2023},
  organization={PMLR}
}

@article{velez2024tdc,
  title={TDC-2: Multimodal foundation for therapeutic science},
  author={Velez-Arce, Alejandro and Huang, Kexin and Li, Michelle M and Lin, Xiang and Gao, Wenhao and Fu, Tianfan and Kellis, Manolis and Pentelute, Bradley L and Zitnik, Marinka},
  journal={bioRxiv},
  pages={2024--06},
  year={2024},
  publisher={Cold Spring Harbor Laboratory}
}

@article{team2023gemini,
  title={Gemini: a family of highly capable multimodal models},
  author={Team, Gemini and Anil, Rohan and Borgeaud, Sebastian and Alayrac, Jean-Baptiste and Yu, Jiahui and Soricut, Radu and Schalkwyk, Johan and Dai, Andrew M and Hauth, Anja and Millican, Katie and others},
  journal={arXiv preprint arXiv:2312.11805},
  year={2023}
}

@article{jumper2021highly,
  title={Highly accurate protein structure prediction with AlphaFold},
  author={Jumper, John and Evans, Richard and Pritzel, Alexander and Green, Tim and Figurnov, Michael and Ronneberger, Olaf and Tunyasuvunakool, Kathryn and Bates, Russ and {\v{Z}}{\'\i}dek, Augustin and Potapenko, Anna and others},
  journal={nature},
  volume={596},
  number={7873},
  pages={583--589},
  year={2021},
  publisher={Nature Publishing Group}
}

@article{tunyasuvunakool2021highly,
  title={Highly accurate protein structure prediction for the human proteome},
  author={Tunyasuvunakool, Kathryn and Adler, Jonas and Wu, Zachary and Green, Tim and Zielinski, Michal and {\v{Z}}{\'\i}dek, Augustin and Bridgland, Alex and Cowie, Andrew and Meyer, Clemens and Laydon, Agata and others},
  journal={Nature},
  volume={596},
  number={7873},
  pages={590--596},
  year={2021},
  publisher={Nature Publishing Group UK London}
}

@article{senior2020improved,
  title={Improved protein structure prediction using potentials from deep learning},
  author={Senior, Andrew W and Evans, Richard and Jumper, John and Kirkpatrick, James and Sifre, Laurent and Green, Tim and Qin, Chongli and {\v{Z}}{\'\i}dek, Augustin and Nelson, Alexander WR and Bridgland, Alex and others},
  journal={Nature},
  volume={577},
  number={7792},
  pages={706--710},
  year={2020},
  publisher={Nature Publishing Group UK London}
}

@article{abramson2024accurate,
  title={Accurate structure prediction of biomolecular interactions with {AlphaFold 3}},
  author={Abramson, Josh and Adler, Jonas and Dunger, Jack and Evans, Richard and Green, Tim and Pritzel, Alexander and Ronneberger, Olaf and Willmore, Lindsay and Ballard, Andrew J and Bambrick, Joshua and others},
  journal={Nature},
  pages={1--3},
  year={2024},
  publisher={Nature Publishing Group UK London}
}

@article{ren2023alphafold,
  title={{AlphaFold} accelerates artificial intelligence powered drug discovery: efficient discovery of a novel {CDK20} small molecule inhibitor},
  author={Ren, Feng and Ding, Xiao and Zheng, Min and Korzinkin, Mikhail and Cai, Xin and Zhu, Wei and Mantsyzov, Alexey and Aliper, Alex and Aladinskiy, Vladimir and Cao, Zhongying and others},
  journal={Chemical science},
  volume={14},
  number={6},
  pages={1443--1452},
  year={2023},
  publisher={Royal Society of Chemistry}
}

@article{m2024augmenting,
  title={Augmenting large language models with chemistry tools},
  author={M. Bran, Andres and Cox, Sam and Schilter, Oliver and Baldassari, Carlo and White, Andrew D and Schwaller, Philippe},
  journal={Nature Machine Intelligence},
  pages={1--11},
  year={2024},
  publisher={Nature Publishing Group UK London}
}

@article{chen2024genept,
  title={{GenePT}: a simple but effective foundation model for genes and cells built from {ChatGPT}},
  author={Chen, Yiqun and Zou, James},
  journal={bioRxiv},
  pages={2023--10},
  year={2024}
}

@article{zhuo2024protllm,
  title={Protllm: An interleaved protein-language llm with protein-as-word pre-training},
  author={Zhuo, Le and Chi, Zewen and Xu, Minghao and Huang, Heyan and Zheng, Heqi and He, Conghui and Mao, Xian-Ling and Zhang, Wentao},
  journal={arXiv preprint arXiv:2403.07920},
  year={2024}
}

@article{kaufmann2023survey,
  title={A survey of reinforcement learning from human feedback},
  author={Kaufmann, Timo and Weng, Paul and Bengs, Viktor and H{\"u}llermeier, Eyke},
  journal={arXiv preprint arXiv:2312.14925},
  year={2023}
}

@article{pei2024biot5,
  title={Biot5: Enriching cross-modal integration in biology with chemical knowledge and natural language associations},
  author={Pei, Qizhi and Zhang, Wei and Zhu, Jinhua and Wu, Kehan and Gao, Kaiyuan and Wu, Lijun and Xia, Yingce and Yan, Rui},
  journal={arXiv preprint arXiv:2310.07276},
  year={2023}
}

@article{yu2024llasmol,
  title={Llasmol: Advancing large language models for chemistry with a large-scale, comprehensive, high-quality instruction tuning dataset},
  author={Yu, Botao and Baker, Frazier N and Chen, Ziqi and Ning, Xia and Sun, Huan},
  journal={arXiv preprint arXiv:2402.09391},
  year={2024}
}

@article{vaswani2023attention,
  title={Attention is all you need},
  author={Vaswani, A},
  journal={Advances in Neural Information Processing Systems},
  year={2017}
}

@article{cui2024scgpt,
  title={scGPT: toward building a foundation model for single-cell multi-omics using generative AI},
  author={Cui, Haotian and Wang, Chloe and Maan, Hassaan and Pang, Kuan and Luo, Fengning and Duan, Nan and Wang, Bo},
  journal={Nature Methods},
  pages={1--11},
  year={2024},
  publisher={Nature Publishing Group US New York}
}

@article{theodoris2023transfer,
  title={Transfer learning enables predictions in network biology},
  author={Theodoris, Christina V and Xiao, Ling and Chopra, Anant and Chaffin, Mark D and Al Sayed, Zeina R and Hill, Matthew C and Mantineo, Helene and Brydon, Elizabeth M and Zeng, Zexian and Liu, X Shirley and others},
  journal={Nature},
  volume={618},
  number={7965},
  pages={616--624},
  year={2023},
  publisher={Nature Publishing Group UK London}
}

@article{mohr2022data,
  title={Data-driven discovery of cardiolipin-selective small molecules by computational active learning},
  author={Mohr, Bernadette and Shmilovich, Kirill and Kleinw{\"a}chter, Isabel S and Schneider, Dirk and Ferguson, Andrew L and Bereau, Tristan},
  journal={Chemical Science},
  volume={13},
  number={16},
  pages={4498--4511},
  year={2022},
  publisher={Royal Society of Chemistry}
}

@article{belenahalli2023development,
  title={Development of machine learning models based on molecular fingerprints for selection of small molecule inhibitors against {JAK2} protein},
  author={Belenahalli Shekarappa, Sharath and Kandagalla, Shivananda and Lee, Julian},
  journal={Journal of Computational Chemistry},
  volume={44},
  number={16},
  pages={1493--1504},
  year={2023},
  publisher={Wiley Online Library}
}

@article{tayyebi2023prediction,
  title={Prediction of organic compound aqueous solubility using machine learning: a comparison study of descriptor-based and fingerprints-based models},
  author={Tayyebi, Arash and Alshami, Ali S and Rabiei, Zeinab and Yu, Xue and Ismail, Nadhem and Talukder, Musabbir Jahan and Power, Jason},
  journal={Journal of Cheminformatics},
  volume={15},
  number={1},
  pages={99},
  year={2023},
  publisher={Springer}
}

@article{torng2019graph,
  title={Graph convolutional neural networks for predicting drug-target interactions},
  author={Torng, Wen and Altman, Russ B},
  journal={Journal of chemical information and modeling},
  volume={59},
  number={10},
  pages={4131--4149},
  year={2019},
  publisher={ACS Publications}
}

@inproceedings{stark2022equibind,
  title={Equibind: Geometric deep learning for drug binding structure prediction},
  author={St{\"a}rk, Hannes and Ganea, Octavian and Pattanaik, Lagnajit and Barzilay, Regina and Jaakkola, Tommi},
  booktitle={International conference on machine learning},
  pages={20503--20521},
  year={2022},
  organization={PMLR}
}

@article{xiong2020pushing,
  title={Pushing the boundaries of molecular representation for drug discovery with the graph attention mechanism},
  author={Xiong, Zhaoping and Wang, Dingyan and Liu, Xiaohong and Zhong, Feisheng and Wan, Xiaozhe and Li, Xutong and Li, Zhaojun and Luo, Xiaomin and Chen, Kaixian and Jiang, Hualiang and others},
  journal={Journal of medicinal chemistry},
  volume={63},
  number={16},
  pages={8749--8760},
  year={2019},
  publisher={ACS Publications}
}

@article{heid2022machine,
  title={Machine learning of reaction properties via learned representations of the condensed graph of reaction},
  author={Heid, Esther and Green, William H},
  journal={Journal of Chemical Information and Modeling},
  volume={62},
  number={9},
  pages={2101--2110},
  year={2021},
  publisher={ACS Publications}
}

@article{yang2019analyzing,
%   title={Analyzing learned molecular representations for property prediction},
%   author={Yang, Kevin and Swanson, Kyle and Jin, Wengong and Coley, Connor and Eiden, Philipp and Gao, Hua and Guzman-Perez, Angel and Hopper, Timothy and Kelley, Brian and Mathea, Miriam and others},
%   journal={Journal of chemical information and modeling},
%   volume={59},
%   number={8},
%   pages={3370--3388},
%   year={2019},
%   publisher={ACS Publications}
% }

@article{morrone2020docking,
  title={Combining docking pose rank and structure with deep learning improves protein--ligand binding mode prediction over a baseline docking approach},
  author={Morrone, Joseph A and Weber, Jeffrey K and Huynh, Tien and Luo, Heng and Cornell, Wendy D},
  journal={Journal of chemical information and modeling},
  volume={60},
  number={9},
  pages={4170--4179},
  year={2020},
  publisher={ACS Publications}
}

@article{lin2023evolutionary,
  title={Evolutionary-scale prediction of atomic-level protein structure with a language model},
  author={Lin, Zeming and Akin, Halil and Rao, Roshan and Hie, Brian and Zhu, Zhongkai and Lu, Wenting and Smetanin, Nikita and Verkuil, Robert and Kabeli, Ori and Shmueli, Yaniv and others},
  journal={Science},
  volume={379},
  number={6637},
  pages={1123--1130},
  year={2023},
  publisher={American Association for the Advancement of Science}
}

@article{ferruz2022protgpt2,
  title={Prot{GPT2} is a deep unsupervised language model for protein design},
  author={Ferruz, Noelia and Schmidt, Steffen and H{\"o}cker, Birte},
  journal={Nature communications},
  volume={13},
  number={1},
  pages={4348},
  year={2022},
  publisher={Nature Publishing Group UK London}
}

@article{alley2019unified,
  title={Unified rational protein engineering with sequence-based deep representation learning},
  author={Alley, Ethan C and Khimulya, Grigory and Biswas, Surojit and AlQuraishi, Mohammed and Church, George M},
  journal={Nature methods},
  volume={16},
  number={12},
  pages={1315--1322},
  year={2019},
  publisher={Nature Publishing Group US New York}
}

@article{rives2019biological,
  title={Biological structure and function emerge from scaling unsupervised learning to 250 million protein sequences},
  author={Rives, Alexander and Meier, Joshua and Sercu, Tom and Goyal, Siddharth and Lin, Zeming and Liu, Jason and Guo, Demi and Ott, Myle and Zitnick, C Lawrence and Ma, Jerry and others},
  journal={Proceedings of the National Academy of Sciences},
  volume={118},
  number={15},
  pages={e2016239118},
  year={2021},
  publisher={National Acad Sciences}
}

@article{stokes2020deep,
  title={A deep learning approach to antibiotic discovery},
  author={Stokes, Jonathan M and Yang, Kevin and Swanson, Kyle and Jin, Wengong and Cubillos-Ruiz, Andres and Donghia, Nina M and MacNair, Craig R and French, Shawn and Carfrae, Lindsey A and Bloom-Ackermann, Zohar and others},
  journal={Cell},
  volume={180},
  number={4},
  pages={688--702},
  year={2020},
  publisher={Elsevier}
}

@article{huang2022artificial,
  title={Artificial intelligence foundation for therapeutic science},
  author={Huang, Kexin and Fu, Tianfan and Gao, Wenhao and Zhao, Yue and Roohani, Yusuf and Leskovec, Jure and Coley, Connor W and Xiao, Cao and Sun, Jimeng and Zitnik, Marinka},
  journal={Nature chemical biology},
  volume={18},
  number={10},
  pages={1033--1036},
  year={2022},
  publisher={Nature Publishing Group US New York}
}

@article{huang2021therapeutics,
  title={Therapeutics data commons: Machine learning datasets and tasks for drug discovery and development},
  author={Huang, Kexin and Fu, Tianfan and Gao, Wenhao and Zhao, Yue and Roohani, Yusuf and Leskovec, Jure and Coley, Connor W and Xiao, Cao and Sun, Jimeng and Zitnik, Marinka},
  journal={arXiv preprint arXiv:2102.09548},
  year={2021}
}

@article{hinkson2020accelerating,
  title={Accelerating therapeutics for opportunities in medicine: a paradigm shift in drug discovery},
  author={Hinkson, Izumi V and Madej, Benjamin and Stahlberg, Eric A},
  journal={Frontiers in pharmacology},
  volume={11},
  pages={770},
  year={2020},
  publisher={Frontiers Media SA}
}

@article{sun2022why,
  title={Why 90\% of clinical drug development fails and how to improve it?},
  author={Sun, Duxin and Gao, Wei and Hu, Hongxiang and Zhou, Simon},
  journal={Acta Pharmaceutica Sinica B},
  volume={12},
  number={7},
  pages={3049--3062},
  year={2022},
  publisher={Elsevier}
}

@inproceedings{longpre2023flan,
  title={The {FLAN} collection: Designing data and methods for effective instruction tuning},
  author={Longpre, Shayne and Hou, Le and Vu, Tu and Webson, Albert and Chung, Hyung Won and Tay, Yi and Zhou, Denny and Le, Quoc V and Zoph, Barret and Wei, Jason and others},
  booktitle={International Conference on Machine Learning},
  pages={22631--22648},
  year={2023},
  organization={PMLR}
}

@article{brown2020language,
  title={Language models are few-shot learners},
  author={Brown, Tom and Mann, Benjamin and Ryder, Nick and Subbiah, Melanie and Kaplan, Jared D and Dhariwal, Prafulla and Neelakantan, Arvind and Shyam, Pranav and Sastry, Girish and Askell, Amanda and others},
  journal={Advances in neural information processing systems},
  volume={33},
  pages={1877--1901},
  year={2020}
}

@article{landrum2016rdkit,
  author = {Landrum, Greg},
  description = {Release 2016_09_4 (Q3 2016) Release · rdkit/rdkit},
  title = {RDKit: Open-Source Cheminformatics Software},
  url = {https://github.com/rdkit/rdkit/releases/tag/Release_2016_09_4},
  year = 2016
}

@article{dalke2019chemfp,
  title={The chemfp project},
  author={Dalke, Andrew},
  journal={Journal of cheminformatics},
  volume={11},
  pages={1--21},
  year={2019},
  publisher={Springer}
}

@article{sievers2011fast,
  title={Fast, scalable generation of high-quality protein multiple sequence alignments using Clustal Omega},
  author={Sievers, Fabian and Wilm, Andreas and Dineen, David and Gibson, Toby J and Karplus, Kevin and Li, Weizhong and Lopez, Rodrigo and McWilliam, Hamish and Remmert, Michael and S{\"o}ding, Johannes and others},
  journal={Molecular systems biology},
  volume={7},
  number={1},
  pages={539},
  year={2011},
  publisher={John Wiley \& Sons, Ltd Chichester, UK}
}

@article{siramshetty2021validating,
  title={Validating ADME QSAR models using marketed drugs},
  author={Siramshetty, Vishal and Williams, Jordan and Nguyen, {\DH}ac-Trung and Neyra, Jorge and Southall, Noel and Math{\'e}, Ewy and Xu, Xin and Shah, Pranav},
  journal={SLAS DISCOVERY: Advancing the Science of Drug Discovery},
  volume={26},
  number={10},
  pages={1326--1336},
  year={2021},
  publisher={SAGE Publications Sage CA: Los Angeles, CA}
}

@article{turon2023first,
  title={First fully-automated {AI/ML} virtual screening cascade implemented at a drug discovery centre in Africa},
  author={Turon, Gemma and Hlozek, Jason and Woodland, John G and Kumar, Ankur and Chibale, Kelly and Duran-Frigola, Miquel},
  journal={Nature Communications},
  volume={14},
  number={1},
  pages={5736},
  year={2023},
  publisher={Nature Publishing Group UK London}
}

@article{bera2022simgcn,
  title={SimGCN for TDC Benchmarks},
  author={Suman Bera and Jason Dent and Gurbinder Gill and Andrew Stolman and Bo Wu},
  year={2022},
}

@misc{fonteno2023predicting,
  title={Predicting a Compounds Blood-Brain-Barrier Permeability with Lantern Pharma's {AI} and {ML} Platform, {RADR}},
  author={Rick Fontenot and Umesh Kathad and Joseph McDermott and Drew Sturtevant and Panna Sharma and Peter Carr},
  year={2023},
}

@article{boral2022accountable,
  title={Accountable prediction of drug ADMET Properties with molecular descriptors},
  author={Boral, Nilavo and Ghosh, Promita and Goswami, Adhish and Bhattacharyya, Malay},
  journal={bioRxiv},
  pages={2022--06},
  year={2022},
  publisher={Cold Spring Harbor Laboratory}
}

@article{plonka2021cyplebrity,
  title={CYPlebrity: Machine learning models for the prediction of inhibitors of cytochrome {P450} enzymes},
  author={Plonka, Wojciech and Stork, Conrad and {\v{S}}{\'\i}cho, Martin and Kirchmair, Johannes},
  journal={Bioorganic \& medicinal chemistry},
  volume={46},
  pages={116388},
  year={2021},
  publisher={Elsevier}
}

@article{hu2019strategies,
  title={Strategies for pre-training graph neural networks},
  author={Hu, Weihua and Liu, Bowen and Gomes, Joseph and Zitnik, Marinka and Liang, Percy and Pande, Vijay and Leskovec, Jure},
  journal={arXiv preprint arXiv:1905.12265},
  year={2019}
}

@article{huang2020deeppurpose,
  title={DeepPurpose: a deep learning library for drug--target interaction prediction},
  author={Huang, Kexin and Fu, Tianfan and Glass, Lucas M and Zitnik, Marinka and Xiao, Cao and Sun, Jimeng},
  journal={Bioinformatics},
  volume={36},
  number={22-23},
  pages={5545--5547},
  year={2020},
  publisher={Oxford University Press}
}

@misc{euclia2023half,
  author = {Euclia},
  year = {2023},
  publisher = {GitHub},
  journal = {GitHub repository},
  howpublished = {\url{https://github.com/euclia/public-models}},
}

@article{rivera2024silico,
  title={In silico Evaluation of the Feasibility of Magnolia officinalis Electron-shuttling Compounds as Parkinson’s Disease Remedy},
  author={Rivera, Zaina Allyson and Tayo, Lemmuel and Chen, Bor-Yann and Tsai, Po-Wei},
  journal={Letters in Drug Design \& Discovery},
  volume={21},
  number={14},
  pages={3039--3048},
  year={2024},
  publisher={Bentham Science Publishers}
}

@article{huang2022unified,
  title={A Unified System for Molecular Property Predictions: Oloren ChemEngine and its Applications},
  author={Huang, David and Chowdhuri, Sauhaarda Raunak and Li, Andrew and Li, Alex and Agrawal, Ayush and Gano, Kameron and Zhu, Andy},
  year={2022}
}

@article{alves2015predicting,
  title={Predicting chemically-induced skin reactions. Part I: QSAR models of skin sensitization and their application to identify potentially hazardous compounds},
  author={Alves, Vinicius M and Muratov, Eugene and Fourches, Denis and Strickland, Judy and Kleinstreuer, Nicole and Andrade, Carolina H and Tropsha, Alexander},
  journal={Toxicology and applied pharmacology},
  volume={284},
  number={2},
  pages={262--272},
  year={2015},
  publisher={Elsevier}
}

@article{lagunin2009computer,
  title={Computer-aided prediction of rodent carcinogenicity by PASS and CISOC-PSCT},
  author={Lagunin, Alexey and Filimonov, Dmitrii and Zakharov, Alexey and Xie, Wei and Huang, Ying and Zhu, Fucheng and Shen, Tianxiang and Yao, Jianhua and Poroikov, Vladimir},
  journal={QSAR \& Combinatorial Science},
  volume={28},
  number={8},
  pages={806--810},
  year={2009},
  publisher={Wiley Online Library}
}

@article{shermukhamedov2023structure,
  title={Structure to Property: Chemical Element Embeddings and a Deep Learning Approach for Accurate Prediction of Chemical Properties},
  author={Shermukhamedov, Shokirbek and Mamurjonova, Dilorom and Probst, Michael},
  journal={arXiv preprint arXiv:2309.09355},
  year={2023}
}

@article{li2021trimnet,
  title={TrimNet: learning molecular representation from triplet messages for biomedicine},
  author={Li, Pengyong and Li, Yuquan and Hsieh, Chang-Yu and Zhang, Shengyu and Liu, Xianggen and Liu, Huanxiang and Song, Sen and Yao, Xiaojun},
  journal={Briefings in Bioinformatics},
  volume={22},
  number={4},
  pages={bbaa266},
  year={2021},
  publisher={Oxford University Press}
}

@article{korotcov2017comparison,
  title={Comparison of deep learning with multiple machine learning methods and metrics using diverse drug discovery data sets},
  author={Korotcov, Alexandru and Tkachenko, Valery and Russo, Daniel P and Ekins, Sean},
  journal={Molecular pharmaceutics},
  volume={14},
  number={12},
  pages={4462--4475},
  year={2017},
  publisher={ACS Publications}
}

@article{karim2021cardiotox,
  title={CardioTox net: a robust predictor for hERG channel blockade based on deep learning meta-feature ensembles},
  author={Karim, Abdul and Lee, Matthew and Balle, Thomas and Sattar, Abdul},
  journal={Journal of Cheminformatics},
  volume={13},
  pages={1--13},
  year={2021},
  publisher={Springer}
}

@article{lie2021covid,
  title={COVID-19 multi-targeted drug repurposing using few-shot learning},
  author={Liu, Yang and Wu, You and Shen, Xiaoke and Xie, Lei},
  journal={Frontiers in Bioinformatics},
  volume={1},
  pages={693177},
  year={2021},
  publisher={Frontiers Media SA}
}

@article{haneczok2021machine,
  title={Machine learning enabled identification of potential SARS-CoV-2 3CLpro inhibitors based on fixed molecular fingerprints and Graph-CNN neural representations},
  author={Haneczok, Jacek and Delijewski, Marcin},
  journal={Journal of Biomedical Informatics},
  volume={119},
  pages={103821},
  year={2021},
  publisher={Elsevier}
}

@article{li2017learning,
  title={Learning graph-level representation for drug discovery},
  author={Li, Junying and Cai, Deng and He, Xiaofei},
  journal={arXiv preprint arXiv:1709.03741},
  year={2017}
}

@article{vu2019bcl,
  title={BCL:: Mol2D—a robust atom environment descriptor for QSAR modeling and lead optimization},
  author={Vu, Oanh and Mendenhall, Jeffrey and Altarawy, Doaa and Meiler, Jens},
  journal={Journal of computer-aided molecular design},
  volume={33},
  pages={477--486},
  year={2019},
  publisher={Springer}
}

@article{probst2022reaction,
  title={Reaction classification and yield prediction using the differential reaction fingerprint DRFP},
  author={Probst, Daniel and Schwaller, Philippe and Reymond, Jean-Louis},
  journal={Digital discovery},
  volume={1},
  number={2},
  pages={91--97},
  year={2022},
  publisher={Royal Society of Chemistry}
}

@article{chen2020predicting,
  title={Predicting antibody developability from sequence using machine learning},
  author={Chen, Xingyao and Dougherty, Thomas and Hong, Chan and Schibler, Rachel and Zhao, Yi Cong and Sadeghi, Reza and Matasci, Naim and Wu, Yi-Chieh and Kerman, Ian},
  journal={biorxiv},
  pages={2020--06},
  year={2020},
  publisher={Cold Spring Harbor Laboratory}
}

@article{leenay2019large,
  title={Large dataset enables prediction of repair after CRISPR--Cas9 editing in primary T cells},
  author={Leenay, Ryan T and Aghazadeh, Amirali and Hiatt, Joseph and Tse, David and Roth, Theodore L and Apathy, Ryan and Shifrut, Eric and Hultquist, Judd F and Krogan, Nevan and Wu, Zhenqin and others},
  journal={Nature biotechnology},
  volume={37},
  number={9},
  pages={1034--1037},
  year={2019},
  publisher={Nature Publishing Group US New York}
}

@article{kalemati2023bicomp,
  title={BiComp-DTA: Drug-target binding affinity prediction through complementary biological-related and compression-based featurization approach},
  author={Kalemati, Mahmood and Zamani Emani, Mojtaba and Koohi, Somayyeh},
  journal={PLOS Computational Biology},
  volume={19},
  number={3},
  pages={e1011036},
  year={2023},
  publisher={Public Library of Science San Francisco, CA USA}
}

@article{kinnings2011machine,
  title={A machine learning-based method to improve docking scoring functions and its application to drug repurposing},
  author={Kinnings, Sarah L and Liu, Nina and Tonge, Peter J and Jackson, Richard M and Xie, Lei and Bourne, Philip E},
  journal={Journal of chemical information and modeling},
  volume={51},
  number={2},
  pages={408--419},
  year={2011},
  publisher={ACS Publications}
}

@article{wei2021deeppla,
  title={DeepPLA: a novel deep learning-based model for protein-ligand binding affinity prediction},
  author={Wei, Bomin and Gong, Xiang},
  year={2021},
}

@article{lam2023otter,
  title={Otter-Knowledge: benchmarks of multimodal knowledge graph representation learning from different sources for drug discovery},
  author={Lam, Hoang Thanh and Sbodio, Marco Luca and Galindo, Marcos Mart{\'\i}nez and Zayats, Mykhaylo and Fernandez-Diaz, Raul and Valls, Victor and Picco, Gabriele and Ramis, Cesar Berrospi and Lopez, Vanessa},
  journal={arXiv preprint arXiv:2306.12802},
  year={2023}
}

@article{pei2023breaking,
  title={Breaking the barriers of data scarcity in drug--target affinity prediction},
  author={Pei, Qizhi and Wu, Lijun and Zhu, Jinhua and Xia, Yingce and Xie, Shufang and Qin, Tao and Liu, Haiguang and Liu, Tie-Yan and Yan, Rui},
  journal={Briefings in Bioinformatics},
  volume={24},
  number={6},
  pages={bbad386},
  year={2023},
  publisher={Oxford University Press}
}

@article{raimondi2021novel,
  title={A novel method for data fusion over entity-relation graphs and its application to protein--protein interaction prediction},
  author={Raimondi, Daniele and Simm, Jaak and Arany, Adam and Moreau, Yves},
  journal={Bioinformatics},
  volume={37},
  number={16},
  pages={2275--2281},
  year={2021},
  publisher={Oxford University Press}
}

@article{lind2019predicting,
  title={Predicting drug activity against cancer cells by random forest models based on minimal genomic information and chemical properties},
  author={Lind, Alex P and Anderson, Peter C},
  journal={PloS one},
  volume={14},
  number={7},
  pages={e0219774},
  year={2019},
  publisher={Public Library of Science San Francisco, CA USA}
}

@article{xia2018predicting,
  title={Predicting tumor cell line response to drug pairs with deep learning},
  author={Xia, Fangfang and Shukla, Maulik and Brettin, Thomas and Garcia-Cardona, Cristina and Cohn, Judith and Allen, Jonathan E and Maslov, Sergei and Holbeck, Susan L and Doroshow, James H and Evrard, Yvonne A and others},
  journal={BMC bioinformatics},
  volume={19},
  pages={71--79},
  year={2018},
  publisher={Springer}
}

@article{preuer2018deepsynergy,
  title={DeepSynergy: predicting anti-cancer drug synergy with Deep Learning},
  author={Preuer, Kristina and Lewis, Richard PI and Hochreiter, Sepp and Bender, Andreas and Bulusu, Krishna C and Klambauer, G{\"u}nter},
  journal={Bioinformatics},
  volume={34},
  number={9},
  pages={1538--1546},
  year={2018},
  publisher={Oxford University Press}
}

@article{wong2020mipdh,
  title={MIPDH: a novel computational model for predicting microRNA--mRNA interactions by DeepWalk on a heterogeneous network},
  author={Wong, Leon and You, Zhu-Hong and Guo, Zhen-Hao and Yi, Hai-Cheng and Chen, Zhan-Heng and Cao, Mei-Yuan},
  journal={ACS omega},
  volume={5},
  number={28},
  pages={17022--17032},
  year={2020},
  publisher={ACS Publications}
}

@article{gfeller2023improved,
  title={Improved predictions of antigen presentation and TCR recognition with MixMHCpred2. 2 and PRIME2. 0 reveal potent SARS-CoV-2 CD8+ T-cell epitopes},
  author={Gfeller, David and Schmidt, Julien and Croce, Giancarlo and Guillaume, Philippe and Bobisse, Sara and Genolet, Raphael and Queiroz, Lise and Cesbron, Julien and Racle, Julien and Harari, Alexandre},
  journal={Cell Systems},
  volume={14},
  number={1},
  pages={72--83},
  year={2023},
  publisher={Elsevier}
}

@article{motmaen2023peptide,
  title={Peptide-binding specificity prediction using fine-tuned protein structure prediction networks},
  author={Motmaen, Amir and Dauparas, Justas and Baek, Minkyung and Abedi, Mohamad H and Baker, David and Bradley, Philip},
  journal={Proceedings of the National Academy of Sciences},
  volume={120},
  number={9},
  pages={e2216697120},
  year={2023},
  publisher={National Acad Sciences}
}

@article{weber2021titan,
  title={{TITAN}: {T-cell} receptor specificity prediction with bimodal attention networks},
  author={Weber, Anna and Born, Jannis and Rodriguez Mart{\'\i}nez, Mar{\'\i}a},
  journal={Bioinformatics},
  volume={37},
  number={Supplement\_1},
  pages={i237--i244},
  year={2021},
  publisher={Oxford University Press}
}

@article{fu2022hint,
  title={Hint: Hierarchical interaction network for clinical-trial-outcome predictions},
  author={Fu, Tianfan and Huang, Kexin and Xiao, Cao and Glass, Lucas M and Sun, Jimeng},
  journal={Patterns},
  volume={3},
  number={4},
  year={2022},
  publisher={Elsevier}
}

@article{zheng2019predicting,
  title={Predicting retrosynthetic reactions using self-corrected transformer neural networks},
  author={Zheng, Shuangjia and Rao, Jiahua and Zhang, Zhongyue and Xu, Jun and Yang, Yuedong},
  journal={Journal of chemical information and modeling},
  volume={60},
  number={1},
  pages={47--55},
  year={2019},
  publisher={ACS Publications}
}
\balance
\clearpage
\end{appendices}
\end{refsection}

\end{document}